\documentclass[10pt,reqno]{amsart}

\usepackage{amsthm,amsmath,amsfonts,amssymb}
\usepackage{bm}
\usepackage[dvipdfmx]{graphicx}
\usepackage{subcaption}
\usepackage[font=small,labelfont=bf]{caption}

\usepackage{xcolor}
\usepackage{fullpage}
\usepackage{appendix}

\usepackage[sort&compress, numbers]{natbib}

\usepackage{hyperref}					
\hypersetup{colorlinks,					
	linkcolor=blue,
	citecolor=red}

\newtheorem{theorem}{Theorem}[section]
\newtheorem{corollary}{Corollary}[section]
\newtheorem{lemma}{Lemma}[section]

\newtheorem{assumption}{Assumption}[section]
\theoremstyle{definition}
\newtheorem{definition}{Definition}[section]
\newtheorem{remark}{Remark}[section]

\usepackage{amsaddr}
\usepackage{verbatim}

\def\R{\mathbb{R}}
\def\E{\mathbb{E}}
\def\P{\mathbb{P}}

\def\II{\mathcal{I}}
\def\JJ{\mathcal{J}}
\def\FF{\mathcal{F}}

\def\init{\mathcal{E}}

\newcommand{\wt}{\widetilde}
\renewcommand{\d}{{\,\rm d}}
\def\llan{\left\langle}
\def\rran{\right\rangle}

\allowdisplaybreaks

\title{Convergence of Stochastic Gradient Methods for Wide Two-Layer Physics-Informed Neural Networks 
for the Poisson Equation}
\author{ Bangti Jin\and Longjun Wu} 
\address{ Department of Mathematics, The Chinese University of Hong Kong, Shatin, N.T., Hong Kong, P.R. China }

\begin{document}

\begin{abstract}
Physics informed neural networks (PINNs) represent a very popular class of neural solvers for partial differential equations. In practice, one often employs stochastic gradient descent type algorithms to train the neural network. Therefore, the convergence guarantee of stochastic gradient descent is of fundamental importance. In this work, we establish the linear convergence of stochastic gradient descent / flow in training over-parameterized 
two layer PINNs with a general class of activation functions for solving one model second-order elliptic problem, i.e., the Poisson equation, in the sense of high probability. These results extend the existing result 
\cite{Gao2023} in which gradient descent was analyzed. The challenge of the analysis lies in handling the dynamic randomness
introduced by stochastic optimization methods. The key of the analysis lies in ensuring the positive definiteness of suitable Gram matrices during the training. The analysis sheds insight into the dynamics of the 
optimization process, and provides guarantees on physics informed neural networks 
trained by stochastic algorithms.
\end{abstract}

\maketitle

\section{Introduction}
Partial differential equations (PDEs) represent one very popular and flexible class of mathematical models in nearly all disciplines in natural science and engineering, and their numerical solutions is of critical importance.
One major hurdle to solve PDEs is the notorious curse of dimensionality, i.e., the computational complexity grows exponentially with the problem dimensionality. Recently, deep learning using deep neural networks (DNNs) has emerged as a powerful tool for solving PDEs, and it  has received a lot of attention due to its tremendous potentials to break the curse 
\cite{E2018,Han2018,Han2020}. Various deep learning methods have been proposed to solve PDEs \cite{Raissi2019,E2018,Zang2020,Chen2023}. We refer interested readers to the reviews \cite{BeckKuckuck:2023,Tanyu2023} for details.

Among various existing neural PDE solvers, the most popular method is physics-informed
neural network (PINN) due to Raissi et al \cite{Raissi2019}, which can be traced back to the seminal work \cite{Lagaris1998} in the 1990s. PINN is based on the principle of PDE residual minimization, i.e., the loss is a suitable weighted combination of the residuals of the PDE (in the domain) and boundary condition(s) (on the boundary), commonly measured in the standard $L^2$ norms, and employs DNNs as the ansatz space to approximate the PDE solutions. By the construction, PINN directly integrates the physical knowledge encoded in the differential equation into neural networks. In practice, one learns neural network parameters by training the empirical losses using suitable optimization algorithms, including gradient descent, stochastic gradient descent, Adam \cite{Kingma:2017}, and limited-memory BFGS \cite{ByrdLu:1995} etc. The method enjoys a number of distinct features, e.g., ease of implementation, flexibility of the equation type (elliptic, parabolic and hyperbolic), and strong empirical performance for many PDEs problems. Indeed, it has shown impressive empirical performance across a wide range of challenging direct and inverse problems associated with PDEs, e.g., Navier-Stokes equations \cite{JinCai:2021,Eivazi:2022}, Hamilton-Jacobi-Bellman equation \cite{Sirignano2018}, 
and various PDE inverse problems \cite{Zhang2023,JinLiQuanZhou:2024,CenJinLiZhou:2025}. These results clearly show its significant potentials in diverse scenarios, and thus it has received immense attention within the scientific and engineering computing community \cite{Cuomo2022}.  
 
Despite the empirical successes of PINNs, the theoretical understanding of PINNs is still largely in its infancy. See \cite{Shin:2024theoretical} for a recent survey of relevant mathematical theory of PINNs.
 Several works have also investigated the generalization error, and derived error bounds between the DNN approximation and exact solution in terms of DNN architecture parameters and the number of sampling points etc. under suitable \textit{a priori} regularity assumption on the exact solution \cite{Mishra2023,DeRyckMishra:2024,Hu2024}. The analysis is mostly conducted under the assumption of zero or small training error in the optimization procedure (see, e.g., \cite{Shin:2020Convergence,shin:2023Error} for the theory for various linear PDEs). However, due to the nonlinearity of the activation function in the neural network, the empirical loss is highly nonconvex with respect to the DNN parameters, and thus it is very challenging to find a global minimizer to the empirical loss, or to ensure a small training error \textit{a priori}. In practice, the choice of a suitable optimization algorithm is crucial for PINNs to achieve satisfactory numerical results \cite{Bihlo:2024,UrbanPons:2025}. The mathematical study on the convergence of the optimizers for PINN training is of fundamental importance towards establishing a complete mathematical theory of PINNs.

There are several works on the analysis of the optimization process
within the framework of neural tangent kernel (NTK) \cite{Jacot2018}. The key observation of the NTK is that it is essentially deterministic at initialization and then keeps unchanged during the 
training process when the width of the network tends to infinity. Based on this important observation, 
Du et al \cite{Du2019} proved that gradient descent (GD) can find a global minimum of the least-squares loss involving over-parameterized 
two-layer neural networks for the regression task, and later the authors extended the analysis to deep neural 
networks (DNNs) \cite{Du2019deep}. Allen-Zhu et al \cite{Allen2019} investigated both the GD and stochastic gradient descent (SGD) for over-parameterized DNN, convolutional neural networks (CNN), and residual neural networks (ResNet). These works focus on the mean squared loss for the standard regression task. 

When compared with the standard regression task, the convergence analysis of PINNs is more involved: the PINN loss includes both interior and
boundary type terms, and involves partial derivatives of the neural network. 
The first breakthrough on the convergence of optimization algorithms for training PINNs for the standard parabolic equation is due to Gao, Gu and Ng \cite{Gao2023}, who proved that GD 
can reach a global minimizer of over-parameterized two-layer PINNs with the \(\mathrm{ReLU}^3\) activation. (The ReLU$^3$ activation ensures the well definedness of the PINN loss.)
The key part of the analysis, compared to the classical regression and classification problems, is to handle the intricate interaction among the trainable parameters caused by the physics-informed loss, and to establish the positive definiteness of the associated Gram matrices under the ReLU$^3$ activation. This analysis was recently refined by Xu et al \cite{XuDuHuang:2024} using a different error decomposition. See also \cite{LuoYang:2024} for an analysis of gradient flow under some technical assumptions.
More recently, Xu et al \cite{Xu2024} studied the implicit gradient descent (IGD) for training PINNs, and proved that for a smooth globally Lipschitz non-polynomial activation $\sigma$,
IGD converges to a globally optimal solution at a linear rate. Moreover, they observed that the learning rate can be chosen more flexibly because of the unconditional stability of IGD. All the above-mentioned works focus on the over-parameterized regime, where the number of DNN parameters far exceeds the number of sampling points.
In the under-parameterized regime, Niessen and Muller \cite{Niessen2025} very recently provided the 
optimization guarantee of projected (stochastic) gradient descent for two-layer PINNs with finite width, and derived an overall error bound based on an existing approximation estimate.
However, these important works focus mostly on deterministic algorithms, and the analysis does not extend directly to stochastic methods.

Stochastic gradient descent (SGD) and its variants are simple and powerful methods for training PINNs in practice and have been widely adopted. 
They are prevalent in training neural PDE solvers because of their low computational cost 
and excellent generalization performance. To the best of our knowledge,  
the convergence analysis of SGD for training PINNs remains missing, although 
the previous works offer the analysis for regression problems \cite{Allen2019,Cao2024}. 
Due to the intricate loss of PINNs (involving various partial derivatives) and the dynamic randomness induced by SGM,
it is more challenging to investigate the training dynamics.
Indeed, the convergence analysis requires the bound on all parameters but the stochasticity of the iterations allow only 
estimating the parameters in a high probability sense.
Moreover, the number of trainable parameters increases as the neural network gets wider, which requires new techniques to control them uniformly. 

In view of the wide adoption of SGD type algorithms for training PINNs in practice, it is important to analyze its convergence. This motivates the convergence analysis of stochastic gradient methods (SGMs) for over-parameterized two-layer PINNs, 
including stochastic gradient descent (SGD) and stochastic gradient flow (SGF), i.e., the optimization
problem with the objective given by (\ref{PINN Loss}) and trained by SGM. This represents the main contribution of this work, i.e., linear convergence of SGMs to a global minimum for training the PINN loss in the high probability sense. In the analysis, we assume that the nonlinear activation function $\sigma$ is locally Lipschitz, which is milder than that in existing works \cite{Gao2023,Xu2024}. First, we prove that with high probability, the relevant Gram matrices stay positive definite during  the training dynamics under some mild assumptions.
Second, we track the parameters and the loss at each step and prove that if the Gram matrices are positive definite, the SGD algorithm can find a zero minimum in the sense of expectation for the optimization problem. 
Our analysis applies to any unbiased SGD with a fixed set of sampling points, i.e., the results hold regardless of the specific sampling method. However, the analysis does not include the variance of the stochastic estimates, which would be necessary for understanding the impact of batch sizes on the convergence; see Remarks \ref{rmk:variance} and \ref{rmk:batch_size} for more details.
Third, we establish and study the stochastic differential equation satisfied by the parameters, and prove that the linear convergence result is also valid for the SGF. In the course of the convergence analysis,
we employ a general concentration inequality to deal with the quantities related to initialization.
For SGD, we analyze each iterate and control the parameters and the loss using the bound from each step. For SGF, we first bound the logarithm of the loss using properties of Ito integral
and then bound the parameters. 

The rest of the paper is organized as follows.
In Section \ref{StaRan}, we describe the problem setting, and
give some preliminary estimates. In Section \ref{SecSGD}, we provide
the convergence analysis of SGD. In Section \ref{SecSGF}, we investigate the continuous time model. In Section \ref{sec:num}, we present numerical results for the two-dimensional Poisson problem. In Section \ref{sec:concl}, we give concluding remarks and discuss future research problems. The proofs are given in the appendices. Throughout,
we use \(m\) to denote the width of the hidden layer of the  PINNs, and both the words 
`over-parameterized' and `wide' mean that \(m\) approaches infinity. Moreover, we also use \([m]\) 
to denote the set \(\{1,2,\ldots,m\}\). Given two quantities \(a\) and \(b\), the symbol 
\(a\gtrsim b\) means that there is an absolute constant \(C\), independent of the neural network width $m$, the numbers $n_1,n_2$ of sampling points and the probability tolerance $\delta$, such that \(a\geq C\cdot b\), and the symbols 
`\(\lesssim\)' and `\(\approx\)' have the similar meaning.

\section{Preliminary}\label{StaRan} 

In this section, we describe the problem formulation and give preliminary estimates. 
\subsection{Physics informed neural networks}
Let $\Omega\subset \mathbb{R}^d$ be an open bounded domain with a smooth boundary $\partial\Omega$.
Consider the Poisson equation with the Dirichlet boundary condition:
\begin{equation}\label{PoissonPDE}
	\left\{
	\begin{aligned}
		\Delta u &=f, \quad \mbox{in } \Omega, \\
		u&=g, \quad  \mbox{on }  \partial \Omega,
	\end{aligned}
	\right.
\end{equation}
where $\bm{x}=(x_1,x_2,\cdots,x_d)\in \mathbb{R}^d$, and $\Delta u = \sum_{i=1}^d 
\frac{\partial^2 u}{\partial x_i^2}$ is the Laplace operator.
The analysis below can also be extended to other linear PDEs and with boundary conditions of other types using the same proof strategy. However, the preliminary estimates in Appendix \ref{Pro-StaRan} and the analysis in Sections \ref{SecSGD} and \ref{SecSGF} are slightly more involved, and the implicit constants in the inequalities for \(m\) and \(\eta\) will also change accordingly; We provide a more detailed discussion about relevant changes in Appendix \ref{Gen-PDE}.

To approximate the PDE solution $u$ using a neural network \(\phi(\bm{x};\theta)\) (with the neural network parameters $\theta$), PINN constructs a physics-informed expected loss based on the principle of PDE residual minimization:
\begin{equation}
    \mathcal{L}(\bm{w},\bm{a}) = 
	\frac12\int_{\Omega} \left(\Delta [\phi(\bm{x};\theta)] - f(\bm{x}) \right)^2 {\rm d}x + \frac\gamma2 
	\int_{\partial\Omega} \left(\phi(\bm{y};\theta) - g(\bm{y}) \right)^2 {\rm d}y,
\end{equation}
where the hyperparameter \(\gamma>0\) balances the interior and boundary losses. This loss reflects how well the neural network function $\phi(\boldsymbol{x};\theta)$ satisfies the differential equation and the boundary condition, henceforth the name physics-informed neural network.

In this work, we employ the standard fully connected two-layer neural networks:
	\begin{align}\label{network}
	\phi(\bm{x}; \bm{w}, \bm{a}) = \frac{1}{\sqrt{m}} \sum_{r=1}^{m} a_r 
	\cdot \sigma \left( [{w}_{r1} \, {w}_{r2} \cdots {w}_{rd}] 
	\bm{x} +  {w}_{r,d+1} \right) 
	= \frac{1}{\sqrt{m}} \sum_{r=1}^{m} a_r \cdot \sigma \left( \bm{w}_r^\top 
	\wt{\bm{x}} \right),
	\end{align}
where $\sigma:\mathbb{R}\to\mathbb{R}$ is a nonlinear activation function and 
$\bm{w} = [\bm{w}_1^\top \, \cdots \, \bm{w}_m^\top]^\top \in \mathbb{R}^{m(d+1)}$, 
$\bm{a} = [a_1 \, \cdots \,a_m]^\top \in \mathbb{R}^m$, 
$\bm{w}_r = [{w}_{r1} \, {w}_{r2} \cdots {w}_{rd}\,{w}_{r,d+1}]^\top \in \mathbb{R}^{d+1}$,     
$ \wt{\bm{x}} = [\bm{x}^\top \, 1]^\top \in \mathbb{R}^{d+1}$.
The bias terms \({w}_{r,d+1}\) are absorbed into the vectors \(\bm{w}_r\) and 
\(\wt{\bm{x}}\in\mathbb{R}^{d+1}\) is the extended point of \(\bm{x}\). Let the vector \(\theta:=[\bm{w}^\top \, \bm{a}^\top]^\top\in \mathbb{R}^{m(d+2)}\) denote
all the parameters in the neural network. Throughout, we make the following assumption about the initialization and the activation function $\sigma$.

\begin{assumption}\label{AssAct} 

\begin{itemize}
    \item[(i)] For all \(r\in[m]\), \(\bm{w}_r(0)\) is sampled from the standard Gaussian distribution and 
    \(a_r(0)\) is sampled from Rademacher distribution, i.e., $\bm{w}_r(0) \sim \mathcal{N}(\bm{0}, \bm{I}_{d+1})$ and $a_r(0) \sim \mathrm{Unif}\{-1, 1\}$.

    \item[(ii)] The activation function \(\sigma\) is piecewise continuously differentiable up to 
	third order, and has locally Lipschitz property, 
	i.e., for all \( |z_1|,|z_2|\leq M\), there exists \(C_M>0\) such that 
	\[  | \sigma^{(k)}(z_1) - \sigma^{(k)}(z_2) |\leq C_M|z_1 - z_2|, \quad  k=0,1,2,3. \]
	The constant \(C_M\) grows at most polynomially with  \(M\):
	there exists \(A>0\) and \(\ell\geq 0\) such that
	\[ C_M \leq A(1+M^\ell), \quad \forall M>0. \]
	For all \(M>0\), the Lipschitz constant \(C_M\geq 1\) and satisfies 
	\(|\sigma^{(k)}(0)|\leq C_M\) for \(k=0,1,2,3\). Then we have for all \(|z|\leq M\)
	\[ |\sigma^{(k)}(z)|\leq |\sigma^{(k)}(z) - \sigma^{(k)}(0)| + |\sigma^{(k)}(0)| \leq C_M(|z|+1). \]
\end{itemize}
\end{assumption}

\begin{remark}
For the initialization of \(a_r(0)\), we only use the property of zero mean and boundedness, and one can also use  other initialization schemes. The assumption on $\sigma$ is quite generous and covers many commonly used functions. Indeed, the RePU function with power not lower than three and some globally Lipschitz smooth activations, e.g., \(\tanh\), \(\sin\), sigmoid and softplus, all satisfy the assumption. Higher growth order of \(C_M\) will induce larger \(m\) for the convergence guarantee; see Remark \ref{rmk:m-dependence}. Indeed, those smooth functions which are widely used in neural PDE solvers have globally bounded derivatives (for any fixed order). In comparison, the works \cite{Gao2023,XuDuHuang:2024} investigates PINN with ReLU$^3$ activation, whereas the work \cite{Xu2024} focuses on nonpolynomial analytic and three-times globally Lipschitz activations. 
\end{remark}

In practice, the population loss $\mathcal{L}(\bm{w},\bm{a})$ has to be further discretized since the involved integrals cannot be evaluated in closed form. This is commonly achieved using Monte Carlo methods, which is especially attractive in the high-dimensional case. Specifically, let the training samples $\{\bm{x}_p\}_{p=1}^{n_1}$ and $\{{\bm{y}}_q\}_{q=1}^{n_2}$ be drawn independently and identically distributed (i.i.d.) from the uniform distributions $\mathcal{U}(\Omega)$ and $\mathcal{U}(\partial\Omega)$ on the domain $\Omega$ and the boundary $\partial\Omega$, respectively. Then the empirical loss $L(\bm{w},\bm{a})$ reads
\begin{equation*}
		L(\bm{w},\bm{a}) :=
		\sum_{p=1}^{n_1}  \frac{|\Omega|}{2n_1}\left( \sum_{i=1}^d \frac{\partial^2 \phi}{\partial x_i^2}
		(\bm{x}_p;\bm{w},\bm{a})-f(\bm{x}_p)\right)^2
		 + \sum_{q=1}^{n_2}\frac{\gamma |\partial\Omega|}{2n_2} 
		\left(\phi(\bm{y}_q;\bm{w},\bm{a})-g(\bm{y}_q)\right)^2,
\end{equation*}
where $|\Omega|$ and $|\partial \Omega|$ denote their respective Lebesgue measures.
To simplify the notation, we absorb $|\Omega|$ and $|\partial\Omega|$ into $\gamma$, and investigate the empirical loss
\begin{equation}\label{PINN Loss}
		L(\bm{w},\bm{a}) :=
		\sum_{p=1}^{n_1}  \frac{1}{2n_1}\left( \sum_{i=1}^d \frac{\partial^2 \phi}{\partial x_i^2}
		(\bm{x}_p;\bm{w},\bm{a})-f(\bm{x}_p)\right)^2
		 + \sum_{q=1}^{n_2}\frac{\gamma }{2n_2} 
		\left(\phi(\bm{y}_q;\bm{w},\bm{a})-g(\bm{y}_q)\right)^2.
\end{equation}
The gradient of the loss $L(\boldsymbol{w},\boldsymbol{a})$ can be evaluated efficiently using automatic differentiation \cite{Baydin:2018AD}, and thus standard first-order algorithms such as SGD can be readily implemented.

Below we assume the samples \(\|\bm{x}_p\|_2,\|{\bm{y}}_q\|_2\leq 1\),
so that the extended samples satisfy \(\|\wt{\bm{x}}_p\|_2,\|\wt{\bm{y}}_q\|_2\leq \sqrt{2}\). 
Throughout, we use the following notation. We denote the loss in the domain $\Omega$ and on the boundary $\partial\Omega$ by $s_p(\bm{w},\bm{a})$ and $h_q(\bm{w},\bm{a})$, respectively:
\begin{gather*}
	s_p(\bm{w},\bm{a})=\frac{1}{\sqrt{n_1}} \left( \sum_{i=1}^d \frac{\partial^2 \phi}{\partial x_i^2}
	(\bm{x}_p;\bm{w},\bm{a})-f(\bm{x}_p)\right)\quad \mbox{and}\quad	h_q(\bm{w},\bm{a})=\sqrt{\frac{\gamma}{n_2}}(\phi(\bm{y}_q;\bm{w},\bm{a})-g(\bm{y}_q)),
\end{gather*}
and accordingly the following loss vectors
\begin{gather*}
	\bm{s}(\bm{w},\bm{a})=[s_1(\bm{w},\bm{a}) \cdots s_{n_1}(\bm{w},\bm{a})]^\top\in \mathbb{R}^{n_1}\quad \mbox{and}\quad	
	\bm{h}(\bm{w},\bm{a})=[h_1(\bm{w},\bm{a}) \cdots h_{n_2}(\bm{w},\bm{a})]^\top \in \mathbb{R}^{n_2}.
\end{gather*}
Then the empirical loss $L$ can be written as
\begin{equation}
	L(\bm{w},\bm{a})=\tfrac{1}{2}(\|\bm{s}(\bm{w},\bm{a})\|_2^2+ \|\bm{h}(\bm{w},\bm{a})\|_2^2).
\end{equation}
The notation \(t\) denotes
the iterate index (for SGD) or time (for SGF). Moreover, we use the following notation interchangeably:
$$\theta(t) = \theta_t  = (\bm{w}_t, \bm{a}_t ) = (\bm{w}(t), \bm{a}(t))\quad \mbox{and}\quad
L(t) = L(\theta_t) = L(\bm{w}_t, \bm{a}_t ).$$
Moreover, when dealing with the randomness of the initialization,
we may abuse the notation \(\theta:=[\bm{w}^\top \, \bm{a}^\top]^\top\) and 
\( \theta(0):=[\bm{w}(0)^\top \, \bm{a}(0)^\top]^\top \). 

\subsection{The idea of convergence analysis}

We first use gradient flow to illustrate the idea of convergence analysis.
In gradient flow, the parameters satisfy the following ordinary differential equations (ODEs):
\begin{gather*}
	\begin{aligned}
		\frac{{\rm d}\bm{w}(t)}{{\rm d}t} = -\frac{\partial L(t)}{\partial \bm{w}}
		= -\sum_{p=1}^{n_1} s_p(t) \cdot 
		\frac{\partial s_p(t)}{\partial \bm{w}} - 
		\sum_{q=1}^{n_2} h_q(t) \cdot 
		\frac{\partial h_q(t)}{\partial \bm{w}},
	\end{aligned}\\
	\begin{aligned}
		\frac{{\rm d}\bm{a}(t)}{{\rm d}t} = -\frac{\partial L(t)}{\partial \bm{a}} 
		= -\sum_{p=1}^{n_1} s_p(t) \cdot 
		\frac{\partial s_p(t)}{\partial \bm{a}} - 
		\sum_{q=1}^{n_2} h_q(t) \cdot 
		\frac{\partial h_q(t)}{\partial \bm{a}}.
	\end{aligned}
\end{gather*}
By the chain rule, the domain loss \(s_p(t)\) satisfies
\begin{equation*}
	\frac{\d s_p(t)}{\d t} =  \left\langle 
		\frac{\partial s_p(t)}{\partial \bm{w}}, 
		\frac{\d\bm{w}(t)}{\d t} \right\rangle +  
		\left\langle \frac{\partial s_p(t)}{\partial \bm{a}},
		\frac{\d\bm{a}(t)}{\d t} \right\rangle
		= \left\langle \frac{\partial s_p(t)}{\partial \theta},
		\frac{\d\theta(t)}{\d t} \right\rangle.
\end{equation*}
Consequently, the loss vector satisfies
\begin{equation*}
	\frac{\d}{\d t}  \begin{bmatrix} \bm{s}(t) \\ \bm{h}(t) \end{bmatrix}
	= -\left( \bm{G}_{\bm{w}}(t) + 
	\bm{G}_{\bm{a}}(t) \right)
	\begin{bmatrix} \bm{s}(t) \\ \bm{h}(t) \end{bmatrix},
\end{equation*}
with 
\begin{equation}
	\bm{G}_{\bm{w}}(\bm{w}, \bm{a}) = 
	\bm{D}_{\bm{w}}^\top \bm{D}_{\bm{w}}, \quad
	\bm{D}_{\bm{w}} =  \begin{bmatrix}
		\frac{\partial s_1(\bm{w}, \bm{a})}{\partial \bm{w}} 
		& \cdots 
		& \frac{\partial s_{n_1}(\bm{w}, \bm{a})}{\partial \bm{w}} 
		& \frac{\partial h_1(\bm{w}, \bm{a})}{\partial \bm{w}} 
		& \cdots 
		& \frac{\partial h_{n_2}(\bm{w}, \bm{a})}{\partial \bm{w}}
		\end{bmatrix},
\end{equation}
\begin{equation}\label{exp-Ga} 
	\bm{G}_{\bm{a}}(\bm{w}, \bm{a}) = 
	\bm{D}_{\bm{a}}^\top \bm{D}_{\bm{a}}, \quad
	\bm{D}_{\bm{a}} = \begin{bmatrix}
		\frac{\partial s_1(\bm{w}, \bm{a})}{\partial \bm{a}} 
		& \cdots 
		& \frac{\partial s_{n_1}(\bm{w}, \bm{a})}{\partial \bm{a}} 
		& \frac{\partial h_1(\bm{w}, \bm{a})}{\partial \bm{a}} 
		& \cdots 
		& \frac{\partial h_{n_2}(\bm{w}, \bm{a})}{\partial \bm{a}}
		\end{bmatrix}.
\end{equation}
Now, if the smallest eigenvalues of the two
Gram matrices \(\bm{G}_{\bm{w}}(t)\) and \(\bm{G}_{\bm{a}}(t)\) have 
positive lower bounds during the training process, then the loss value $L(\theta_t)$ tends to \(0\) when
\(t\) tends to infinity. This can be achieved if the `infinite' Gram matrices
are positive definite and the training is in the `lazy training' regime \cite{Chizat2019}. For the former condition, we make the following assumption, where $\mathbb{E}_{\theta(0)}$ denotes taking expectation with respect to the distribution of the initialization $\theta(0)$. These quantities essentially govern the training dynamics. We refer to readers to Appendix \ref{PosiGram} for the positivity result of the Gram matrix with the RePU or smooth activation functions.

\begin{assumption}\label{InfGram} 
	Let \( \bm{G}_{\bm{w}}^{\infty}\) and 
	\(\bm{G}_{\bm{a}}^{\infty} \in \R^{(n_1 + n_2)\times (n_1 + n_2)} \)
	be the expectation of the initial Gram matrices, i.e.,
	\begin{equation}\label{exp-G_ify}
		\bm{G}_{\bm{w}}^{\infty} := \E_{\theta(0)}\left[ \bm{G}_{\bm{w}}(\theta(0)) \right]\quad \mbox{and} \quad
		\bm{G}_{\bm{a}}^{\infty} := \E_{\theta(0)}\left[ \bm{G}_{\bm{a}}(\theta(0)) \right].
	\end{equation}
	The infinite Gram matrices \(\bm{G}_{\bm{w}}^{\infty}\) and
	\(\bm{G}_{\bm{a}}^{\infty}\) are positive definite: 
	\begin{equation}
		\lambda_{\bm{w}} := \lambda_{\min}(\bm{G}_{\bm{w}}^{\infty})>0\quad \mbox{and} \quad
		\lambda_{\bm{a}} := \lambda_{\min}(\bm{G}_{\bm{a}}^{\infty})>0.
	\end{equation}
\end{assumption}

\begin{remark}
Since the losses \(s_p(\theta)\) and \(h_q(\theta)\) depend on the value of the neural network \(\phi(\bm{x}; \bm{w}, \bm{a})\) at the sampling points,	the Gram matrices \( \bm{G}_{\bm{w}}^{\infty}\) and \(\bm{G}_{\bm{a}}^{\infty}\) 
	are related to the neural network and sampling points, while their sizes depend on
	the number of samples. However, they are independent of \(m\) even though 
	\(\phi(\bm{x}; \bm{w}, \bm{a})\) depends on \(m\). Indeed, for \(1\leq j,k,\leq n_1\), the analysis in Appendix \ref{Pro-StaRan} gives
	\begin{align*}
		\llan \frac{\partial s_j(\theta)}{\partial \bm{w}}, 
		\frac{\partial s_k(\theta)}{\partial \bm{w}} \rran
		 =  \frac{1}{m} \sum_{r=1}^{m} \llan \bm{v}(\bm{w}_r,a_r;\bm{x}_j), 
		 \bm{v}(\bm{w}_r,a_r;\bm{x}_k) \rran
		 =: \frac{1}{m} \sum_{r=1}^{m} V(\bm{w}_r,a_r;\bm{x}_j,\bm{x}_k),
	\end{align*}
	with
	\[ \bm{v}(\bm{w}_r,a_r;\bm{x}_j) = \frac{1}{\sqrt{n_1}}\sum_{i=1}^{d} 
	\left( a_r \sigma'''(\bm{w}_r^\top \wt{\bm{x}}_j)w_{ri}^2 \wt{\bm{x}}_j 
	+ 2a_r \sigma''(\bm{w}_r^\top \wt{\bm{x}}_j)w_{ri}\bm{e}_i \right).  \]
	The \((j,k)\)-th entry $\bm{G}_{\bm{w}}^{\infty}[j,k]$ of the matrix \( \bm{G}_{\bm{w}}^{\infty}\) is given by
	\begin{align*}
		\bm{G}_{\bm{w}}^{\infty}[j,k] = \E_\theta\left[ \frac{1}{m} 
		\sum_{r=1}^{m} V(\bm{w}_r,a_r;\bm{x}_j,\bm{x}_k) \right]
		= \frac{1}{m} \sum_{r=1}^{m} \E_{\bm{w}_r,a_r}\left[ V(\bm{w}_r,a_r;\bm{x}_j,\bm{x}_k) \right].
	\end{align*}
Since all the entries \((\bm{w}_r,a_r)\) are i.i.d., the summation reduces to 
	\(\E_{\bm{w}_r,a_r}\left[ V(\bm{w}_r,a_r;\bm{x}_j,\bm{x}_k)\right]\).
Thus the smallest eigenvalues \(\lambda_{\bm{w}}\) and \(\lambda_{\bm{a}}\) of 
(\( \bm{G}_{\bm{w}}^{\infty} \) and \(\bm{G}_{\bm{a} }^{\infty}\)) are fixed once the sampling points 
are chosen. 
\end{remark}

\begin{remark}\label{rmk:GramInit}
For the NTK of regression problems (i.e., without differential operators), when the activation \(\sigma\) is non-polynomial piecewise differentiable and the sampling points $\{\bm{x}_p\}_{p=1}^{n_1}$ are pairwise non-parallel, the positivity of the infinite Gram matrix is guaranteed; see the recent work \cite{Carvalho:2025} and the references therein for detailed discussions.
The verification of Assumption \ref{InfGram} for PINNs is generally highly nontrivial due to the presence of multiple differential operators. So far the positivity has been established for some specific PDE problems and activation functions \cite{Gao2023,Xu2024,ZhaoLuo:2025}; see Appendix \ref{PosiGram} for these important results as well as detailed discussions about adapting the positivity result to the Poisson problem \eqref{PoissonPDE} with the RePU or smooth activation functions.
\end{remark}

Below we also use the notation 
\begin{equation} \label{eqn:lambda-G}
\lambda_\theta = \lambda_{\bm{w}} + \lambda_{\bm{a}}\quad \mbox{and}\quad 
	\bm{G}_{\theta}(\bm{w}, \bm{a}) = \bm{G}_{\bm{a}}(\bm{w}, \bm{a}) + \bm{G}_{\bm{w}}(\bm{w}, \bm{a})
	= \bm{D}_\theta^\top \bm{D}_\theta,
\end{equation}
with
\begin{equation*}
	\bm{D}_\theta = \begin{bmatrix}
		\frac{\partial s_1(\bm{w}, \bm{a})}{\partial \theta} 
		& \cdots 
		& \frac{\partial s_{n_1}(\bm{w}, \bm{a})}{\partial \theta} 
		& \frac{\partial h_1(\bm{w}, \bm{a})}{\partial \theta} 
		& \cdots 
		& \frac{\partial h_{n_2}(\bm{w}, \bm{a})}{\partial \theta}
		\end{bmatrix}.
\end{equation*}
Note that by Weyl's inequality, we have 
\[\lambda_{\min}(\bm{G}_{\theta}^{\infty}) =
\lambda_{\min}(\bm{G}_{\bm{w}}^{\infty}+\bm{G}_{\bm{a}}^{\infty}) 
\geq \lambda_{\bm{w}} + \lambda_{\bm{a}} = \lambda_\theta. \]

\begin{remark}
Tight bounds on the smallest eigenvalue $\lambda_\theta$ of the Gram matrix in the case of regression (i.e., in the abscence of differential operators) using deep ReLU nets can be found in \cite{Nguyen2021}. In contrast, the spectral analysis of eigenvalues of the Gram matrix for PINNs remain largely completely open due to its intrinsic delicacy. The smallest eigenvalue $\lambda_\theta$ depends on several factors, e.g., the activation function $\sigma$ and the distribution of the sampling points $\{\bm{x}_i\}_{i=1}^{n_1}$ and $\{\bm{y}_j\}_{j=1}^{n_2}$. As a general rule of thumb, the smoother is the activation function $\sigma$, the faster is the decay of the eigenvalues of the Gram matrices. Moreover, badly distributed sampling points (e.g., with two points nearly identical) can lead to severe ill-conditioning of the Gram matrices. However, currently a precise characterization of $\lambda_\theta$ is still unavailable for the NTK of PINNs in terms of these important influencing factors. 
\end{remark}

For stochastic gradient-type algorithms, the parameters $\theta$ are updated as 
\begin{align}
    \mathrm{SGD}: & \quad \theta(t+1) = \theta(t) - \eta \frac{\partial \wt{L}(\theta_t)}{\partial \theta},\label{eqn:SGD-0} \\
    \mathrm{SGF}: & \quad \d\theta_t = - \nabla L(\theta_t)\d t + \sqrt{\eta} \sigma(\theta_t) \d W_t,\label{eqn:SGF-0}
\end{align}
with \(\eta\) being the fixed learning rate and \(\wt{L}(\theta)\) the mini-batch loss which is constructed by subsampling the given set of sampling points $\{\bm{x}_i\}_{i=1}^{n_1}$ and $\{\bm{y}_j\}_{j=1}^{n_2}$ uniformly at random; see Sections \ref{SecSGD} and \ref{SecSGF} for further details of the schemes (including a short discussion of the impact of the batch size).
In this work, we establish the following result on the convergence of the SGD and SGF. 
\begin{theorem}[informal statements]\label{thm:informal}
Let Assumptions \ref{AssAct} and \ref{InfGram} hold, and use SGD \eqref{eqn:SGD-0} / SGF \eqref{eqn:SGF-0} to train the empirical loss $L(\boldsymbol{w,a})$. 
Let \(S\) be the first time that some parameter leaves a given neighborhood of the initialization.
If the width $m$ of the neural network is large enough and the step size \(\eta\) is small enough, then with high probability {\color{blue} \(S=\infty\)}, and the expected value of the loss \(L(t)\) decays exponentially
\begin{align*}
	\mathrm{SGD}: &\quad
		\E \left[ L(t) \cdot \mathbf{1}_{S=\infty} \right] 
			\leq \left( 1- \eta \tfrac{\lambda_\theta}{2} \right)^t L(0),\\
\mathrm{SGF}: &\quad
			\E \left[ L(t) \cdot \mathbf{1}_{S=\infty} \right] 
			\leq \exp\left( - \tfrac{\lambda_\theta}{2}t \right) L(0).
		\end{align*}
\end{theorem}

Theorem \ref{thm:informal} shows that among all possible iteration trajectories of the neural network parameters, the majority of the trajectories stay in the vicinity of its starting point. Moreover, the average loss of such trajectories decays exponentially to zero.

\subsection{Static randomness}

First we deal with the quantities related to random initialization, the so-called static randomness
compared with the dynamic randomness of the algorithm, which arises from the random selection of the data at each iteration. The proofs can be found in Appendix \ref{Pro-StaRan}.

\begin{lemma}\label{wInit} 
	For any \(\delta\in (0,1)\), with probability at least \(1-\delta\), there holds
	\[ \|\boldsymbol{w}_r(0)\|_2 \leq \sqrt{2(d+1)\log\left(\frac{2m(d+1)}{\delta}\right)},
	\quad \forall r\in [m]. \] 
\end{lemma}

\begin{remark}
Gaussian initialization is crucial to ensure the positivity of infinite Gram matrix (and then the initial Gram matrix) \cite{Carvalho:2025,Gao2023,Xu2024,ZhaoLuo:2025}, which is central to the convergence analysis. In practice, one may also employ uniformly bounded initializations, in which case the bound of the initial parameters holds automatically. Note that existing analysis of the positivity of the associated infinite Gram matrix depends crucially on the full support of the distribution, and the analysis strategy relies crucially on reducing the positivity issue to the linear independence of certain functions (determined by the activation $\sigma$ and sampling points) over the entire Euclidean space; see \eqref{inner-product_G} and \eqref{funs-LI} in Appendix \ref{PosiGram} for details. For bounded initializations, one would need to establish the linear independence over a bounded domain instead, which may impose additional restrictions on the activation $\sigma$ and sampling points. It is of much interest to investigate the impact of the initialization scheme on the Gram matrix, including its positivity and spectral behavior. 
\end{remark}

Given any \(M>0\), let \(E(M)\) be the event that \( \|\bm{w}_r(0)\|_2 \leq M \) 
for all \(r\in [m]\), namely,
\[ E(M):= \bigcap_{r=1}^{m} \left\{ \|\bm{w}_r(0)\|_2 \leq M  \right\}. \]
 Define the quantity
\[B:= 1 + \sqrt{2(d+1)\log\left(\frac{2m(d+1)}{\delta}\right)}. \] 
Then the event in Lemma \ref{wInit} is \(E(B-1)\). 
This lemma indicates that with high probability, all the \(\bm{w}_r(0)\) are bounded by \(B-1\).
We conduct the analysis within the event $E(B-1)$ below.
The choice of \(B\) is to ensure that all the weights 
\(\|\bm{w}_r(t)\|_2\) are bounded by \(B\) during the iteration, which can be realized by the initial bound 
\(B-1\) in the `lazy training' regime. Note that the quantity \(B\) and thus the Lipschitz constant \(C_B\) are related to the width $m$ of the neural network, which both grow with $\log m$ polynomially and can be dominated by the term linear in \(m\).

Next we analyze the initial loss $L(0)$, Gram matrices and the continuity of Gram matrices with respect to
the NN parameters $\theta$. The main tool in the analysis is concentration inequalities. 
\begin{lemma}\label{Loss0}
	If \(m\gtrsim \log^2\left(\frac{n_1+n_2}{\delta}\right)\), then conditioned on the event \(E(B-1)\), 
	we have with probability at least $1 - \delta$,
	\begin{align*}
		L(0) \lesssim C_B^2 d^3\log\left( \frac{n_1+n_2}{\delta} \right).
	\end{align*}
\end{lemma}

\begin{lemma}\label{GramInit} 
	If \(m\gtrsim \frac{C_B^4d^6}
	{\min\left\{\lambda_{\bm{w}}^2,\lambda_{\bm{a}}^2\right\}} 
	\log^3\left(\frac{n_1+n_2}{\delta}\right)\), then conditioned on the event \(E(B-1)\), 
	we have with probability at least $1 - \delta$,
	\begin{gather*}
		\|\bm{G}_{\bm{w}}(0) - \bm{G}_{\bm{w}}^\infty\|_2 \leq \frac{\lambda_{\bm{w}}}{4}\quad
	\text{and} \quad \|\bm{G}_{\bm{a}}(0) - \bm{G}_{\bm{a}}^\infty\|_2 \leq \frac{\lambda_{\bm{a}}}{4}.
	\end{gather*}
\end{lemma}

Lemma \ref{GramInit} shows that the difference between initial and infinite
Gram matrices is small for wide NNs, which ensures the positive definiteness of the
initial Gram matrices. This agrees with the result in \cite{Jacot2018}, which asserts that
the initial Gram matrices are deterministic and positive when \(m\) tends to infinity
and the loss is convex in the function space. Since the differential 
operator in PDE (\ref{PoissonPDE}) is linear,  the loss \(L(\theta)\) is still 
convex with respect to the neural network \(\phi\).
The next lemma shows the continuity of the Gram matrices, i.e., $\boldsymbol{G_w}$ and $\boldsymbol{G_a}$ are stable under small perturbations. These two results are the key points for the positive definiteness of Gram matrices during training.


\begin{lemma} \label{GramDyn} 
    If \(m\gtrsim\log^3 \left(\frac{1}{\delta}\right)\) and the constants \(0< R_{\bm{w}},R_{\bm{a}}\leq 1 \) satisfy
	\begin{equation}\label{RwRa}
        R_{\bm{w}}, R_{\bm{a}} \approx 
 		\frac{\min\left\{\lambda_{\bm{w}},\lambda_{\bm{a}}\right\}}{C_B^2 d^3},
	\end{equation}
    then conditioned on the \(E(B-1)\) and the event in Lemma \ref{GramInit}, we have with probability at least $1 - \delta$, the following statement holds: Whenever the parameters \(\bm{w}_r, a_r, r\in[m]\) satisfy
    \[\|\bm{w}_r - \bm{w}_r(0)\|_2 \leq R_{\bm{w}},\quad \|a_r - a_r(0)\|_2 \leq R_{\bm{a}},\quad \forall r\in [m], \]
    then $\lambda_{\min}(\bm{G}_{\bm{w}}(\bm{w},\bm{a})) \geq \frac{1}{2}\lambda_{\bm{w}}$ and  $ \lambda_{\min}(\bm{G}_{\bm{a}}(\bm{w},\bm{a})) \geq \frac{1}{2}\lambda_{\bm{a}}.$
\end{lemma}

\begin{remark}
Lemma \ref{GramDyn} indicates that in the NTK regime, if each parameter changes only slightly then the training dynamics can be quantitatively approximated by a related kernel. 
In fact, from the proof of Lemma \ref{w(S)-w(0)}, a larger \(m\) makes each \(\bm{w}_r\) and \(a_r\) stays closer from their initialization for all \(r\in[m]\), within an \(\mathcal{O}(\frac{1}{\sqrt{m}})\) neighborhood, when ignoring the terms depending logarithmically on \(m\).
However, the collective variation can be significant in wide neural networks when all the parameters are trained. This paradigm shares the spirit of freezing the parameters in the hidden layer, which is known as extreme learning machine (ELM) in the literature \cite{Huang:2006ELM}. The key difference of the NTK from the ELM lies in the fact that ELM is a linear model when the training parameters and the associated optimization problem is convex, whereas the former is nonlinear with respect to the parameters and the associated optimization problem is often nonconvex. Indeed, one key contribution of the work is to rigorously prove the guarantee that the parameters change only slightly with high probability when training wide PINNs by SGM, i.e., such models operate within the lazy training regime; see Theorems \ref{SGDThm} and \ref{ConThm}.
\end{remark}

Now we summarize the consequences of the preceding lemmas before analyzing the stochastic dynamics. 
We denote by \(E(L(0)),E(\bm{G}(0))\) and \(E(\bm{G})\) the events in Lemmas \ref{Loss0}, \ref{GramInit}  and \ref{GramDyn}, respectively, and \(\init\) the intersection of all these three events, i.e.,
\begin{equation}\label{initialization}
    \init = E(B-1) \cap E(L(0)) \cap E(\bm{G}(0)) \cap E(\bm{G}).
\end{equation}
Note that usually we have \(\min\left\{\lambda_{\bm{w}}^2,\lambda_{\bm{a}}^2\right\}\leq 1\). Thus if \(m\gtrsim \frac{C_B^4d^6}
	{\min\left\{\lambda_{\bm{w}}^2,\lambda_{\bm{a}}^2\right\}} 
	\log^3\left(\frac{n_1+n_2}{\delta}\right) \geq \log^3\left(\frac{n_1+n_2}{\delta}\right)\), we have
\begin{align*}
    \P(\init) & = \P(E(B-1))\P(E(L(0))\cap E(\bm{G}(0)) \cap E(\bm{G}) |E(B-1)) \\
    & \geq \P(E(B-1)) \Big( \P(E(L(0))|E(B-1)) + \P( E(\bm{G}(0)) \cap E(\bm{G})|E(B-1)) - 1\Big) \\
    & = \P(E(B-1))(\P(E(L(0))|E(B-1))-1) + \P(E(B-1)\cap E(\bm{G}(0)))\P( E(\bm{G}) |  E(B-1)\cap E(\bm{G}(0)) ) \\
    &\geq -\delta + (1-\delta)^2(1-\delta) \geq 1-4\delta,
\end{align*}
where we have used the elementary inequality \(\P(A\cap B| E) \geq \P(A| E) + \P(B| E) - 1 \) for any events $A,B$ and $E$. Therefore, 
with probability at least \(1-4\delta\) over initialization, the event \(\init\) happens. 
With the occurrence of \(\init\), the initial Gram matrices are positive definite and Lemma \ref{GramDyn} ensures the positivity of Gram matrices during training if the parameters lie in the vicinity of their initializations.

\section{Convergence Analysis of SGD}\label{SecSGD}

In this part, we consider any unbiased SGD, i.e., the parameters are updated using a mini-batch loss constructed from sampling points randomly subsampled from the fixed sets $\{\bm{x}_p\}_{p=1}^{n_1}$ and $\{{\bm{y}}_q\}_{q=1}^{n_2}$:
\begin{equation}
	\begin{aligned}
		\wt{L}(\bm{w},\bm{a}) 
		& = \sum_{p\in \mathcal{I}} \frac{1}{2I}\left( \sum_{i=1}^d \frac{\partial^2 \phi}{\partial x_i^2}
		(\bm{x}_p;\bm{w},\bm{a})-f(\bm{x}_p)\right)^2
		 + \sum_{q\in \mathcal{J}}\frac{\gamma }{2J} 
		\left(\phi(\bm{y}_q;\bm{w},\bm{a})-g(\bm{y}_q)\right)^2 \\
		& = \sum_{p\in \mathcal{I}} \frac{n_1}{2I} s_p(\bm{w},\bm{a})^2  + 
		\sum_{q\in \mathcal{J}} \frac{n_2}{2J} h_q(\bm{w},\bm{a})^2,
	\end{aligned}
\end{equation}
where the indices \(\mathcal{I}\subset [n_1]\) with \(I=|\mathcal{I}|\) and 
\(\mathcal{J}\subset [n_2]\) with \(J=|\mathcal{J}|\), and the choices of the subsets 
\(\mathcal{I}\) and \(\mathcal{J}\) are independent. The specific schemes for \(\mathcal{I}\) and \(\mathcal{J}\) can take different forms, e.g., mini-batch  with replacement. Then the SGD iteration is given by
\begin{gather}
	\bm{w}_r(t+1) = \bm{w}_r(t) - \eta 
	\frac{\partial \wt{L}(\bm{w}(t), \bm{a}(t))}{\partial \bm{w}_r}, \\
	a_r(t+1) = a_r(t) - \eta
	\frac{\partial \wt{L}(\bm{w}(t), \bm{a}(t))}{\partial a_r}.
\end{gather}
Since the considered scheme is unbiased, we have
\begin{align*}
		\E_{\II,\JJ}\left[ \wt{L}(\bm{w},\bm{a}) \right] 
		 &= \E_{\II}\left[ \sum_{p\in \mathcal{I}} \frac{n_1}{2I} s_p(\bm{w},\bm{a})^2 \right] + 
		\E_{\JJ}\left[ \sum_{q\in \mathcal{J}} \frac{n_2}{2J} h_q(\bm{w},\bm{a})^2 \right]
		= L(\bm{w},\bm{a}),\\
		\E_{\II,\JJ}\left[ \frac{\partial \wt{L}(\bm{w}, \bm{a})}{\partial \bm{w}_r} \right]  
		 &=  \E_{\II,\JJ}\left[ \sum_{p\in \mathcal{I}} \frac{n_1}{I} s_p(\bm{w}, \bm{a}) \cdot 
		\frac{\partial s_p(\bm{w}, \bm{a})}{\partial \bm{w}_r} +
		 \sum_{q\in \mathcal{J}} \frac{n_2}{J} h_q(\bm{w}, \bm{a}) \cdot 
		\frac{\partial h_q(\bm{w}, \bm{a})}{\partial \bm{w}_r} \right] 
		 = \frac{\partial L(\bm{w}, \bm{a})}{\partial \bm{w}_r},\\
		\E_{\II,\JJ}\left[ \frac{\partial \wt{L}(\bm{w}, \bm{a})}{\partial a_r} \right]  
		 &=  \E_{\II,\JJ}\left[ \sum_{p\in \mathcal{I}} \frac{n_1}{I} s_p(\bm{w}, \bm{a}) \cdot 
		\frac{\partial s_p(\bm{w}, \bm{a})}{\partial a_r} +
		 \sum_{q\in \mathcal{J}} \frac{n_2}{J} h_q(\bm{w}, \bm{a}) \cdot 
		\frac{\partial h_q(\bm{w}, \bm{a})}{\partial a_r} \right] 
		 = \frac{\partial L(\bm{w}, \bm{a})}{\partial a_r}.
\end{align*}

\begin{remark}\label{rmk:variance}
The discussed SGD algorithms have the following setting \cite{An2023,Liu:2022almost,Weissmann:2025almost}: at each iteration 
\( \theta_{t+1} = \theta_t - \eta(\nabla_\theta L(\theta_t) + Z(\theta_t)), \)
with \(Z(\theta_t)\) an unbiased random noise of the gradient \(\nabla_\theta L(\theta_t)\). Our analysis focuses exclusively on the first-order moment of the random noise, i.e., convergence in expectation. All the stated results hold independently of the batch sizes $I$ and $J$, e.g., the change of parameters and the decay of loss after one iteration in Lemmas \ref{StepGap-w_t} and \ref{LossGap} below, while lacking a detailed study of the variance. Hence the final result is valid only in expectation sense and regardless of the specific sampling method. This however obscures the effect of the batch sizes. In the analysis, 
we employ a strategic decomposition in deriving the key estimate \eqref{eli-BatSize}, which completely eliminates the effect of the batch size, see Remark \ref{rmk:batch_size} for more details, and the result is consistent with the unbiased property of the iteration. 
Although our expectation estimates hold for any unbiased SGD, the variance depends heavily on the sampling schemes. For example, if we fix \(I=|\mathcal{I}|\) and \(J=|\mathcal{J}|\) and choose \(\mathcal{I}\) from \([n_1]\) and \(\mathcal{J}\) from \([n_2]\) uniformly at random, then by the independence of the random index sets \(\mathcal{I}\) and \(\mathcal{J}\), the variance of the stochastic estimator \(\wt{L}(\bm{w},\bm{a})\) is given by 
\begin{align*}
  \mathrm{Var}(\wt{L}) =& \E_{\II,\JJ}\left[ (\wt{L} - L)^2 \right] \\
   =& \frac{n_1-I}{4I(n_1-1)}\left( n_1\sum_{p=1}^{n_1}s_p^4 - \left(\sum_{p=1}^{n_1}s_p^2\right)^2 \right)
        + \frac{n_2-J}{4J(n_2-1)}\left( n_2\sum_{q=1}^{n_2}h_q^4 - \left(\sum_{q=1}^{n_2}h_q^2\right)^2 \right).
    \end{align*}
Thus small \(I\) and \(J\) result in larger variance in the loss estimator $\widetilde{L}(\bm{w},\bm{a})$, while the choice \(I=n_1\) and \(J=n_2\) has a vanishing variance. The latter case recovers the empirical loss $L(\bm{w},\bm{a})$, and the resulting algorithm is  deterministic (i.e., gradient descent). To obtain an explicit dependence of the convergence results on the batch sizes $I$ and $J$, one requires more refined estimates on the variances at each iteration in Lemmas \ref{StepGap-w_t} and \ref{LossGap}, and then integrate their cumulative impact along the trajectories to arrive at the final result in Theorem \ref{SGDThm}, which remains very challenging. Such an analysis is beyond the scope of this work.
\end{remark}

\begin{remark}
Our analysis covers random sampling with replacement. The mini-batch scheme without replacement is also commonly used in practice but the estimated stochastic gradient is biased, and thus the analysis herein does not apply to the case. In practice, it is also common to sample new i.i.d. points on-the-fly from the underlying probability measure on the domain / boundary at each iteration \cite{Shin:2024theoretical}. Despite its unbiasedness, it may generate infinitely many points and the associated Gram matrices will be infinite-dimensional (i.e., operators) and does not satisfy the positivity, due to the continuity of the associated kernel (e.g., for any smooth activation $\sigma$). Since the analysis strategy in the NTK framework relies crucially on the positivity of the associated Gram matrix, it does not apply directly to the setting. It remains an interesting future research problem to develop relevant analysis strategies for these practically relevant variants of the SGD.
\end{remark}

Next we handle the dynamic randomness, and the proofs of the lemmas can be found in Appendix \ref{Pro-SecSGD}. 
Let \(\FF_t\) be the known information after \(t\) iterations  (i.e., the filtration generated by the random initialization and the random iteration index selection up to the $t$th iteration) and let
$R_{\bm{w}}$ and $R_{\bm{a}}$ be taken as \eqref{RwRa}.
Let \(S\) be the stopping time such that whenever \(t< S\), we have
\begin{equation}\label{Stopping}
	\|\bm{w}_r(t) - \bm{w}_r(0)\|_2 \leq R_{\bm{w}}, \;
	\|a_r(t) - a_r(0)\|_2 \leq R_{\bm{a}}, \quad \forall r\in [m], 
\end{equation}
while at least one of the above inequalities fails for \( t=S \). In other words, \(S\) is the first time that some parameter leaves a given neighborhood of the initialization. If the event \(\init\) happens, then for \(t<S\), we have by Lemma \ref{GramDyn}, 
\[ \lambda_{\min}(\bm{G}_{\bm{w}}(t)) \geq \frac{1}{2}\lambda_{\bm{w}},\;
\lambda_{\min}(\bm{G}_{\bm{a}}(t)) \geq \frac{1}{2}\lambda_{\bm{a}} \quad \Longrightarrow \quad
\lambda_{\min}(\bm{G}_{\theta}(t)) \geq \frac{1}{2}\lambda_{\bm{w}}
+ \frac{1}{2}\lambda_{\bm{a}} = \frac{1}{2}\lambda_{\theta}. \]
During the iteration, we need to control the distance of \(\bm{w}_r(t+1)\) from \(\bm{w}_r(0)\) 
for all \(r \in [m]\). We will control them uniformly and thus let:
\begin{gather*}
	\left\| \bm{w}_*(t) - \bm{w}_*(0) \right\|_2 := 
	\sup_{r\in [m]} \left\{ \left\| \bm{w}_r(t) - \bm{w}_r(0) \right\|_2 \right\}, \\
	\left\| \bm{w}_*(t+1) - \bm{w}_*(t) \right\|_2 := 
	\sup_{r\in [m]} \left\{ \left\| \bm{w}_r(t+1) - \bm{w}_r(t) \right\|_2 \right\}, \\
	\left\|\frac{\partial \wt{L}(\bm{w}, \bm{a})}{\partial \bm{w}_*} \right\|_2 := 
	\sup_{r\in [m]} \left\{ 
	\left\| \frac{\partial \wt{L}(\bm{w}, \bm{a})}{\partial \bm{w}_r} \right\|_2 \right\}.
\end{gather*}
This also applies to other quantities and the counterpart related to \(a_r, r\in [m]\).

\begin{lemma}\label{StepGap-w_t}
	If the event \(E(B-1)\) happens, then for all \(t>0\), we have 
	\begin{align*}
		\mathbb{E} \left[ \left\| \bm{w}_*(t+1) - \bm{w}_*(t) \right\|_2^2 
		\cdot \mathbf{1}_{S> t} \mid \mathcal{F}_t \right]& 
		\lesssim \eta^2 \frac{C_B^2 B^6}{m}L(t),  \\
		\mathbb{E} \left[ \left| a_*(t+1) - a_*(t) \right|^2 
		\cdot \mathbf{1}_{S> t} \mid \mathcal{F}_t \right] 
		&\lesssim \eta^2 \frac{C_B^2 B^6}{m}L(t). 
	\end{align*}
\end{lemma}

While the estimates in Lemma \ref{StepGap-w_t} hold for all \(t>0\), they make sense only  
within the stopping time $S$ since \(\mathbf{1}_{S> t} = 0\) for \(t\geq S\). 
Next we bound the expectation of the loss by mathematical induction with the occurrence of the event \(\init\), cf. \eqref{initialization}.

\begin{lemma}\label{LossGap}
	Let \(\lambda_\theta = \lambda_{\bm{w}} + \lambda_{\bm{a}} \). If the event \(\init\) happens, 
    \[ m\gtrsim \max\left\{C_B^2 B^6,\log^3\left(\frac{n_1+n_2}{\delta}\right)\right\}\quad \mbox{and}\quad
    \eta \lesssim \frac{\lambda_\theta }{C_B^3 B^{9}}\sqrt{\frac{m}{L(0)}}, \]  
    then for all \(t>0\),
	\begin{gather*}
		\mathbb{E} \left[ L(t+1)
		\cdot \mathbf{1}_{S> t+1} \mid \mathcal{F}_{t} \right] 
		\leq \left(1-\eta \frac{\lambda_\theta}{2}\right)L(t) \cdot \mathbf{1}_{S>t}.
	\end{gather*}
\end{lemma}

For all \(t>0\), using Lemma \ref{LossGap}, we can estimate the expectation of \(L(t)\) by mathematical induction:
\begin{align*}
	\E \left[ L(t) \cdot \mathbf{1}_{S> t} \right] 
	& = \E \left[ \mathbb{E} \left[ L(t)
	\cdot \mathbf{1}_{S> t} \mid \mathcal{F}_{t-1} \right] \right] 
	 \leq \left( 1- \eta \frac{\lambda_\theta}{2} \right) 
	\E[  L(t-1) \cdot \mathbf{1}_{S> t-1} ]  \leq \left( 1- \eta \frac{\lambda_\theta}{2} \right)^t L(0).
\end{align*}
Thus, the loss  in the expectation sense can be controlled within the stopping time $S$.

\begin{corollary}\label{E[L(t)]}
With the settings in Lemma \ref{LossGap}, there holds for all \(t>0\), 
	\begin{align*}
		\E \left[ L(t) \cdot \mathbf{1}_{S> t} \right] 
		 \leq \left( 1- \eta \frac{\lambda_\theta}{2} \right)^{t} L(0).
	\end{align*}
\end{corollary}

Using the above estimates, we can now bound the distance of the parameters
\(\bm{w}_*(S)\) and \(a_*(S)\) at time \(S\) from the initialization. 
\begin{lemma}\label{w(S)-w(0)}
	With the settings in Lemma \ref{LossGap}, we have
	\begin{gather*}
		\E \left[ \left\| \bm{w}_*(S) - \bm{w}_*(0) \right\|_2 \right] 
		\lesssim \frac{C_B B^3}{\lambda_\theta} \sqrt{ \frac{L(0)}{m} }\quad \mbox{and} \quad
		\E \left[ | a_*(S) - a_*(0) | \right] 
		\lesssim \frac{C_B B^3}{\lambda_\theta} \sqrt{ \frac{L(0)}{m} }.
	\end{gather*}
\end{lemma}

Finally, by combining the preceding lemmas, we deduce that \(S=\infty\) holds with high probability, and then the estimate for the loss holds
during the whole training process.
\begin{theorem}\label{SGDThm}
Let \(\lambda_\theta = \lambda_{\bm{w}} + \lambda_{\bm{a}} \). For \(\delta, \wt{\delta}\in (0,1)\), if \(m\) and \(\eta\) satisfy
\begin{equation}\label{m-eta}
m\gtrsim \frac{C_B^4 d^6}{\min\left\{\lambda_{\bm{w}}^2,\lambda_{\bm{a}}^2\right\}}
        \log\left( \frac{n_1+n_2}{\delta} \right) \cdot
        \max\left\{ \log^2\left( \frac{n_1+n_2}{\delta} \right),
		\frac{C_B^4 d^3 B^{6} }{{\wt{\delta}}^2 \lambda_\theta^2 }
            \right\}\quad \mbox{and}
		\quad \eta \lesssim \frac{\lambda_\theta }{C_B^3 B^{9}}\sqrt{\frac{m}{L(0)}},
	\end{equation}
	where 
	\[ B = 1 + \sqrt{2(d+1)\log\left(\frac{2m(d+1)}{\delta}\right)} \]
	and \(C_B\) grows at most polynomially in \(B\),
	then with probability at least \( 1- 4\delta \) over the 
	initialization, we have
	$\P\left( S=\infty \right) \geq  1-\wt{\delta} $, and then, for all \(t\in \mathbb{N}\),
	\begin{align*}
		\E \left[ L(t) \cdot \mathbf{1}_{S=\infty}  \right] 
		\leq \left( 1- \eta \frac{\lambda_\theta}{2} \right)^t L(0).
	\end{align*}
\end{theorem}

\begin{remark}\label{rmk:m-dependence} 
By Markov's inequality, the exponential decay in expectation ensures that the loss \(L(t)\) can be arbitrarily small with high probability provided the number of iterations \(t\) is sufficiently large. Note that the inequality \eqref{m-eta} on the width \(m\) of the neural network should be interpreted as an implicit constraint, since \(B\) is of order \(\sqrt{\log m}\), while \(C_B\) has at  most polynomial growth of \(B\). Due to the dominance of the linear factor over the logarithmic factors, the constraint can be fulfilled by taking $m$ sufficiently large and the condition on \(m\) in (\ref{m-eta}) is not excessively restrictive. 
For \(C_B\) satisfying the polynomial growth \(C_B\lesssim 1 + B^\ell\) in Assumption \ref{AssAct} (e.g., \(\ell = 2\) for \(\sigma = \mathrm{ReLU}^3\)), then the condition requires
\[ m\gtrsim B^{8\ell + 6}  \quad \Longrightarrow \quad m\gtrsim \log^{4\ell + 3} m . 
\]
However, the analysis may be improved for the RePU activation as the current estimates hold for very general \(\sigma\).
For the case \(C_B=1\) (i.e., the activation \(\sigma\) is globally Lipschitz), we have approximately
\[ 
m\geq CB^{6} \quad \Longrightarrow \quad m\geq C \log^{3} m \quad \Longrightarrow \quad m \geq C \log^3 C^{1+\epsilon} \quad (\epsilon>0 \text{ is small when } C \text{ is large}). 
\]
There is an extra term \(\wt{\delta^2}\) for \(m\) in \eqref{m-eta} arising from the inequalities \eqref{OverRan} in the proof, which is used to overcome the randomness. Moreover, if the scheme is deterministic, then a sharper estimate can be exploited in Lemma \ref{LossGap} and a lower bound for \(\eta\) is derived. Thus, a larger \(m\) and a smaller \(\eta\) are needed due to the stochasticity of the optimization algorithm, which agrees with our intuition; see also the recent work \cite{GaoTan:2025KAN} on a similar phenomenon in the training of Kolmogorov-Arnold networks.
In addition, the problem data \(f\) and \(g\) in the PDE \eqref{PoissonPDE} also affects the implicit constant in the inequality.
\end{remark}

\section{Continuous Time Model}\label{SecSGF}
Now, consider the stochastic gradient flow. The analysis requires the local Lipschitz property of the activation function \(\sigma\) up to order four in Assumption \ref{AssAct} in order to apply Ito's formula
and to estimate the Hessian of the loss.
For the continuous time model, consider the following stochastic differential equation (SDE):
\begin{equation}\label{ParaSDE}
	\d\theta_t = - \nabla L(\theta_t)\d t + \sqrt{\eta} \sigma(\theta_t) \d W_t,
\end{equation} 
or equivalently, the parameters \(\theta\) satisfy
\begin{equation}
	\theta(t) = \theta(0) - \int_{0}^{t} \nabla L(\theta_\tau) \d\tau + 
	\sqrt{\eta} \int_{0}^{t} \sigma(\theta_\tau) \d W_\tau,
\end{equation}
where \(W_t\in \R^{m(d+2)} \) is the standard Wiener process, \(\eta\) is the step size
and \( \sigma(\theta_t) \in \R^{m(d+2)\times m(d+2)} \) satisfies
\[ \sigma(\theta) = \sqrt{\Sigma(\theta)}, \quad \mbox{with }\Sigma(\theta) = 
\E\left[ ( \nabla \wt{L}(\theta) -\nabla L(\theta) )
( \nabla \wt{L}(\theta) -\nabla L(\theta) )^\top \right]. \]
See \cite{Li2019} for more details about the derivation of the continuous time model \eqref{ParaSDE}. For the component $\boldsymbol{w}_r$ of \(\theta\), we have 
\[ \bm{w}_r(t) = \bm{w}_r(0) - \int_{0}^{t} \frac{\partial L(\theta_\tau) }{\partial \bm{w}_r}  \d\tau
+ \sqrt{\eta}\xi_{\bm{w}_r}, \]
where \(\xi_{\bm{w}_r}\) denotes the corresponding part of the noise vector
\( \int_{0}^{t} \sigma(\theta_t) \d W_t \). There is a similar formula for the component \(a_r\).
Similar to the case of SGD, 
we define the stopping time \(S\) by the infimum \(t>0\) 
such that at least one of the following inequalities fails, i.e.,
\begin{equation}
	\|\bm{w}_r(t) - \bm{w}_r(0)\|_2 \leq R_{\bm{w}}, \;
	\|a_r(t) - a_r(0)\|_2 \leq R_{\bm{a}}, \quad \forall r\in [m].
\end{equation}
Throughout we assume that the SDE \eqref{ParaSDE} for \(\theta\) has a unique maximal local solution up to its (random) blow-up time, which is obviously behind the stopping time $S$. 
For the rigorous treatment of the deterministic case based on the PL property of \(\nabla L\) discussed below, we refer
to \cite[Theorem 1.2]{Chatterjee2022}.

We now prove that the loss function $L(\theta_t)$ is \(\lambda_\theta\)-Polyak-Lojasiewicz within stopping time $S$; 
see \cite{Liu2022,Chatterjee2022,An2023,Lugosi2024} for more about the convergence analysis related to 
Polyak-Lojasiewicz property. Similar to Section \ref{SecSGD}, we always assume that the event \(\init\) in \eqref{initialization} happens. First, by the chain rule
\begin{gather*}
	\nabla L(\theta_t) = \sum_{p=1}^{n_1} s_p(\theta_t) \nabla s_p(\theta_t) +
	\sum_{q=1}^{n_2} h_q(\theta_t) \nabla h_q(\theta_t).
\end{gather*}
For \(t<S\), we have \(\lambda_{\min}(\bm{G}_{\theta}(t))\geq \frac{\lambda_\theta}{2}\). Thus
\begin{align*}
 \llan \nabla L(\theta_t) , \nabla L(\theta_t) \rran 
	& = \llan \sum_{p=1}^{n_1} s_p(\theta_t) \nabla s_p(\theta_t) +
	\sum_{q=1}^{n_2} h_q(\theta_t) \nabla h_q(\theta_t), 
	  \sum_{p=1}^{n_1} s_p(\theta_t) \nabla s_p(\theta_t) +
	 \sum_{q=1}^{n_2} h_q(\theta_t) \nabla h_q(\theta_t) \rran \\
	& = \begin{bmatrix}
		\bm{s}(\theta_t) \\
		\bm{h}(\theta_t)
	\end{bmatrix}^\top \bm{G}_{\theta}(t) \begin{bmatrix}
		\bm{s}(\theta_t) \\
		\bm{h}(\theta_t)
	\end{bmatrix} \geq \frac{1}{2}\lambda_\theta \cdot 2L(\theta_t) 
	= \lambda_\theta L(\theta_t). 
\end{align*}
Since the loss function \(L(\theta_t)\) is nonnegative, we obtain the $\lambda_\theta$-Polyak-Lojasiewicz inequality 
\[ L(\theta_t) - \inf L \leq L(\theta_t) 
\leq \frac{1}{\lambda_\theta} \|\nabla L(\theta_t)\|_2^2. \]

Then using Ito's formula and the \(\lambda_\theta\)-Polyak-Lojasiewicz property, we get an estimate on the logarithm of the loss $L(\theta_t)$. Note that we discuss the problem only under the circumstance
such that the loss \(L(\theta)\) does not attain zero in finite time, otherwise we are done in the opposite case. The proofs can be found in Appendix \ref{Pro-SecSGF}.

\begin{lemma}\label{SDElogLoss}
    Let \(\lambda_\theta = \lambda_{\bm{w}} + \lambda_{\bm{a}} \).
	If the event \(\init\) happens, then for \(t < S\),
	\begin{align*}
		\log L(\theta_t) & \leq \log L(\theta_0)  - \lambda_\theta t 
		- \frac{1}{2} \langle M \rangle_t + M_t  + \frac{\eta}{2} \int_{0}^{t} \frac{1}{L(\theta_\tau)}\mathrm{tr}
		\left( \sigma(\theta_\tau)^\top 
		\mathrm{Hess}( L(\theta_\tau)) \sigma(\theta_\tau) \right) \d\tau,
	\end{align*}
	where \(M_t\) and \( \langle M \rangle_t \) are the local martingale and its quadratic variation defined respectively by 
    \begin{equation}\label{Martin-M_t}
         M_t :=  \int_{0}^{t}  \sqrt{\eta}
	\frac{\nabla L(\theta_\tau) ^\top \sigma(\theta_\tau)}{L(\theta_\tau)} \d W_\tau\quad \mbox{and}\quad        \langle M \rangle_t = \int_{0}^{t} \left\| \sqrt{\eta}
	\frac{\nabla L(\theta_\tau) ^\top \sigma(\theta_\tau)}{L(\theta_\tau)} \right\|_2^2 \d\tau. 
    \end{equation}
\end{lemma}

%

We first analyze the terms related to the local martingale \(M_t\).
Let 
\[
v(\tau) := \sqrt{\eta}\frac{\nabla L(\theta_\tau) ^\top \sigma(\theta_\tau)}{L(\theta_\tau)}.
\]
Then we have \(\d M_t = v(t) \d W_t \) and the exponential martingale 
\(\mathcal{E}_t := \exp \left( M_t - \frac{1}{2}\langle M \rangle_t \right)\)   
of \(M_t\) takes the form
\[ \mathcal{E}_t := \exp \left( M_t - \frac{1}{2}\int_{0}^{t} 
	\left\|v(\tau)\right\|_2^2 \d\tau  \right). \]
Using the multi-dimensional Ito's formula, we have
\begin{align*}
	\d\mathcal{E}_t =  \mathcal{E}_t v(t) \d W_t,
\end{align*}
which implies it is a supermartingale, i.e., \(\E\left[ \mathcal{E}_t \mid \FF_s \right]\leq \mathcal{E}_s\), or more rigorously \(\E\left[ \mathcal{E}_t \cdot \mathbf{1}_{S> t} \mid \FF_s \right]\leq \mathcal{E}_s \cdot \mathbf{1}_{S> s}\) by Lemma \ref{CondiExpe}, for \(s\leq t\). We refer interested readers to \cite[Theorem 1.2]{Kazamaki:2006} and the explanation after Example 1.1 therein for further details.
\begin{lemma}
	Within the stopping time $S$, the stochastic process 
	\(\mathcal{E}_t = \exp \left( M_t - \frac{1}{2}\langle M \rangle_t \right)\) 
	is a supermartingale. In particular, there holds \(\E[\mathcal{E}_t \cdot \mathbf{1}_{S> t}] \leq \E[\mathcal{E}_0] = 1\).
\end{lemma}

We next deal with the trace term in Lemma \ref{SDElogLoss}.
\begin{lemma}\label{tr(HessL)}
	If the event \(\init\) happens and \(m\gtrsim \log^3\left(\frac{n_1+n_2}{\delta}\right)\), then for \(t< S\), we have
	\[  \frac{1}{L(\theta_t)}\mathrm{tr}
	\left( \sigma(\theta_t)^\top \mathrm{Hess}( L(\theta_t)) \sigma(\theta_t) \right)
	\lesssim C_B^3 B^6 \left( C_B d^3 + \sqrt{d^3 L(\theta_t)} \right). \]
\end{lemma}

Now we can establish the decay of the loss, and then bound the distance of 
parameters from the initialization within the stopping time $S$.
\begin{lemma}\label{ConE[L(t)]} 
Let \(\lambda_\theta = \lambda_{\bm{w}} + \lambda_{\bm{a}} \).
If the event \(\init\) happens,
\[ m\gtrsim \log^3\left(\frac{n_1+n_2}{\delta}\right)\quad \mbox{and} \quad 
\eta \lesssim \lambda_\theta \left(C_B^4 d^3 B^6\log^{\frac{1}{2}}\left(\frac{n_1+n_2}{\delta}\right) \right)^{-1}, \]
then for all \(t>0\), 
	\begin{align*}
		\E \left[ L(t) \cdot \mathbf{1}_{S> t} \right] 
		 \leq \exp\left( -\frac{\lambda_\theta}{2}t \right) L(0).
	\end{align*}
\end{lemma}

\begin{lemma}\label{Con_w(S)-w(0)} 
With the settings in Lemma \ref{ConE[L(t)]}, we have for all \(t>0\),
	\begin{gather*}
		\E\left[ \|\bm{w}_*(t) - \bm{w}_*(0)\|_2 \cdot \mathbf{1}_{S> t} \right] \leq 
		\frac{C_B B^3}{\lambda_\theta} \sqrt{ \frac{L(0)}{m} } 
		+ C_B B^3 \sqrt{ \frac{\eta L(0)}{\lambda_\theta} }, \\
		\E\left[ \|a_*(t) - a_*(0)\|_2 \cdot \mathbf{1}_{S> t} \right] \leq 
		\frac{C_B B^3}{\lambda_\theta} \sqrt{ \frac{L(0)}{m} } 
		+ C_B B^3 \sqrt{ \frac{\eta L(0)}{\lambda_\theta} }.
	\end{gather*}
\end{lemma}

The following theorem is obtained by combining the preceding lemmas like Theorem \ref{SGDThm}. Specifically, first, let \(t\) approaches the stopping time $S$ from left, then using Lemmas \ref{ConE[L(t)]}, \ref{Con_w(S)-w(0)} and Markov inequality, we deduce that \(S=\infty\) with high probability.
Finally, by conditioning on \(S=\infty\), we get the exponential decay of the loss by Lemma \ref{ConE[L(t)]} again.
We omit the details of the proof.
\begin{theorem}\label{ConThm}
Let \(\lambda_\theta = \lambda_{\bm{w}} + \lambda_{\bm{a}} \). For \(\delta, \wt{\delta}\in (0,1)\), if \(m\) and \(\eta\) satisfy
	\begin{equation*}
		m\gtrsim \frac{C_B^4 d^6}{\min\left\{\lambda_{\bm{w}}^2,\lambda_{\bm{a}}^2\right\}}
        \log\left( \frac{n_1+n_2}{\delta} \right) \cdot
        \max\left\{ \log^2\left( \frac{n_1+n_2}{\delta} \right),
		\frac{C_B^4 d^3 B^{6} }{{\wt{\delta}}^2 \lambda_\theta^2 }
            \right\},
		\quad \eta \lesssim \frac{\wt{\delta}^2 \lambda_\theta \min\left\{\lambda_{\bm{w}}^2,\lambda_{\bm{a}}^2\right\}}{C_B^8 d^9 B^{6}\log\left( \frac{n_1+n_2}{\delta} \right)},
	\end{equation*}
	where 
	\[ B = 1 + \sqrt{2(d+1)\log\left(\frac{2m(d+1)}{\delta}\right)} \]
	and \(C_B\) grows at most polynomially in \(B\),
	then with probability at least \( 1- 4\delta \) over the initialization, we have
$\P\left( S=\infty \right) \geq 1-\wt{\delta}$, and then, for all \(t\in \mathbb{R}_+\),
	\begin{align*}
		\E \left[ L(t) \cdot \mathbf{1}_{S=\infty} \right] 
		\leq \exp\left( - \frac{\lambda_\theta}{2}t \right) L(0).
	\end{align*}
\end{theorem}

\section{Numerical example}\label{sec:num}
In this section, we present numerical results for the two-dimensional Poisson equation on the unit square $\Omega = (0,1)^2$, with the source $f  = -2\pi^2\sin(\pi x)\sin(\pi y)$ and a zero Dirichlet boundary condition. Then the exact solution $u$ for the Poisson problem is given by \(u = \sin(\pi x)\sin(\pi y)\). In the PINN, we employ two-layer neural networks with the hidden layer width \(m\) and tanh activation, and the neural network parameters $\theta$ are initialized to the standard Gaussian. To form the empirical loss $L(\theta)$, we randomly sample 2000 points in the domain $\Omega$ and 400 points on the boundary $\partial\Omega$ (100 points on each side) with the penalty parameter \(\gamma\) in the PINN loss $L(\theta)$ in \eqref{PINN Loss} set to 10. 

\begin{figure}[h]
    \centering
    \begin{subfigure}{0.48\textwidth}
        \centering
        \includegraphics[width=0.9\linewidth]{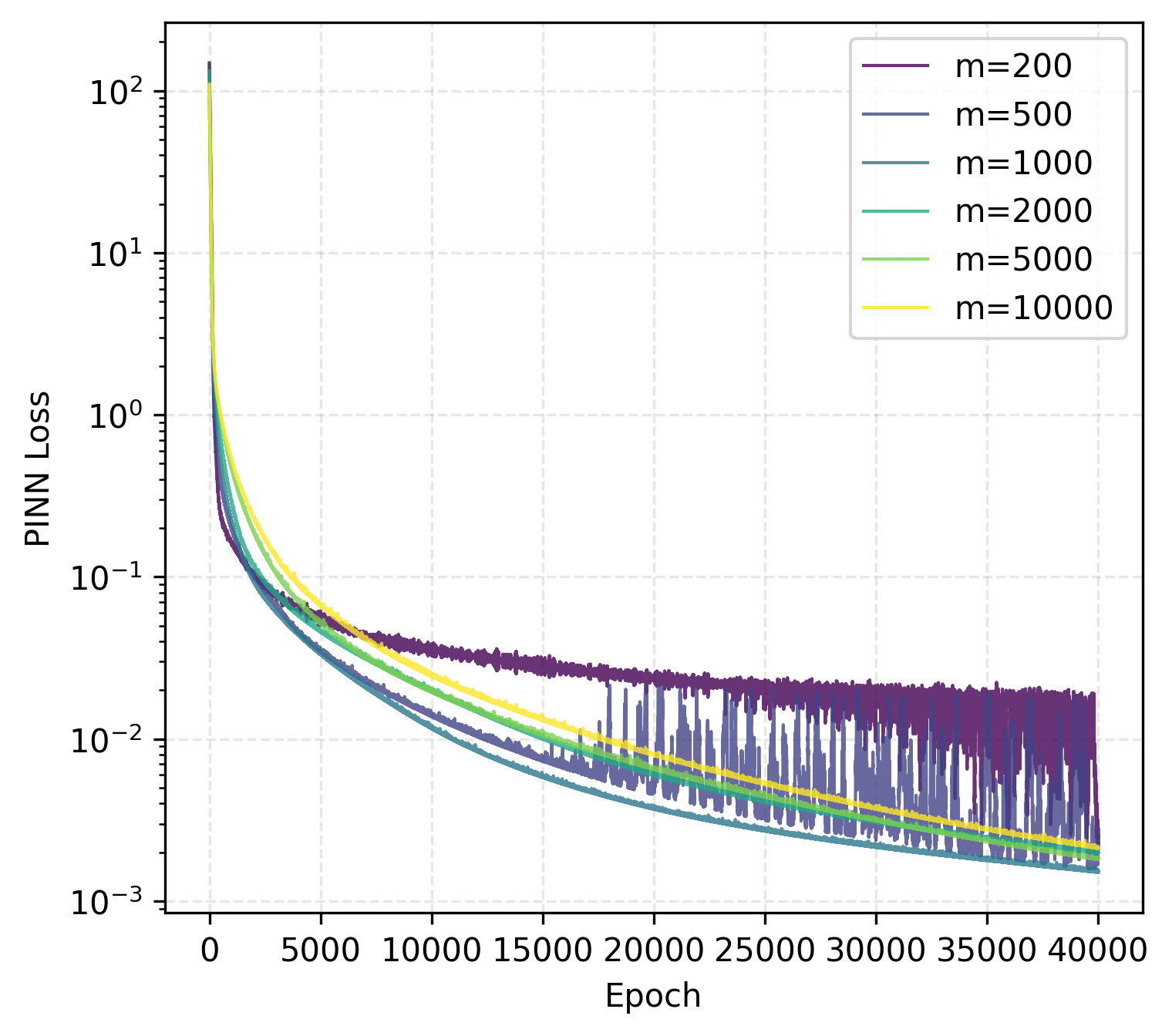}
    \end{subfigure}
    \hfill
    \begin{subfigure}{0.48\textwidth}
        \centering
        \scalebox{0.93}[1]{\includegraphics[width=0.97\linewidth]{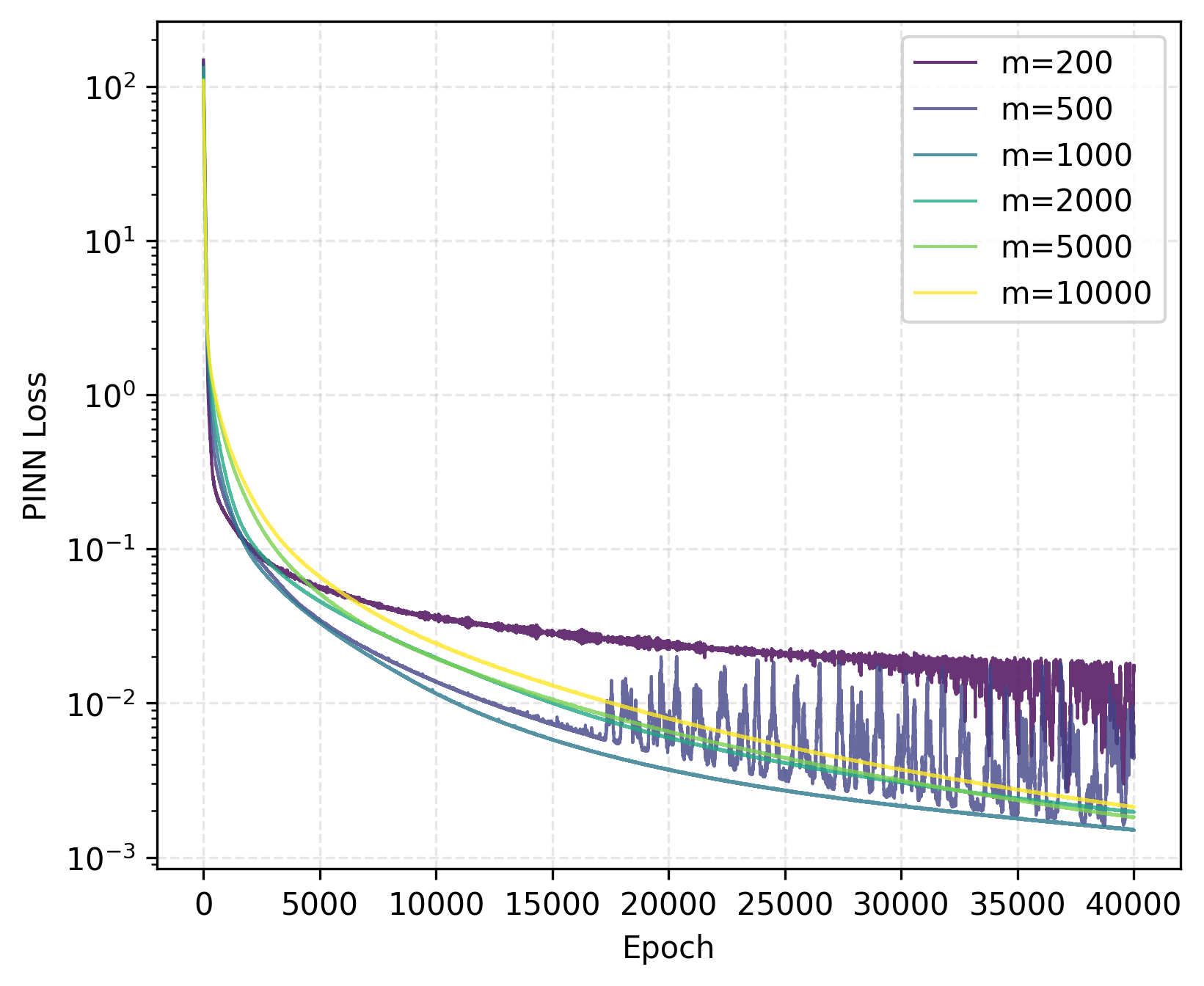}}
\end{subfigure}

    \begin{subfigure}{0.48\textwidth}
        \centering
        \includegraphics[width=0.9\linewidth]{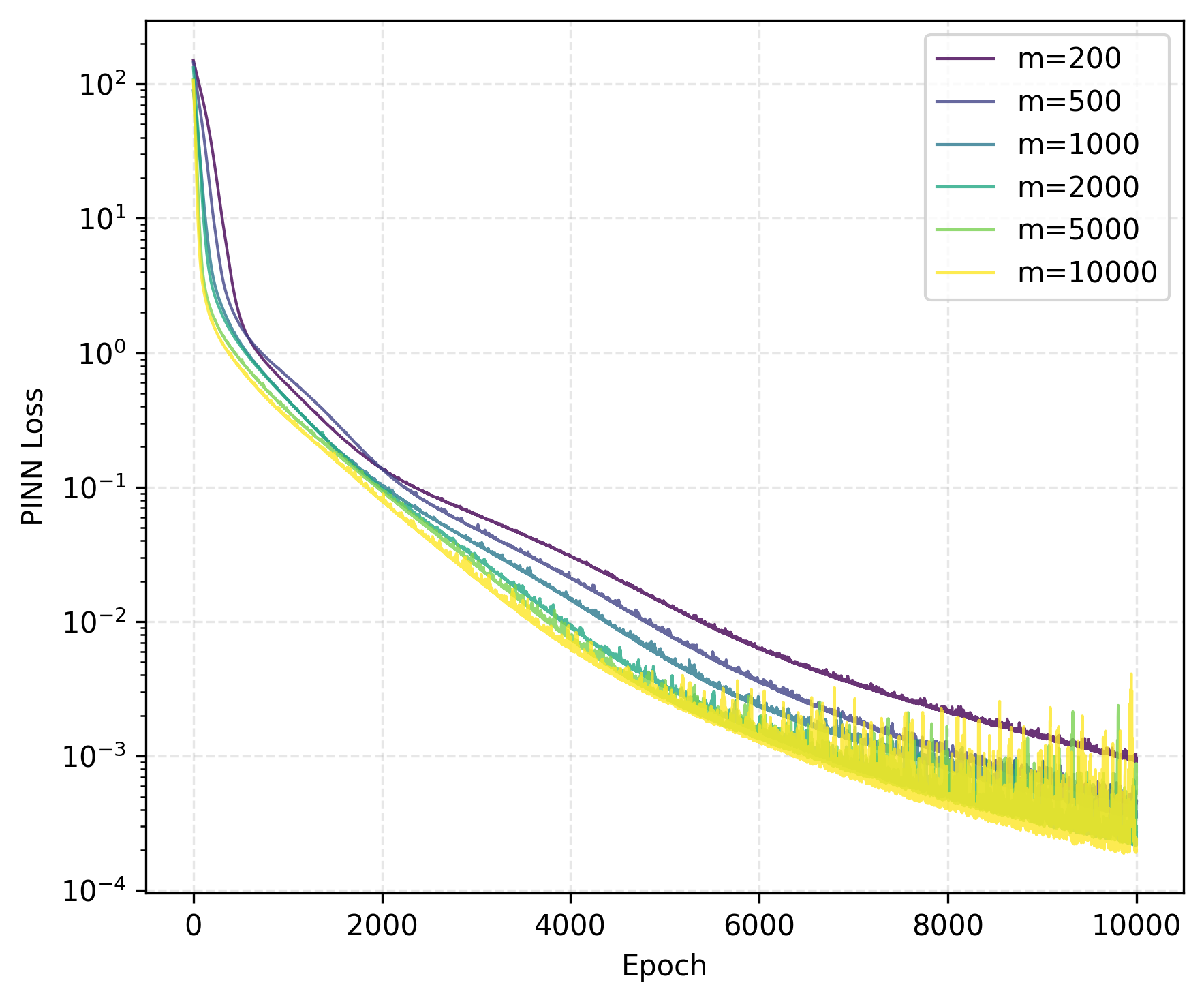}
        \caption{batch size = (512,128)}
        \label{SmaBat}
    \end{subfigure}
    \hfill
    \begin{subfigure}{0.48\textwidth}
        \centering
        \includegraphics[width=0.9\linewidth]{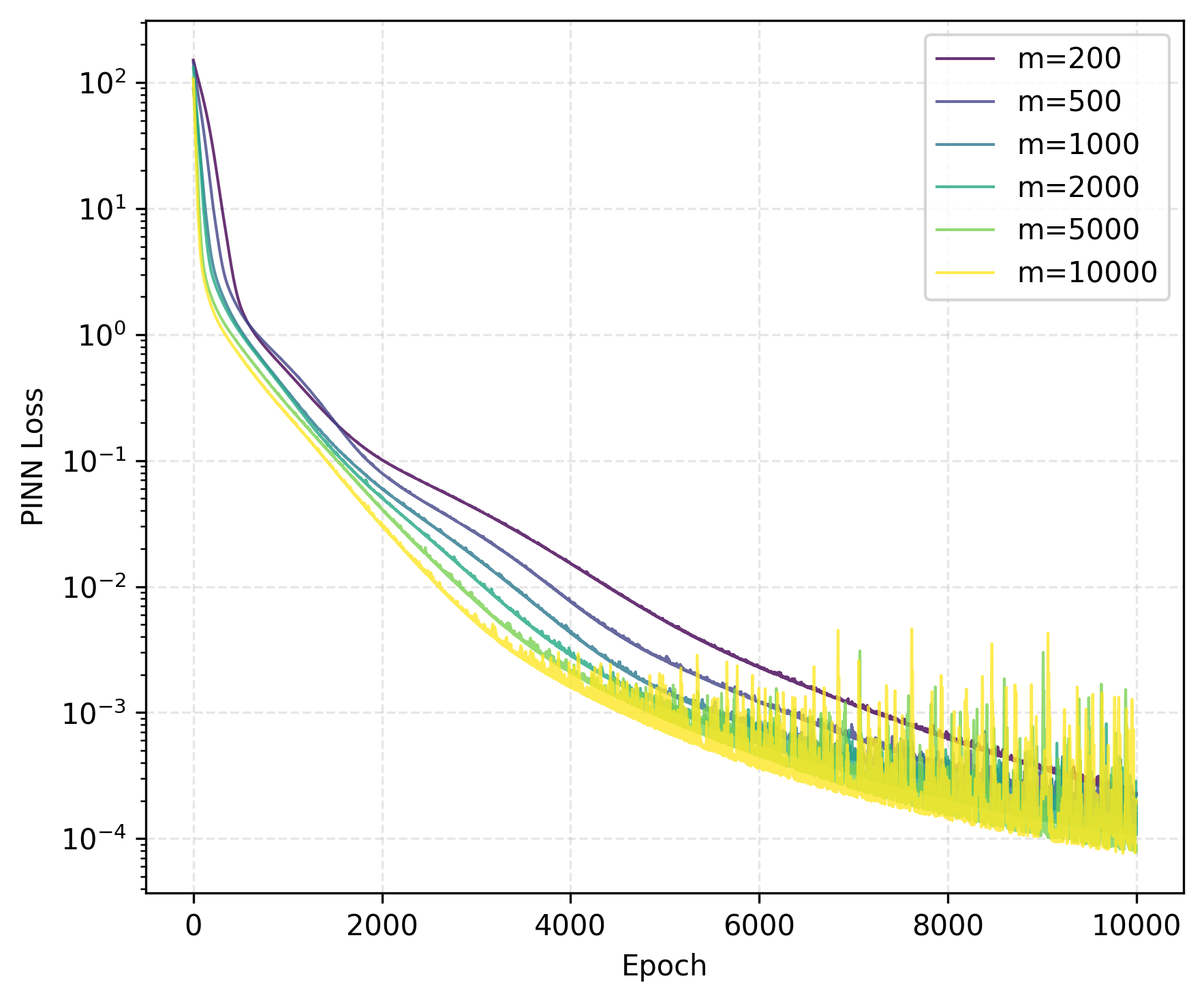}
        \caption{batch size = (1024,256)}
        \label{BigBat}
    \end{subfigure}\vskip -0.5em

    \caption{The loss versus epoch for different network width \(m\), trained with SGD (top) and Adam (bottom).}
    \label{Fig_Loss}
\end{figure}

\begin{figure}[h]
    \centering
    \begin{subfigure}{0.48\textwidth}
        \centering
        \includegraphics[width=0.9\linewidth]{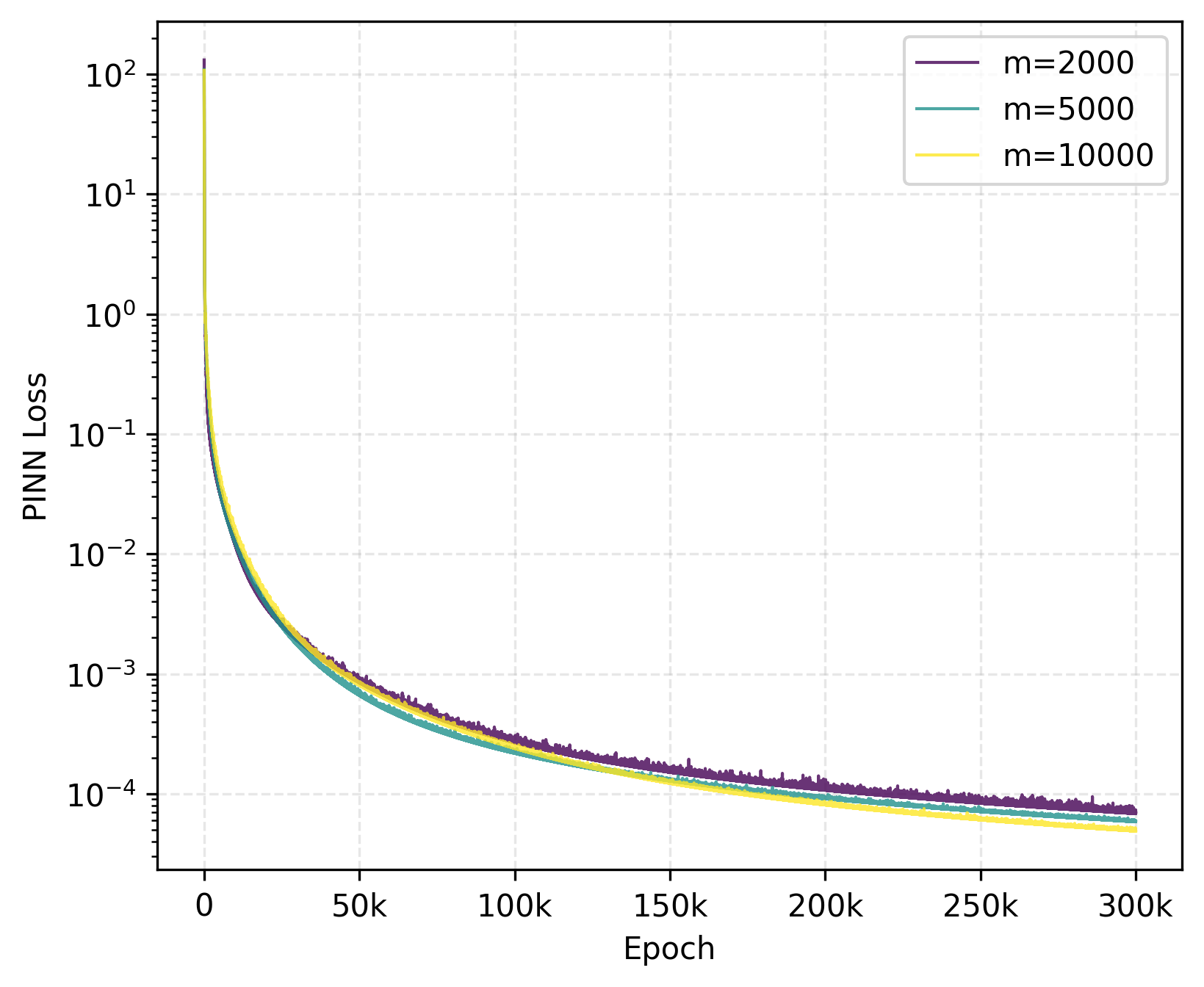}
        \caption{SGD}
    \end{subfigure}
    \hfill
    \begin{subfigure}{0.48\textwidth}
        \centering
        \includegraphics[width=0.9\linewidth]{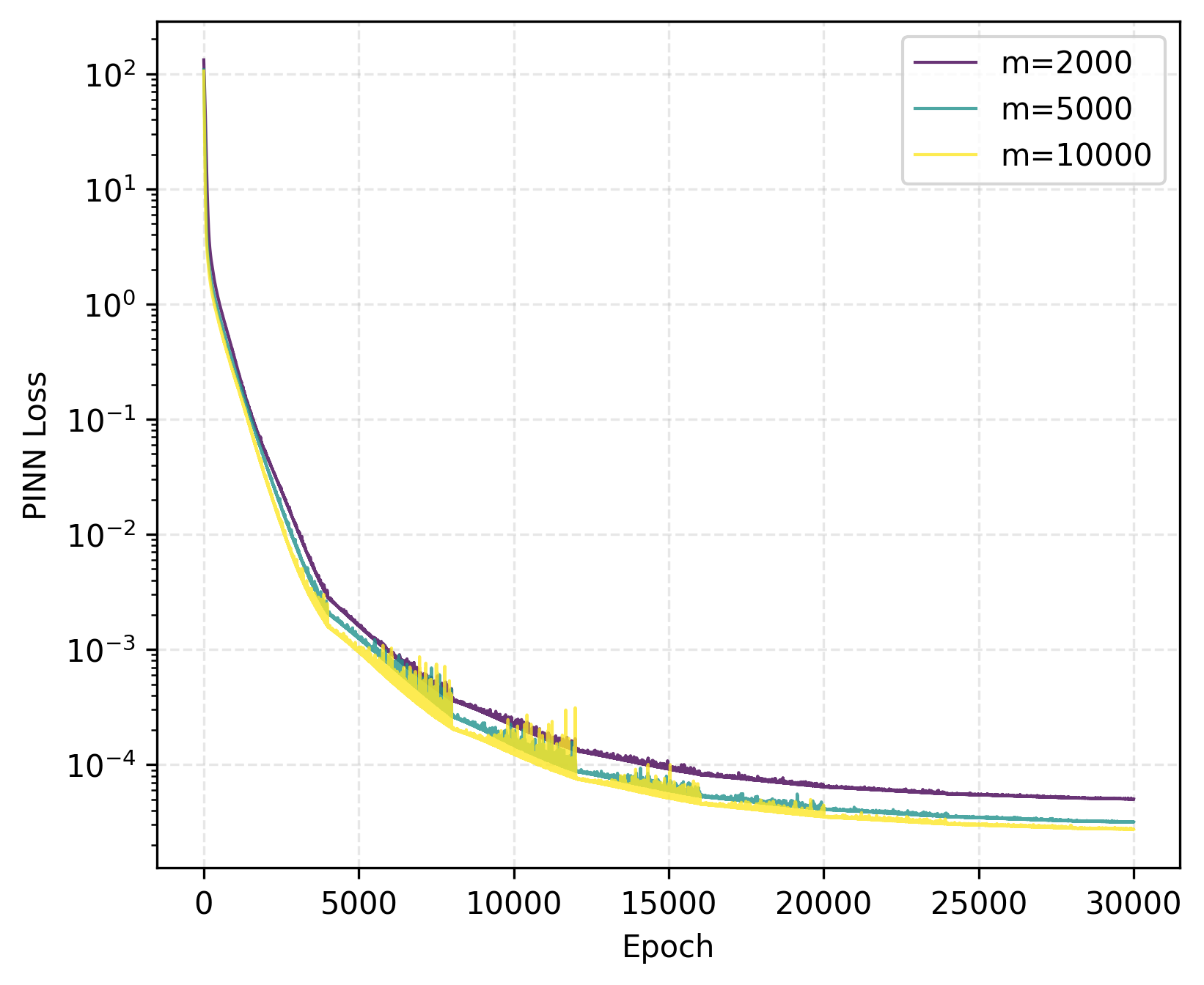}
        \caption{Adam}
    \end{subfigure}\vskip -0.5em

    \caption{The loss versus epoch for large neural network widths \(m\) with batch size = (1024,256), with SGD (left, learning rate $0.082$) and Adam (right, with adaptive learning rate).}
    \label{Fig_Loss-large_m}
\end{figure}

We present numerical results for both SGD and Adam \cite{Kingma:2017}. Adam is currently one of the most popular stochastic optimizers for training PINN type losses, and thus it is included as a benchmark.
Fig. \ref{Fig_Loss} shows the numerical results for different widths \(m\), based on the same training data. The top row shows the results for the SGD (with a learning rate 0.06), whereas the bottom row shows the results by Adam (with a learning rate 0.001). The left and right columns correspond to batch sizes of interior and boundary points given by (512, 128) and (1024, 256), respectively. The final loss exhibits a decreasing trend with increasing width \(m\).
However, the results by the SGD optimizer deviate slightly from the trend, occasionally showing  a small increase in the loss value. This may be attributed to the inherent variance of stochastic algorithms, as a larger 
\(m\) implies more complicated trajectory, thereby amplifying the stochastic effects. In contrast, the Adam optimizer consistently follows the decreasing trend, which aligns well with its known capability to adaptively reduce the variance of gradient estimates. 
Furthermore, the comparison of the plots within each row indicates that increasing the batch size reduces the oscillations along the iteration trajectories during the training, which is consistent with the intuition, as illustrated in Remark \ref{rmk:variance}. 
In Fig. \ref{Fig_Loss-large_m}, we present the numerical results for a large width \(m\) with batch size set to (1024,256). Both SGD (left) and Adam (right) exhibit a linear convergence rate, while a larger \(m\) results in a smaller final loss. This observation partially aligns with our theoretical findings. Notably, as the network width \(m\)
increases, the improvement in the final loss nearly diminishes. This suggests that a reasonably large width \(m\) already places the neural network model in the over-parameterized regime.

\section{Concluding remarks and discussions}\label{sec:concl}
In this work, we have established that both stochastic gradient descent and stochastic gradient flow can find a global minimum of the empirical PINN loss, under very generous assumptions on the activation function $\sigma$ so long as the strict positivity of the infinite Grammian holds (cf. Assumption \ref{InfGram}). When compared with the standard gradient descent, stochastic algorithms requires less computational effort per iterations but require  more neurons to guarantee the convergence. 

There are several avenues deserving further investigations. First, the analysis can only handle linear PDEs since the positive definiteness of Gram matrices may fail in the nonlinear case \cite{Bonfanti2024}. It remains an interesting open question to develop the convergence analysis for nonlinear PDEs. The existing positivity results \cite{Gao2023,Xu2024,ZhaoLuo:2025} cover only a limited range of activation functions. One more tractable theoretical issue is the positivity of the infinite Gram matrix for PINNs for a broad class of activation functions. Second, the smallest eigenvalues of Gram matrices play a central role in the convergence analysis. However, a precise characterization of the smallest eigenvalues is still largely missing. Empirically, they are deeply connected to the neural network architecture, activation functions and the distribution of the sampling points. We refer interested readers to the works \cite{Wang2021,Wang2022} for an empirical investigation of the spectral bias of PINNs.
A thorough theoretical analysis of the spectral behavior of the Gram matrices for PINNs is an important open problem. Third, it is of much interest to establish more fine-grained convergence theory for SGD type algorithms, e.g., with an explicit dependence on the batch size, as well as various variants of the standard SGD, e.g., Adam, SGD without replacement and i.i.d. sampling on the fly. Fourth and last, it is important to close the gap between the lazy training regime in NTK and practical scenarios in order to explain known failure modes of PINNs in challenging scenarios, e.g., nonlinear, stiff, or multiscale problems.

\section*{Acknowledgments}

The authors are grateful to the two anonymous referees and the associate editor handling the paper for their constructive comments which have significantly improved the quality of the paper. Bangti Jin is supported by Hong Kong RGC General Research Fund (Project 14306423 and Project 14306824), and a start-up fund from The Chinese University of Hong Kong. Longjun Wu is partly supported by International Joint PhD Supervision Scheme of The Chinese University of Hong Kong.

\bibliographystyle{abbrv}
\bibliography{reference}

\appendix

\section{Proofs for section \ref{StaRan}} \label{Pro-StaRan}

\subsection{Preliminary estimates}

First we estimate the derivatives of the loss  and then give relevant concentration inequalities. 
The neural network $\phi(\bm{x};\bm{w},\bm{a})$ and its derivatives with respect to the input $\bm{x}$ are given by
\begin{align*}
\phi(\bm{x}; \bm{w}, \bm{a}) &= \frac{1}{\sqrt{m}} \sum_{r=1}^{m} a_r \cdot \sigma \left( [\bm{w}_{r0} \, \bm{w}_{r1} \cdots \bm{w}_{rd}] \bm{x} + \bm{w}_{r,d+1} \right) 
= \frac{1}{\sqrt{m}} \sum_{r=1}^{m} a_r \cdot \sigma \left( \bm{w}_r^\top \wt{\bm{x}} \right),\\
\frac{\partial \phi(\bm{x};\bm{w},\bm{a})}{\partial x_i} &= \frac{1}{\sqrt{m}} \sum_{r=1}^{m} a_r \sigma'(\bm{w}_r^\top \wt{\bm{x}})w_{ri},\\
\frac{\partial^2 \phi(\bm{x};\bm{w},\bm{a})}{\partial x_i \partial x_j} &= \frac{1}{\sqrt{m}} 
\sum_{r=1}^{m} a_r \sigma''(\bm{w}_r^\top \wt{\bm{x}})w_{ri}w_{rj}.
	\end{align*}
Then the interior loss $s_p(\bm{w},\bm{a})$ and its derivatives with respect to the neural network parameters \(\bm{w}_r, a_r, r\in [m],\) are given by
\begin{equation}\label{int-loss} 
    \begin{aligned}
		s_p(\bm{w},\bm{a}) & =\frac{1}{\sqrt{n_1}} \left( 
		\sum_{i=1}^{d} \frac{1}{\sqrt{m}} 
		\sum_{r=1}^{m} a_r \sigma''(\bm{w}_r^\top \wt{\bm{x}}_p)w_{ri}^2 -f(\bm{x}_p) \right) \\
		& = \frac{1}{\sqrt{n_1 m}} \left( 
		\sum_{i=1}^{d} 
		\sum_{r=1}^{m} a_r \sigma''(\bm{w}_r^\top \wt{\bm{x}}_p)w_{ri}^2 - \sqrt{m} f(\bm{x}_p) \right),\\
		\frac{\partial s_p(\bm{w},\bm{a})}{\partial \bm{w}_r} & =  \frac{1}{\sqrt{n_1 m}} 
	 	\sum_{i=1}^{d} \left( a_r \sigma'''(\bm{w}_r^\top \wt{\bm{x}}_p)w_{ri}^2\wt{\bm{x}}_p + 
		a_r \sigma''(\bm{w}_r^\top \wt{\bm{x}}_p)2w_{ri}\bm{e}_i \right), \\
		\frac{\partial s_p(\bm{w},\bm{a})}{\partial a_r}  &= \frac{1}{\sqrt{n_1 m}}
			\sum_{i=1}^{d} \sigma''(\bm{w}_r^\top \wt{\bm{x}}_p)w_{ri}^2,
	\end{aligned}
\end{equation}
where \(\{\bm{e}_1,\ldots,\bm{e}_{d},\bm{e}_{d+1}\}\) denotes the standard Cartesian basis 
of \(\R^{d+1}\). Similarly, the boundary loss $h_q(\bm{w},\bm{a})$ and its derivatives are given by
\begin{align*}
		h_q(\bm{w},\bm{a}) & = \sqrt{\frac{\gamma}{n_2}}\left(\frac{1}{\sqrt{m}} \sum_{r=1}^{m} 
		a_r\sigma \left( \bm{w}_r^\top \wt{\bm{y}}_q \right)- 
		g(\wt{\bm{y}}_q) \right) \\
		& = \sqrt{\frac{\gamma}{n_2 m}}\left( \sum_{r=1}^{m} 
		a_r\sigma \left( \bm{w}_r^\top \wt{\bm{y}}_q \right)-
		\sqrt{m} g(\wt{\bm{y}}_q)\right),\\
		\frac{\partial h_q(\bm{w},\bm{a})}{\partial \bm{w}_r}  &=  
		\sqrt{\frac{\gamma}{n_2 m}}
		a_r\sigma' \left( \bm{w}_r^\top \wt{\bm{y}}_q \right)\wt{\bm{y}}_q, \\
		\frac{\partial h_q(\bm{w},\bm{a})}{\partial a_r} & = 
		\sqrt{\frac{\gamma}{n_2 m}} \sigma\left( \bm{w}_r^\top \wt{\bm{y}}_q \right).
	\end{align*}
The derivatives of the loss $L(\bm{w},\bm{a})$ can also be computed by the chain rule
\begin{gather*}
	\begin{aligned}
		\frac{\partial L(\bm{w}, \bm{a})}{\partial \bm{w}_r} 
		= \sum_{p=1}^{n_1} s_p(\bm{w}, \bm{a}) \cdot 
		\frac{\partial s_p(\bm{w}, \bm{a})}{\partial \bm{w}_r} +
		\sum_{q=1}^{n_2} h_q(\bm{w}, \bm{a}) \cdot 
		\frac{\partial h_q(\bm{w}, \bm{a})}{\partial \bm{w}_r},
	\end{aligned}\\
	\begin{aligned}
		 \frac{\partial L(\bm{w}, \bm{a})}{\partial a_r} 
		= \sum_{p=1}^{n_1} s_p(\bm{w}, \bm{a}) \cdot 
		\frac{\partial s_p(\bm{w}, \bm{a})}{\partial a_r} +
		\sum_{q=1}^{n_2} h_q(\bm{w}, \bm{a}) \cdot 
		\frac{\partial h_q(\bm{w}, \bm{a})}{\partial a_r}.
	\end{aligned}
\end{gather*}
If \(\bm{w}_r\) is bounded by \(B\geq 1\) and \(a_r\) is bounded by \(2\), then by
Assumption \ref{AssAct} and the Cauchy-Schwarz inequality, we have the following bounds on
the derivatives of the interior loss $s_p(\bm{w},\bm{a})$:
\begin{equation}\label{sDeriBound}
	\begin{gathered}
		\left\| \frac{\partial s_p(\bm{w},\bm{a})}{\partial \bm{w}_r} \right\|_2  
		\lesssim \frac{C_B}{\sqrt{n_1 m}} \left( \|\bm{w}_r\|_2^3 + 1  \right)
		\lesssim \frac{C_B B^3}{\sqrt{n_1 m}},\\
		\left\| \frac{\partial s_p(\bm{w},\bm{a})}{\partial a_r} \right\|_2  
		\lesssim \frac{C_B}{\sqrt{n_1 m}} \left( \|\bm{w}_r\|_2^3 + 1  \right)
		\lesssim \frac{C_B B^3}{\sqrt{n_1 m}}.
	\end{gathered}
\end{equation}
The counterparts for the boundary loss $h_q(\bm{w},\bm{a})$ are given by
\begin{equation}\label{hDeriBound}
	\begin{gathered}
		\left\| \frac{\partial h_q(\bm{w},\bm{a})}{\partial \bm{w}_r} \right\|_2  
		\lesssim \frac{C_B}{\sqrt{n_2 m}} \left( \|\bm{w}_r\|_2 + 1  \right)
		\lesssim \frac{C_B B}{\sqrt{n_2 m}}, \\
		\left\| \frac{\partial h_q(\bm{w},\bm{a})}{\partial a_r} \right\|_2  
		\lesssim \frac{C_B}{\sqrt{n_2 m}} \left( \|\bm{w}_r\|_2 + 1  \right)
		\lesssim \frac{C_B B}{\sqrt{n_2 m}}.
	\end{gathered}
\end{equation}

Note that the expressions of $ s_p$ and its derivatives involve $w_{ri}^2$, which are not necessarily bounded, and thus one cannot apply Hoeffding's inequality to bound these relevant quantities.  
Instead we employ the concentration inequality for sub-Weibull random variables defined below, which is a 
generalization of Hoeffding's inequality. 
\begin{definition}
 Let \(\psi_\alpha := \exp(x^\alpha) - 1\), with \(\alpha>0\). For a real-valued random variable \(X\), its \(\psi_\alpha\)-Orlicz norm is defined by 
\[ \|X\|_{\psi_\alpha} := \inf\left\{\epsilon > 0 : 
\mathbb{E}\left[\psi_\alpha\left(\frac{|X|}{\epsilon}\right)\right] \leq 1\right\}. \]
The random variable \(X\) is said to be sub-Weibull\((\alpha)\) if 
\(\|X\|_{\psi_\alpha}<\infty\). For a sub-Weibull\((\alpha)\) random variable \(X\), 
we have for any \(t\geq 0\),
\[ \mathbb{P}\left(|X| \geq t\right) \leq 
2 \exp\left(-\frac{t^\alpha}{\|X\|_{\psi_\alpha}^\alpha}\right). \]
Moreover, \(X\) is sub-Weibull\((\alpha)\) if it satisfies this inequality.
\end{definition}

The Orlicz norm $\|\cdot\|_{\psi_\alpha}$ essentially indicates the decay rate of the tail of random variables, e.g., \(\alpha = 2\) for Gaussians while \(\alpha = 1\) for sub-exponential variables.
The following important properties hold for sub-Weibull random variables \cite{Vladimirova2020}:
\begin{itemize}
	\item First, bounded random variables are sub-Weibull\((\alpha)\) for all \(\alpha>0\);
	\item Second, if \(X\) is sub-Weibull\((\alpha)\), then 
			\(X^{k}\) is sub-Weibull\(\left(\frac{\alpha}{k}\right)\) for \(k>0\);
	\item Third, \(\|\cdot\|_{\psi_\alpha}\) is a norm for \(\alpha\geq 1\) and a quasi-norm for 
    \(0<\alpha<1\).
        In particular, if both \(X\) and \(Y\) are sub-Weibull\((\alpha)\), then so is their sum \(X+Y\), and
        \begin{equation*}
            \begin{cases}
                \|X+Y\|_{\psi_\alpha}\leq  \|X\|_{\psi_\alpha} + \|Y\|_{\psi_\alpha}, & \text{if } \alpha\geq 1; \\
                \|X+Y\|_{\psi_\alpha}\leq  2^{\frac1\alpha}\left( \|X\|_{\psi_\alpha} + \|Y\|_{\psi_\alpha} \right), & \text{if } \alpha< 1.
            \end{cases}
        \end{equation*}    
\end{itemize}

\begin{lemma}[{\cite[Theorem 3.1]{Kuchibhotla2022}}]\label{Concen}
If $X_1, \ldots, X_n$ are independent mean zero random variables with $\|X_i\|_{\psi_\alpha} < \infty$ for all $1 \leq i \leq n$ and some $\alpha > 0$,  then for any vector $a = (a_1, \ldots, a_n) \in \mathbb{R}^n$, the following bound holds:
\begin{align*}
	\mathbb{P}\left(\left|\sum_{i=1}^{n} a_i X_i\right| \geq 2eC(\alpha) \|b\|_2 \sqrt{t} + 2eL_n^*(\alpha) t^{\frac1\alpha} \|b\|_{\beta(\alpha)}\right) &\leq 2e^{-t}, \quad \forall t \geq 0,
\end{align*}
where $b = (a_1 \|X_1\|_{\psi_\alpha}, \ldots, a_n \|X_i\|_{\psi_\alpha}) \in \mathbb{R}^n$,
\[
C(\alpha) := \max\{\sqrt{2}, 2^{\frac1\alpha}\} \times \begin{cases} 
	\sqrt{8}e^3(2\pi)^{\frac14}e^{\frac{1}{24}}\left(\frac{e^{\frac{2}{e}}}{\alpha}\right)^{\frac{1}{\alpha}}, & \text{if } \alpha < 1, \\
	4e + 2(\log 2)^{\frac{1}{\alpha}}, & \text{if } \alpha \geq 1,
	\end{cases}
	\]
	and for $\beta(\alpha) = \infty$ when $\alpha \leq 1$ and $\beta(\alpha) = 
	\frac{\alpha}{\alpha-1}$ when $\alpha > 1$,
	\[
	L_n(\alpha) := \frac{4^{\frac{1}{\alpha}}}{\sqrt{2}\|b\|_2} \times \begin{cases} 
	\|b\|_{\beta(\alpha)}, & \text{if } \alpha < 1, \\
	\frac{4e\|b\|_{\beta(\alpha)}}{C(\alpha)}, & \text{if } \alpha \geq 1,
	\end{cases}
	\]
	and the quantity $L_n^*(\alpha) = L_n(\alpha)C(\alpha)\frac{\|b\|_2}{\|b\|_{\beta(\alpha)}}$.
\end{lemma}

\subsection{Positivity of Gram matrices}\label{PosiGram}

The theoretical treatment to the Gram matrices in the context of PINN is still largely in its infancy. To the best of our knowledge, the positivity of NTK for PINNs has been rigorously proved in the following three cases: 
\begin{itemize}
    \item The heat equation with a Dirichlet boundary condition with RePU activation \cite{Gao2023};
    \item The heat equation with a Dirichlet boundary condition with a smooth activation \cite{Xu2024};
    \item A general class of admissible linear operators with a Dirichlet boundary condition with the tanh activation \cite{ZhaoLuo:2025}.
\end{itemize} 
The general procedure in these works proceeds as follows. First, suppose that no two samples of $\{\bm{x}_p\}_{p=1}^{n_1}\cup \{\bm{y}_q\}_{q=1}^{n_2}$  are identical, which implies that no two samples of the extended variables $\{\widetilde{\bm{x}}_p\}_{p=1}^{n_1}\cup \{\widetilde{\bm{y}}_q\}_{q=1}^{n_2}$ are parallel. Then one proves that the columns of the Jacobian matrices $\bm{D_w}(\bm{\theta}(0))$ and $\bm{D_a}(\bm{\theta}(0))$ are linearly independent in the Hilbert space $\mathcal{H}$ (which consists of measurable functions from $\mathbb{R}^{m(d+2)}$ to $\mathbb{R}^{m(d+1)}$ or $\mathbb{R}^m$ such that $\mathbb{E}_{\bm{\theta}(0)}[\|f(\bm{\theta}(0)\|_2^2]<\infty$). 

The proof techniques in these works apply also to problem \eqref{PoissonPDE}. Below we sketch the proof of the positivity for problem \eqref{PoissonPDE} for the convenience of readers. Let \(\theta = (\bm{w},\bm{a})\) be the initial parameters and we only consider the matrix \(\bm{G}_{\bm{a}}^{\infty} := \E_{\theta}\left[ \bm{G}_{\bm{a}}(\theta) \right]\) since the matrix \(\bm{G}_{\bm{w}}^{\infty}\) can be treated similarly. For any vector \(\bm u = [\alpha, \beta]^\top \in \R^{n_1+n_2}\), by the expressions \eqref{exp-Ga} and \eqref{exp-G_ify}, we have 
\begin{equation} \label{inner-product_G}
\begin{aligned}
 \bm{u}^\top \bm{G}_{\bm{a}}^{\infty} \bm{u} & = \bm{u}^\top \E_\theta \left[ D_{\bm{a}}^\top D_{\bm{a}} \right] \bm{u}
= \E_\theta \left[ (D_{\bm{a}}\bm{u})^\top (D_{\bm{a}}\bm{u}) \right] = \E_\theta \left[ \|D_{\bm{a}}\bm{u}\|_2^2 \right] \\
&= \E_\theta \left[ \sum_{r=1}^{m} \left\| \sum_{p=1}^{n_1}\alpha_p \frac{\partial s_p}{\partial a_r} + \sum_{q=1}^{n_2}\beta_q \frac{\partial h_q}{\partial a_r}\right\|_2^2 \right] \\
& = \E_{\bm{v}\sim \mathcal{N}(0,I_{d+1})} \left[ \left\| \sum_{p=1}^{n_1} \frac{\alpha_p }{\sqrt{n_1 }}
			\sum_{i=1}^{d} \sigma''(\bm{v}^\top \wt{\bm{x}}_p)w_{i}^2 + \sum_{q=1}^{n_2}\frac{\beta_q }{\sqrt{n_2/\gamma}} \sigma\left( \bm{v}^\top \wt{\bm{y}}_q \right) \right\|_2^2  \right].
\end{aligned}    
\end{equation}

Note that the Gaussian distribution \(\mathcal{N}(0,I_{d+1})\) has full support over \(\R^{d+1}\). By the continuity of \(\sigma\) and its derivatives, we only need to prove the linear independence of the functions 
\begin{equation}\label{funs-LI}
    \left\{ \sum_{i=1}^{d} \sigma''(\bm{v}^\top \wt{\bm{x}}_p)v_{i}^2, ~ \sigma\left( \bm{v}^\top \wt{\bm{y}}_q \right) : p\in[n_1], q\in[n_2]\right\}  
\end{equation}
given the sampling points $\{\bm{x}_p\}_{p=1}^{n_1}$ and $\{{\bm{y}}_q\}_{q=1}^{n_2}$. This is established in the following lemma. 
\begin{lemma}
If the sampling points $\{\bm{x}_p\}_{p=1}^{n_1}\cup \{{\bm{y}}_q\}_{q=1}^{n_2}$ are pairwise distinct and the activation \(\sigma\) is \(\mathrm{ReLU}^3\) or smooth but non-polynomial, then the functions \(\left\{ \sum_{i=1}^{d} \sigma''(\bm{v}^\top \wt{\bm{x}}_p)v_{i}^2, ~ \sigma\left( \bm{v}^\top \wt{\bm{y}}_q \right)\right\}\) for \(p\in [n_1], q\in [n_2]\) are linearly independent and thus the Gram matrix \(\bm{G}_{\bm{a}}^{\infty} \) is positive definite.
\end{lemma}
\begin{proof}
By assumption, the extended points $\{\wt{\bm{x}}_p\}_{p=1}^{n_1}\cup \{\wt{\bm{y}}_q\}_{q=1}^{n_2}$ are pairwise non-parallel. If the linear independence fails, there exists some \(0\neq \bm{u} = [\alpha, \beta]^\top \in \R^{n_1+n_2} \) such that 
    \begin{equation}\label{Lin-Rel} 
        \sum_{p=1}^{n_1} \alpha_p
			\sum_{i=1}^{d} \sigma''(\bm{v}^\top \wt{\bm{x}}_p)v_{i}^2 + \sum_{q=1}^{n_2} \beta_q \sigma\left( \bm{v}^\top \wt{\bm{y}}_q \right) = 0, \quad \forall \bm{v}\in\R^{d+1}.
    \end{equation}
If \(\sigma = \mathrm{ReLU}^3\), then we define the hyperplanes 
$$ I_p := \left\{ \bm{v}\in\R^{d+1} : \bm{v}^\top \wt{\bm{x}}_p = 0 \right\} \quad \mbox{and}\quad J_q := \left\{ \bm{v}\in\R^{d+1} : \bm{v}^\top \wt{\bm{y}}_q = 0 \right\}.$$ 
Fix \(k\in[n_1]\). Then we can find \(\bm{z}\in I_k\setminus ((\cup_{p\neq k}I_p) \cup (\cup_{q}J_q))\) such that \(\sum_{i=1}^d z_i^2\neq 0\) since $\{\wt{\bm{x}}_p\}_{p=1}^{n_1}\cup \{\wt{\bm{y}}_q\}_{q=1}^{n_2}$ are pairwise non-parallel. Since the hyperplanes are closed, we can find \(r>0\) such that the ball \(B(\bm{z};r)\) centered at \(\bm{z}\) with radius \(r\) satisfies 
$$
B(\bm{z};r)\cap ((\cup_{p\neq k}I_q) \cup (\cup_{q}J_q)) = \varnothing.$$ 
Let 
$$B^+_r := \{\bm{v}\in B(\bm{z};r): \bm{v}^\top \wt{\bm{x}}_k >0\}\quad \mbox{and}\quad B^-_r := \{\bm{v}\in B(\bm{z};r): \bm{v}^\top \wt{\bm{x}}_k <0\}.
$$
Then by Lebesgue differentiation theorem, we have
    \begin{gather*}
        \lim_{r\to 0}\frac{1}{\mu(B^+_r)} \int_{B^+_r} \nabla_{\bm{v}} \sum_{i=1}^{d} \sigma''(\bm{v}^\top \wt{\bm{x}}_p)w_{i}^2 \,\d{\bm{v}} - \frac{1}{\mu(B^-_r)} \int_{B^-_r} \nabla_{\bm{v}} \sum_{i=1}^{d} \sigma''(\bm{v}^\top \wt{\bm{x}}_p)w_{i}^2 \,\d{\bm{v}} = \bm{0}, \quad  q\neq k, \\
        \lim_{r\to 0}\frac{1}{\mu(B^+_r)} \int_{B^+_r} \nabla_{\bm{v}} \sigma\left( \bm{v}^\top \wt{\bm{y}}_q \right) \,\d{\bm{v}} - \frac{1}{\mu(B^-_r)} \int_{B^-_r} \nabla_{\bm{v}}\sigma\left( \bm{v}^\top \wt{\bm{y}}_q \right) \,\d{\bm{v}} = \bm{0}, \quad q\in [n_2].
    \end{gather*}
    While for \(\wt{\bm{x}}_k\), we have 
    \begin{align*}
        & \lim_{r\to 0}\frac{1}{\mu(B^+_r)} \int_{B^+_r} \nabla_{\bm{v}} \sum_{i=1}^{d} \sigma''(\bm{v}^\top \wt{\bm{x}}_k)w_{i}^2 \,\d{\bm{v}} - \frac{1}{\mu(B^-_r)} \int_{B^-_r} \nabla_{\bm{v}} \sum_{i=1}^{d} \sigma''(\bm{v}^\top \wt{\bm{x}}_k)w_{i}^2 \,\d{\bm{v}} \\
        =  &\lim_{r\to 0}\frac{1}{\mu(B^+_r)} \int_{B^+_r} \nabla_{\bm{v}} \sum_{i=1}^{d} w_{i}^2 (\bm{v}^\top \wt{\bm{x}}_k)_+ \,\d{\bm{v}} = \wt{\bm{x}}_k \sum_{i=1}^{d}z_i^2 \neq \bm{0}.
    \end{align*}
    Therefore, by applying these operations to equation \eqref{Lin-Rel}, we deduce \(\alpha_k = 0\). Similarly, we have \(\bm{u} = [\alpha,\beta]^\top = \bm{0}\), which is a contradiction to the assumption $\bm{u}\neq0$. Next, if \(\sigma\) is smooth, we differentiate equation \eqref{Lin-Rel} with respect to \(\bm{v}\) for \(k\) times and obtain 
    \begin{align*}
        \sum_{p=1}^{n_1} \alpha_p & \left( \sigma^{(k+2)}(\bm{v}^\top \wt{\bm{x}}_p) \sum_{i\in[d]}v_{i}^2 \wt{\bm{x}}_p^{\otimes k} +  C^1_k \sigma^{(k+1)}(\bm{v}^\top \wt{\bm{x}}_p) \sum_{i\in[d]}2v_{i} \bm{e}_i \otimes \wt{\bm{x}}_p^{\otimes k-1} \right. \\
        &\left.+ C^2_k \sigma^{(k)}(\bm{v}^\top \wt{\bm{x}}_p) \sum_{i\in[d]}2 \bm{e}_i^{\otimes 2} \otimes \wt{\bm{x}}_p^{\otimes k-2} \right)  + \sum_{q=1}^{n_2} \beta_q \sigma^{(k)}\left( \bm{v}^\top \wt{\bm{y}}_q \right)\wt{\bm{y}}_q^{\otimes k} = 0, \quad \forall \bm{v}\in\R^{d+1},
    \end{align*}
    where \(\{\bm{e}_i:i=1,\ldots,d+1\}\) denotes the canonical Cartesian basis of \(\R^{d+1}\). Then using the linear independence of the tensors \(\left\{ \wt{\bm{x}}_1^{\otimes n_1+n_2-1},\cdots,\wt{\bm{x}}_{n_1}^{\otimes n_1+n_2-1},\wt{\bm{y}}_1^{\otimes n_1+n_2-1},\cdots,\wt{\bm{y}}_{n_2}^{\otimes n_1+n_2-1} \right\}\) (due to the assumption $\{\wt{\bm{x}}_p\}_{p=1}^{n_1}\cup \{\wt{\bm{y}}_q\}_{q=1}^{n_2}$ are pairwise non-parallel \cite[Lemma G.6]{Du2019deep}), one can derive that all the \(\alpha_p\) and \(\beta_q\) are zero. The lengthy argument is identical with that in \cite[Lemma 2]{Xu2024} and hence is omitted. 
\end{proof}

\subsection{Proof of Lemma \ref{wInit}}

Since \(\bm{w}_r\sim \mathcal{N}(\bm{0}, \bm{I}_{d+1})\) for each \(r\in [m]\), it suffices to control each of its entries.
For any \(M>0\), if the length of \( \bm{w}_r \) is not less than \(M\), then there exists
at least one of its components whose length is not less than \(\frac{M}{\sqrt{d+1}}\).
By the concentration inequality for the Gaussian random variable \(X\sim \mathcal{N}(0,\sigma^2)\) \cite[Lemma 1.3]{Rigollet2023}
\[ \mathbb P[|X| > t] \leq 2\exp \left( - \frac{t^2}{2\sigma^2} \right), \]
we get 
\begin{align*}
	\mathbb{P}\left(\|\boldsymbol{w}_r(0)\|_2 > M\right)\leq (d+1) \cdot 
	\mathbb{P}_{w \sim \mathcal{N}(0,1)}\left(|w| > \frac{M}{\sqrt{d+1}}\right)
	\leq 2(d+1) \exp\left(-\frac{1}{2} \cdot \frac{M^2}{d+1}\right).
\end{align*}
The assertion is obtained by setting the right hand side to \( \delta/m \).

\subsection{Proof of Lemma \ref{Loss0}}
For the interior loss $s_p(\bm{w},\bm{a})$, by the Cauchy-Schwarz inequality, we have
\begin{align*}
		\frac{1}{2}s_p(\bm{w},\bm{a})^2 & =\frac{1}{2n_1} \left( 
		\sum_{i=1}^{d} \frac{1}{\sqrt{m}} 
		\sum_{r=1}^{m} a_r \sigma''(\bm{w}_r^\top \wt{\bm{x}}_p)w_{ri}^2 -f(\bm{x}_p) \right)^2 \\
		& \leq \frac{1}{n_1} \left(  \left(  \frac{1}{\sqrt{m}} \sum_{r=1}^{m}\sum_{i=1}^{d}a_r 
		\sigma''(\bm{w}_r^\top \wt{\bm{x}}_p)w_{ri}^2 \right)^2 + f(\bm{x}_p)^2 \right).
	\end{align*}
Next we define the independent random variables 
	\[ X_r = \frac{1}{C_B} \sum_{i=1}^{d} a_r \sigma''(\bm{w}_r^\top \wt{\bm{x}}_p)w_{ri}^2 = \frac{1}{C_B}a_r \sigma''(\bm{w}_r^\top \wt{\bm{x}}_p) \sum_{i=1}^{d} w_{ri}^2,\quad r\in [m]. \]
Conditioned on the event \(E(B-1)\), by the local Lipschitz property of \(\sigma''\), we have
	\[ |X_r|\lesssim \|\bm{w}_r\|_2^3 + 1, \quad \E_\theta[X_r]=0, \quad r\in [m]. \]
Note that 
$$\|\bm{w}_r\|_2^3 = \left(\sum_{i=1}^{d+1} w_{ri}^2\right)^{\frac32} \lesssim \sqrt{d} \sum_{i=1}^{d+1} w_{ri}^3,$$
with each \(w_{ri}^3\) in the summation being sub-Weibull \((\frac{2}{3})\), since all the components \(w_{ri}\) are Gaussian. Thus for \(\alpha = \frac{2}{3}\), we have
\[ \Big\|\|\bm{w}_r\|_2^3\Big\|_{\psi_\alpha} \lesssim \sqrt{d} \left\|\sum_{i=1}^{d+1} w_{ri}^3\right\|_{\psi_\alpha} \lesssim  \sqrt{d}\sum_{i=1}^{d+1}\big\|w_{ri}^3\big\|_{\psi_\alpha}\lesssim d^\frac{3}{2}, \]
from which we derive
	\[\|X_r\|_{\psi_\alpha} \lesssim \Big\|\|\bm{w}_r\|_2^3\Big\|_{\psi_\alpha} +  \|1\|_{\psi_\alpha} \lesssim d^\frac{3}{2}, \quad \mbox{with }\alpha = \frac{2}{3}. \]
	We then have, by Lemma \ref{Concen}, with probability at least \(1-2\delta\),
	\[  \left| \frac{C_B}{\sqrt{m}} \sum_{r=1}^{m} X_r \right| \lesssim
	C_B d^\frac{3}{2}\log^{\frac{1}{2}}\left( \frac{1}{\delta} \right) + 
	\frac{C_B d^{3/2}}{\sqrt{m}} \log^{\frac{3}{2}}\left( \frac{1}{\delta} \right). \]
By discarding the constant term \(f(\bm{x}_p)^2\), we have
	\begin{align*}
		\frac{1}{2}s_p(\bm{w},\bm{a})^2 \lesssim \frac{C_B^2 d^3}{n_1} \left(  
		\log\left( \frac{1}{\delta} \right) + 
		\frac{1}{m} \log^3\left( \frac{1}{\delta} \right) \right).
	\end{align*}
	Likewise, for the boundary loss $h_q(\bm{w},\bm{a})$, we have
	\begin{align*}
		\frac{1}{2}h_q(\bm{w},\bm{a})^2 & =\frac{\gamma}{2n_2} \left( 
			\frac{1}{\sqrt{m}} \sum_{r=1}^{m} 
			a_r\sigma \left( \bm{w}_r^\top \wt{\bm{y}}_q \right)- 
			g(\wt{\bm{y}}_q)\right)^2 \\
		& \leq \frac{\gamma}{n_2} \left(  \left(  \frac{1}{\sqrt{m}} \sum_{r=1}^{m}
		a_r\sigma \left( \bm{w}_r^\top \wt{\bm{y}}_q \right) \right)^2 
		+ g(\wt{\bm{y}}_q)^2 \right).
	\end{align*}
With \(\alpha = 2 \), by Lemma \ref{Concen}, we have with probability at least \(1-2\delta\),
\begin{align*}
	\frac{1}{2}h_q(\bm{w},\bm{a})^2 \lesssim \frac{C_B^2 d}{n_2} \left(  
		\log\left( \frac{1}{\delta} \right) + 
		\frac{1}{m} \log\left( \frac{1}{\delta} \right) \right).
\end{align*}
Therefore, conditioned on the event \(E(B-1)\), we have with probability at least \(1-\delta\),
\begin{align*}
	L(0) \lesssim C_B^2 d^3\log\left( \frac{n_1+n_2}{\delta} \right) + \frac{C_B^2 d^3}{m} \log^3\left( \frac{n_1+n_2}{\delta} \right).
\end{align*}
Under the given assumption on the neural network width \(m\), the second term is smaller than the first term, and then the desired conclusion follows.

\subsection{Proof of Lemma \ref{GramInit}}
Since the spectral norm of a matrix is bounded by its Frobenius norm, it suffices to bound each entry of the difference matrix 
\[ 
\Delta \bm{G}_{\bm{w}}^\infty:= \bm{G}_{\bm{w}}(0) - \bm{G}_{\bm{w}}^\infty. 
\] 
There are three forms of its entries: 
\begin{equation*}
\Delta \bm{G}_{\bm{w}}^\infty[j,k] = \left\{\begin{aligned}
	    \sum_{r=1}^m \left\langle \frac{\partial s_j(\theta)}{\partial \bm{w}_r}, 
		\frac{\partial s_k(\theta)}{\partial \bm{w}_r} \right\rangle - 
		\mathbb{E}_{\theta} \left\langle \frac{\partial s_j(\theta)}{\partial \bm{w}_r}, 
		\frac{\partial s_k(\theta)}{\partial \bm{w}_r} 
		\right\rangle, &\quad 1\leq j \leq n_1,\; 1\leq k \leq n_1,\\
        \sum_{r=1}^m \left\langle \frac{\partial s_j(\theta)}{\partial \bm{w}_r}, 
		\frac{\partial h_k(\theta)}{\partial \bm{w}_r} \right\rangle - 
		\mathbb{E}_{\theta} \left\langle \frac{\partial s_j(\theta)}{\partial \bm{w}_r}, 
		\frac{\partial h_k(\theta)}{\partial \bm{w}_r} 
		\right\rangle, &\quad 1 \leq j \leq n_1 ,\; n_1 + 1\leq k \leq n_1 + n_2,\\
        \sum_{r=1}^m \left\langle \frac{\partial h_j(\theta)}{\partial \bm{w}_r}, 
		\frac{\partial h_k(\theta)}{\partial \bm{w}_r} \right\rangle - 
		\mathbb{E}_{\theta} \left\langle \frac{\partial h_j(\theta)}{\partial \bm{w}_r}, 
		\frac{\partial h_k(\theta)}{\partial \bm{w}_r} 
		\right\rangle, & \quad n_1 + 1\leq j,k \leq n_1 + n_2.
     \end{aligned}\right.
\end{equation*}
The case \(n_1 + 1\leq j \leq n_1 + n_2,\; 1\leq k \leq n_1\) is the same as the second case. We analyze the three cases of $\Delta \bm{G}_{\bm{w}}^\infty[j,k]$ separately. For any fixed pair \(1\leq j \leq n_1, 1\leq k \leq n_1\), we write the $(j,k)$-th entry $\bm{G}_{\bm{w}}^\infty[j,k]$ of the difference matrix \(\Delta \bm{G}_{\bm{w}}^\infty\) as 
	\[ \Delta \bm{G}_{\bm{w}}^\infty[j,k] = \frac{C_B^2}{n_1 m} \sum_{r=1}^{m} X_r, \]
where \(\{X_r\}_{r=1}^{m}\) are independent random variables. 
Conditioned on the event \(E(B-1)\), \(X_r\) satisfies
	\[ |X_r|\lesssim \|\bm{w}_r\|_2^6 + 1, \quad \E_\theta[X_r]=0, \quad r\in [m]. \]
Note that 
$$\|\bm{w}_r\|_2^6 = \left(\sum_{i=1}^{d+1} w_{ri}^2\right)^{3} \lesssim d^2 \sum_{i=1}^{d+1} w_{ri}^6$$ 
with each \(w_{ri}^6\) being sub-Weibull\((\frac{1}{3})\). Thus for \(\alpha = \frac{1}{3}\), we have
	\[\|X_r\|_{\psi_\alpha} \lesssim \Big\|\|\bm{w}_r\|_2^6\Big\|_{\psi_\alpha} + \|1\|_{\psi_\alpha}  \lesssim d^3, \quad r\in [m].\]
	Thus we have with probability at least \(1-2\delta\),
	\[  \Big| \Delta \bm{G}_{\bm{w}}^\infty[j,k] \Big| = 
	\left| \frac{C_B^2}{n_1 m} \sum_{r=1}^{m} X_r \right| \lesssim
	\frac{C_B^2 d^3}{n_1\sqrt{m}}\log^{\frac{1}{2}}\left( \frac{1}{\delta} \right) + 
	\frac{C_B^2 d^3}{n_1m} \log^3\left( \frac{1}{\delta} \right) .  \]
Similarly for \(1\leq j \leq n_1, n_1+1\leq k \leq n_1+n_2\) and 
	\(n_1 + 1\leq j \leq n_1 + n_2, 1\leq k \leq n_1\),
we have \(\alpha=\frac12\), with probability at least \(1-2\delta\),
\[  
\Big| \Delta \bm{G}_{\bm{w}}^\infty[j,k] \Big| = 
	\left| \frac{C_B^2}{\sqrt{n_1n_2} m} \sum_{r=1}^{m} X_r \right| \lesssim
	\frac{C_B^2 d^2}{\sqrt{n_1n_2}\sqrt{m}}\log^{\frac{1}{2}}\left( \frac{1}{\delta} \right) + 
	\frac{C_B^2 d^2}{\sqrt{n_1n_2}m}\log^2 \left( \frac{1}{\delta} \right).  
\]
For \(n_1 + 1\leq j,k \leq n_1 + n_2\), we have \(\alpha=1\), with probability at least \(1-2\delta\),
\[  
\Big| \Delta \bm{G}_{\bm{w}}^\infty[j,k] \Big| = 
	\left| \frac{C_B^2}{n_2 m} \sum_{r=1}^{m} X_r \right| \lesssim
	\frac{C_B^2 d}{n_2\sqrt{m}}\log^{\frac{1}{2}}\left( \frac{1}{\delta} \right) + 
	\frac{C_B^2 d}{n_2m}\log\left( \frac{1}{\delta} \right).  \]
By squaring and summing over these elements, we have with probability at least \(1-\delta\),
\[ \|\Delta \bm{G}_{\bm{w}}^\infty\|_F^2 \lesssim C_B^4 d^6\left( \frac{1}{m} \log\left( \frac{n_1+n_2}{\delta}  \right) + \frac{1}{m^2} \log^{6}\left( \frac{n_1+n_2}{\delta} \right) \right). \]
In order to bound \(\|\Delta \bm{G}_{\bm{w}}^\infty\|_F\) by \(\lambda_{\bm{w}}\), we require that the right hand side be smaller than \(\lambda_{\bm{w}}^2\),
from which we get 
	\[ m \gtrsim \frac{C_B^4d^6}{\lambda_{\bm{w}}^2} 
	\log^3\left(\frac{n_1+n_2}{\delta}\right). \]
Similarly, using the bound on the derivative with respect to \(a_r\) given in \eqref{hDeriBound}, to bound \(\|\Delta \bm{G}_{\bm{a}}^\infty\|_F\) by \(\lambda_{\bm{a}}\), we need 
	\[ m \gtrsim \frac{C_B^4d^6}{\lambda_{\bm{a}}^2} 
	\log^3\left(\frac{n_1+n_2}{\delta}\right). \]

\subsection{Proof of Lemma \ref{GramDyn}}
Conditioned on the event \(E(B-1)\), we have for all \(r\in [m]\),
\begin{align*}
		\|\bm{w}_r \|_2 &\leq  \|\bm{w}_r - \bm{w}_r(0)\|_2 + \|\bm{w}_r(0)\|_2 
		\leq 1 + \|\bm{w}_r(0)\|_2, \\
		\|a_r \|_2 &\leq  \|a_r - a_r(0)\|_2 + \|a_r(0)\|_2 \leq 2. 
	\end{align*}
	The derivatives of the interior loss $s_j$ have been computed in \eqref{int-loss}:
    \[ \frac{\partial s_p(\bm{w},\bm{a})}{\partial \bm{w}_r}  =  \frac{1}{\sqrt{n_1 m}} 
	 	\sum_{i=1}^{d} \left( a_r \sigma'''(\bm{w}_r^\top \wt{\bm{x}}_p)w_{ri}^2\wt{\bm{x}}_p + 
		a_r \sigma''(\bm{w}_r^\top \wt{\bm{x}}_p)2w_{ri}\bm{e}_i \right). \]
	Using the Lipschitz property of the activation function $\sigma$ and polynomials, we get
	\begin{align*}
		& \left\| \frac{\partial s_j(\bm{w},\bm{a})}{\partial \bm{w}_r} - 
		\frac{\partial s_j(\bm{w}(0),\bm{a}(0))}{\partial \bm{w}_r} \right\|_2 \\
		  \leq &\left\| \frac{\partial s_j(\bm{w},\bm{a})}{\partial \bm{w}_r} - 
		\frac{\partial s_j(\bm{w}(0),\bm{a})}{\partial \bm{w}_r} \right\|_2 +
		\left\| \frac{\partial s_j(\bm{w}(0),\bm{a})}{\partial \bm{w}_r} - 
		\frac{\partial s_j(\bm{w}(0),\bm{a}(0))}{\partial \bm{w}_r} \right\|_2 \\
		 \lesssim& \frac{C_B}{\sqrt{n_1 m}} (\|\bm{w}_r(0)\|_2^2+1) R_{\bm{w}}
		+ \frac{C_B}{\sqrt{n_1 m}} (\|\bm{w}_r(0)\|_2^3+1) R_{\bm{a}},
	\end{align*}
where the last step follows from the following inequalities
\begin{align*}
 &   \begin{aligned}
       & \left|\sum_{i=1}^{d}\sigma'''(\bm{w}_r^\top \wt{\bm{x}}_p)w_{ri}^2 - \sum_{i=1}^{d}\sigma'''(\bm{w}_r(0)^\top \wt{\bm{x}}_p)w_{ri}(0)^2\right| \\
       \leq& 
         \left|\sigma'''(\bm{w}_r^\top \wt{\bm{x}}_p)\sum_{i=1}^{d}w_{ri}^2 - \sigma'''(\bm{w}_r(0)^\top \wt{\bm{x}}_p)\sum_{i=1}^{d}w_{ri}^2\right| \\
         &+\left|\sigma'''(\bm{w}_r(0)^\top \wt{\bm{x}}_p)\sum_{i=1}^{d}w_{ri}^2 - \sigma'''(\bm{w}_r(0)^\top \wt{\bm{x}}_p)\sum_{i=1}^{d}w_{ri}(0)^2\right| \\
      \lesssim& C_B  R_{\bm{w}} \cdot (\|\bm{w}_r(0)\|_2^2+1) + C_B (\|\bm{w}_r(0)\|_2+1) \cdot (\|\bm{w}_r(0)\|_2+1)R_{\bm{w}} \approx C_B (\|\bm{w}_r(0)\|_2^2+1) R_{\bm{w}},
    \end{aligned}\\
  & \left|  \sum_{i=1}^{d}  a_r \sigma'''(\bm{w}_r(0)^\top \wt{\bm{x}}_p)w_{ri}^2(0)-
    \sum_{i=1}^{d}  a_r(0) \sigma'''(\bm{w}_r(0)^\top \wt{\bm{x}}_p)w_{ri}^2(0)\right|\lesssim C_B (\|\bm{w}_r(0)\|_2^3+1) R_{\bm{a}}.
\end{align*}
The difference for the boundary loss $h_j$ is given by
\begin{align*}
	& \left\| \frac{\partial h_j(\bm{w},\bm{a})}{\partial \bm{w}_r} - 
		\frac{\partial h_j(\bm{w}(0),\bm{a}(0))}{\partial \bm{w}_r} \right\|_2 \\
		&  \leq \left\| \frac{\partial h_j(\bm{w},\bm{a})}{\partial \bm{w}_r} - 
		\frac{\partial h_j(\bm{w}(0),\bm{a})}{\partial \bm{w}_r} \right\|_2 +
		\left\| \frac{\partial h_j(\bm{w}(0),\bm{a})}{\partial \bm{w}_r} - 
		\frac{\partial h_j(\bm{w}(0),\bm{a}(0))}{\partial \bm{w}_r} \right\|_2 \\
		& \lesssim \frac{C_B}{\sqrt{n_2 m}} R_{\bm{w}} 
		+ \frac{C_B}{\sqrt{n_2 m}}(\|\bm{w}_r(0)\|_2+1) R_{\bm{a}}.
	\end{align*}
The entries of the difference matrix \(\Delta \bm{G}_{\bm{w}}:= 
	\bm{G}_{\bm{w}}(\bm{w},\bm{a}) - \bm{G}_{\bm{w}}(\bm{w}(0),\bm{a}(0))\) take three forms.
	For any fixed pair \(1\leq j \leq n_1, 1\leq k \leq n_1\), using the estimate \eqref{sDeriBound}, 
        we have
	\begin{align*}
		\Big|\Delta \bm{G}_{\bm{w}}[j,k]\Big| 
		 =& \left| \sum_{r=1}^{m} 
		\left\langle \frac{\partial s_j(\bm{w},\bm{a})}{\partial \bm{w}_r}, 
		\frac{\partial s_k(\bm{w},\bm{a})}{\partial \bm{w}_r} \right\rangle - 
		\left\langle \frac{\partial s_j(\bm{w}(0),\bm{a}(0))}{\partial \bm{w}_r}, 
		\frac{\partial s_k(\bm{w}(0),\bm{a}(0))}{\partial \bm{w}_r} \right\rangle \right| \\
		 \leq& \left| \sum_{r=1}^{m} 
		\left\langle \frac{\partial s_j(\bm{w},\bm{a})}{\partial \bm{w}_r}, 
		\frac{\partial s_k(\bm{w},\bm{a})}{\partial \bm{w}_r} - 
		\frac{\partial s_k(\bm{w}(0),\bm{a}(0))}{\partial \bm{w}_r} \right\rangle \right|  \\
		& + \left| \sum_{r=1}^{m} 
		\left\langle \frac{\partial s_j(\bm{w},\bm{a})}{\partial \bm{w}_r} 
		- \frac{\partial s_j(\bm{w}(0),\bm{a}(0))}{\partial \bm{w}_r}, 
		\frac{\partial s_k(\bm{w}(0),\bm{a}(0))}{\partial \bm{w}_r} \right\rangle \right| \\
		 \lesssim& \frac{C_B^2}{n_1 m} \sum_{r=1}^{m} (\|\bm{w}_r(0)\|_2^5+1) R_{\bm{w}} +
		\frac{C_B^2}{n_1 m} \sum_{r=1}^{m} (\|\bm{w}_r(0)\|_2^6+1) R_{\bm{a}}  \\
        \leq &\frac{C_B^2}{n_1 }(R_{\bm{w}} + R_{\bm{a}})\frac{1}{m} \sum_{r=1}^{m}(\|\bm{w}_r(0)\|_2^6 +1) .
	\end{align*}
	Meanwhile,
	for \(1\leq j \leq n_1, n_1+1\leq k \leq n_1+n_2\) and 
	\(n_1 + 1\leq j \leq n_1 + n_2, 1\leq k \leq n_1\), we have
	\begin{align*}
		\Big|\Delta \bm{G}_{\bm{w}}[j,k]\Big| 
		\lesssim \frac{C_B^2}{\sqrt{n_1n_2}}(R_{\bm{w}} + R_{\bm{a}})\frac{1}{m}\sum_{r=1}^{m}(\|\bm{w}_r(0)\|_2^4+1).
	\end{align*}
	For \(n_1 + 1\leq j,k \leq n_1 + n_2\), we have
	\begin{align*}
		\Big|\Delta \bm{G}_{\bm{w}}[j,k]\Big| 
		\lesssim \frac{C_B^2}{\sqrt{n_1n_2}}(R_{\bm{w}} + R_{\bm{a}})\frac{1}{m}\sum_{r=1}^{m}(\|\bm{w}_r(0)\|_2^2+1).
	\end{align*}
By combining the preceding estimates, we deduce
	\begin{align*}
		\|\Delta \bm{G}_{\bm{w}}\|_2 & \leq \|\Delta \bm{G}_{\bm{w}}\|_F \lesssim
		C_B^2 (R_{\bm{w}} + R_{\bm{a}})\frac{1}{m}\sum_{r=1}^{m}(\|\bm{w}_r(0)\|_2^6+1) \\
        & \lesssim C_B^2(R_{\bm{w}} + R_{\bm{a}})\left(\frac{1}{m}\sum_{r=1}^{m}(\|\bm{w}_r(0)\|_2^6-\E\left[\|\bm{w}_r(0)\|_2^6\right]) + d^3 +1\right) .
	\end{align*}
With \(\alpha = \frac{1}{3} \), by Lemma \ref{Concen}, we have with probability at least \(1-2\delta\)
\begin{equation}\label{moment-w(0)}
    \frac{1}{m}\sum_{r=1}^{m}(\|\bm{w}_r(0)\|_2^6-\E\left[\|\bm{w}_r(0)\|_2^6\right]) \lesssim 
\frac{d^3}{\sqrt{m}}\log^{\frac{1}{2}}\left( \frac{1}{\delta} \right) + 
	\frac{d^3}{m}\log^3\left( \frac{1}{\delta} \right).
\end{equation}
Conditioned on the events in Lemma \ref{GramInit}, we have \(\|\Delta \bm{G}_{\bm{w}}\|_2\leq \frac{\lambda_{\bm{w}}}{4}\). Similar estimates hold for \(\bm{G}_{\bm{a}}\), since we have 
\begin{align*}
		 \left\| \frac{\partial s_j(\bm{w},\bm{a})}{\partial a_r} - 
		\frac{\partial s_j(\bm{w}(0),\bm{a}(0))}{\partial a_r} \right\|_2 
		 &= \left\| \frac{\partial s_j(\bm{w})}{\partial a_r} - 
		\frac{\partial s_j(\bm{w}(0))}{\partial a_r} \right\|_2 
		 \lesssim \frac{C_B}{\sqrt{n_1 m}} R_{\bm{w}}(\|\bm{w}_r(0)\|_2^2+1),\\
		 \left\| \frac{\partial h_j(\bm{w},\bm{a})}{\partial a_r} - 
		\frac{\partial h_j(\bm{w}(0),\bm{a}(0))}{\partial a_r} \right\|_2 
		 &= \left\| \frac{\partial h_j(\bm{w})}{\partial a_r} - 
		\frac{\partial h_j(\bm{w}(0))}{\partial a_r} \right\|_2 
		 \lesssim \frac{C_B}{\sqrt{n_2 m}} R_{\bm{w}}.
	\end{align*}
Consequently, we deduce
	\begin{align*}
		\|\bm{G}_{\bm{a}}(\bm{w},\bm{a}) - \bm{G}_{\bm{a}}(\bm{w}(0),\bm{a}(0))\|_2 
		\leq \|\Delta \bm{G}_{\bm{a}}\|_F 
		\lesssim C_B^2 R_{\bm{w}}\left(\frac{1}{m}\sum_{r=1}^{m}\|\bm{w}_r(0)\|_2^4+1\right) .
	\end{align*}
Combining the preceding estimates completes the proof of the lemma.

\section{Proofs for section \ref{SecSGD}} \label{Pro-SecSGD}

\subsection{Proof of Lemma \ref{StepGap-w_t}}
First we recall the vector form of the Cauchy-Schwarz inequality:
\begin{equation}\label{eqn:CS-vector}
   \left\| \sum_{i=1}^na_v\boldsymbol{v}_i\right\|_2^2\leq \left(\sum_{i=1}^na_i^2\right)\left(\sum_{i=1}^n \|\boldsymbol{v}_i\|_2^2\right), \quad \forall \{a_i\}_{i=1}^n\subset\mathbb{R}, \{\boldsymbol{v}_i\}_{i=1}^n\subset \mathbb{R}^d.
\end{equation}
It can be derived by applying the standard scalar form of the inequality to each component of the vector.
Then according to the iteration process, we have
	\begin{equation*}
		\|\bm{w}_*(t+1) - \bm{w}_*(t)\|_2 = \eta 
		\left\| \frac{\partial \wt{L}(\bm{w}(t), \bm{a}(t))}{\partial \bm{w}_*} \right\|_2
		= \eta \left\| \frac{\partial \wt{L}(t)}{\partial \bm{w}_*} \right\|_2.
	\end{equation*}
	Conditioned on the event \(t<S\), we take conditional expectation
    \begin{equation}\label{eli-BatSize} 
     \begin{aligned}
		&\quad \mathbb{E} \left[ \left\| \frac{\partial \wt{L}(\bm{w}_t, \bm{a}_t)}{\partial \bm{w}_*} 
		\right\|_2^2 \cdot \mathbf{1}_{S> t} \mid \mathcal{F}_t \right] \\
		& = \mathbb{E} \left[ \left\| \sum_{p\in \mathcal{I}} \frac{n_1}{I} s_p(\bm{w}_t, \bm{a}_t) \cdot 
		\frac{\partial s_p(\bm{w}_t, \bm{a}_t)}{\partial \bm{w}_*} +
		 \sum_{q\in \mathcal{J}} \frac{ n_2}{J} h_q(\bm{w}_t, \bm{a}_t) \cdot 
		\frac{\partial h_q(\bm{w}_t, \bm{a}_t)}{\partial \bm{w}_*}
		\right\|_2^2 \cdot \mathbf{1}_{S> t} \mid \mathcal{F}_t \right] \\
		& \leq \mathbb{E} \left[ \left( \sum_{p\in \mathcal{I}} \frac{n_1}{I} s_p(\bm{w}_t, \bm{a}_t)^2 +
		 \sum_{q\in \mathcal{J}} \frac{ n_2}{J} h_q(\bm{w}_t, \bm{a}_t)^2
		\right) \cdot \right. \\
		& \qquad \qquad \qquad 
		\left. \left( \sum_{p\in \mathcal{I}} \frac{n_1}{I} 
		\left\|\frac{\partial s_p(\bm{w}_t, \bm{a}_t)}{\partial \bm{w}_*}\right\|_2^2 +
		\sum_{q\in \mathcal{J}} \frac{ n_2}{J} 
		\left\|\frac{\partial h_q(\bm{w}_t, \bm{a}_t)}{\partial \bm{w}_*}\right\|_2^2
	    \right) \cdot \mathbf{1}_{S> t} \mid \mathcal{F}_t \right] \\
		& \lesssim   \mathbb{E} \left[ \left( \sum_{p\in \mathcal{I}} \frac{n_1}{I} s_p(\bm{w}_t, \bm{a}_t)^2 +
		\sum_{q\in \mathcal{J}} \frac{ n_2}{J} h_q(\bm{w}_t, \bm{a}_t)^2
	    \right) \cdot \right. \\
		& \qquad \qquad \qquad 
		\left. \left(I \cdot \frac{n_1}{I} \frac{C_B^2 B^6}{n_1 m} 
		+ J\cdot \frac{ n_2}{J} \frac{C_B^2 B^2}{n_2 m}   \right)  
		\cdot \mathbf{1}_{S> t} \mid \mathcal{F}_t \right]\\
		& \lesssim \frac{C_B^2 B^6}{m}L(t),
	\end{aligned}   
    \end{equation}
where we have used the Cauchy-Schwarz inequality \eqref{eqn:CS-vector} in the third line and the estimates (\ref{sDeriBound}) and (\ref{hDeriBound}) in the fourth line. Similarly, we have 
\begin{align*}
\mathbb{E} \left[ \left\|\frac{\partial \wt{L}(\bm{w}, \bm{a})}{\partial a_*} \right\|_2^2 
\cdot \mathbf{1}_{S> t} \mid \mathcal{F}_t \right] \lesssim \frac{C_B^2 B^6}{m}L(t).
\end{align*}

\begin{remark}\label{rmk:batch_size}
In the derivation of the crucial estimate \eqref{eli-BatSize},  we have split the factors \(\frac{n_1}{I}\) and \(\frac{n_2}{J}\) equally into the two terms. Note that in the second factor, the batch sizes cancel out due to the uniform upper bound of the terms \(\frac{\partial s_p}{\partial \bm{w}_*}\) and \(\frac{\partial h_q}{\partial \bm{w}_*}\), while in the first part we have obtained the loss \(L(t)\) upon taking expectation, due to the unbiasedness of the stochastic loss $\widetilde{L}$. Consequently, the effect of the batch sizes $I$ and $J$ is eliminated. In contrast, using the the slightly looser estimate 
    \begin{align*}
         \mathbb{E} \left[ \left\| \frac{\partial \wt{L}}{\partial \bm{w}_*} 
		\right\|_2^2 \right] & = \mathbb{E} \left[ \left\| \sum_{p\in \mathcal{I}} \frac{n_1}{I} s_p \cdot 
		\frac{\partial s_p}{\partial \bm{w}_*} +
		 \sum_{q\in \mathcal{J}} \frac{ n_2}{J} h_q \cdot 
		\frac{\partial h_q}{\partial \bm{w}_*}
		\right\|_2^2  \right] \\
		& \leq  \max\left\{ \frac{n_1}{I}, \frac{n_2}{J} \right\} \left\| \left[ 
		\frac{\partial s_1}{\partial \bm{w}_*}, \cdots, \frac{\partial h_{n_2}}{\partial \bm{w}_*}  \right] \right\|_2^2
        \mathbb{E} \left[  \sum_{p\in \mathcal{I}} \frac{n_1}{I} s_p^2 +
		 \sum_{q\in \mathcal{J}} \frac{ n_2}{J} h_q^2
		 \right],
    \end{align*}
 the factor \(\max\left\{ \frac{n_1}{I}, \frac{n_2}{J} \right\}\) would appear, and this is the case in the prior works \cite{Allen2019,GaoTan:2025KAN}.
\end{remark}

\subsection{Proof of Lemma \ref{LossGap}}

The proof of the lemma requires the following results.
\begin{lemma}\label{CondiExpe} 
Let \((\Omega,\mathcal{A},\mathbb{P})\) be a probability space and 
\(\mathcal{F}\subset \mathcal{A}\) be a sub-\(\sigma\)-algebra. Let \(X\) be a random variable with \(\E\left[X^2\right]< \infty\). If \(X=0 \; a.e.\)  on \(A^c\) for some \(A\in \mathcal{F}\), then
the conditional expectation \( Y:= \E\left[X \mid \mathcal{F} \right] \)
also vanishes \(a.e.\) on \(A^c\).
\end{lemma}
\begin{proof}
Let \(\wt{Y} = Y\mathbf{1}_{A} \). Since \(A\in \mathcal{F}\) and \(Y\) is \(\mathcal{F}\)-measurable by the definition of conditional expectation,  \(\wt{Y}\) is \(\mathcal{F}\)-measurable. By the property of conditional expectation,
\begin{align*}
\E\left[(X-Y)^2\right]  
\leq \E\left[(X-\wt{Y})^2\right] = \E\left[(X-Y)^2\cdot \mathbf{1}_{A}\right].
\end{align*}
Meanwhile, we have
\begin{align*}
\E\left[(X-Y)^2\right] 
& = \E\left[(X-Y)^2\cdot \mathbf{1}_{A}\right] + 
\E\left[(X-Y)^2\cdot \mathbf{1}_{A^c}\right] \\
& = \E\left[(X-Y)^2\cdot \mathbf{1}_{A}\right] + \E\left[Y^2\cdot \mathbf{1}_{A^c}\right].
\end{align*}
Combining these identities yields \(Y^2\cdot \mathbf{1}_{A^c} \equiv 0\) almost everywhere. 
\end{proof}

\begin{lemma}\label{Loss(t)}
    If the event \(\init\) happens and \(m\gtrsim \log^3\left( \frac{n_1+n_2}{\delta}\right)\), then for any \(t<S\), we have \( L(t)\lesssim L(0)\).
\end{lemma}
\begin{proof}
    Similar to the proof of Lemma \ref{GramDyn}, we have
    \begin{align*}
	& \left|\sum_{i=1}^{d} a_r(t) \sigma''(\bm{w}_r(t)^\top \wt{\bm{x}}_p)w_{ri}(t)^2
	- a_r(0) \sigma''(\bm{w}_r(0)^\top \wt{\bm{x}}_p)w_{ri}(0)^2\right| \\
	 \leq& \left|a_r(t)\left(\sum_{i=1}^{d}\sigma''(\bm{w}_r(t)^\top \wt{\bm{x}}_p)w_{ri}(t)^2 
	- \sigma''(\bm{w}_r(0)^\top \wt{\bm{x}}_p)w_{ri}(0)^2 \right)\right| \\
	& + \left| (a_r(t)-a_r(0))a_r(0) \sum_{i=1}^{d}\sigma''(\bm{w}_r(0)^\top \wt{\bm{x}}_p)w_{ri}(0)^2 \right|\\
	 \lesssim& 2 C_B (\|\bm{w}_r(0)\|_2^2+1) R_{\bm{w}} + C_B (\|\bm{w}_r(0)\|_2^3+1) R_{\bm{a}} \\
    \lesssim &C_B  (R_{\bm{w}}+R_{\bm{a}})(\|\bm{w}_r(0)\|_2^3+1).
\end{align*}
Then, the difference of the interior loss $s_p(\bm{w},\bm{a})$ is given by
\begin{align*}
	& \left|\frac{1}{2}s_p(\bm{w}(t),\bm{a}(t))^2 - \frac{1}{2}s_p(\bm{w}(0),\bm{a}(0))^2 \right| \\
	 =& \frac{1}{2n_1} \left| \frac{1}{\sqrt{m}} \sum_{r=1}^{m} \sum_{i=1}^{d} 
	 a_r(t) \sigma''(\bm{w}_r(t)^\top \wt{\bm{x}}_p)w_{ri}(t)^2
	 - a_r(0) \sigma''(\bm{w}_r(0)^\top \wt{\bm{x}}_p)w_{ri}(0)^2  \right|\times \\
	&  \left| \frac{1}{\sqrt{m}} \sum_{r=1}^{m} \sum_{i=1}^{d} 
	 a_r(t) \sigma''(\bm{w}_r(t)^\top \wt{\bm{x}}_p)w_{ri}(t)^2
	+ a_r(0) \sigma''(\bm{w}_r(0)^\top \wt{\bm{x}}_p)w_{ri}(0)^2 - 2f(\bm{x}_p) \right| \\
 \lesssim & \frac{1}{2n_1} C_B (R_{\bm{w}}+R_{\bm{a}}) \frac{1}{\sqrt{m}}\sum_{r=1}^{m} (\|\bm{w}_r(0)\|_2^3+1)
	\left( C_B (R_{\bm{w}}+R_{\bm{a}}) \frac{1}{\sqrt{m}}\sum_{r=1}^{m} (\|\bm{w}_r(0)\|_2^3+1) + 2 |s_p(0)|  \right).
\end{align*}
The same inequality holds for the boundary loss $h_q(\bm{w},\bm{a})$. Since the event \(\init\) happens, by the inequality \eqref{moment-w(0)}, 
we have \[\frac{1}{\sqrt{m}}\sum_{r=1}^{m} (\|\bm{w}_r(0)\|_2^3+1)\lesssim \frac{d^{\frac32}}{\sqrt[4]{m}}
\log^{\frac{1}{4}}\left( \frac{1}{\delta} \right) + 
	\frac{d^{\frac32}}{\sqrt{m}}\log^{\frac32}\left( \frac{1}{\delta} \right) + d^{\frac32} \lesssim d^{\frac32}. \] 
Thus the estimate 
\(L(t) \lesssim L(0)\) holds by equation \eqref{RwRa}.
\end{proof}

In the following discussion, we always condition on the event \(t+1<S\). We may omit the 
function \(\mathbf{1}_{S>t+1}\) or \(\mathbf{1}_{S>t}\) whenever there is no ambiguity. For \(\tau\in[0,1]\), we define the linearly interpolated weight
	\[ \theta_{t+\tau} := \theta_t - \tau\eta \nabla \wt{L}(\theta_t) = 
	\tau\theta_{t+1} + (1-\tau)\theta_{t}.  \] 
	Note that \(\|\bm{w}_r(t+1)\|_2\leq B, \|\bm{w}_r(t)\|_2\leq B\) and 
	\(\|a_r(t+1)\|_r\leq 2, \|a_r(t)\|_2\leq 2\). Then, by the convexity of norms, we also have
	\[\|\bm{w}_r(t+\tau)\|_2\leq B, \quad \|a_r(t+\tau)\|_2\leq 2, \] 
	for all \(\tau\in[0,1]\), where \(\bm{w}_r(t+\tau)\) and \(a_r(t+\tau)\) 
	are the corresponding components of \(\theta_{t+\tau}\).

	First, \(L(t+1) \cdot \mathbf{1}_{S> t+1}\) is square integrable by Lemma \ref{Loss(t)}. Note that the characteristic function \( \mathbf{1}_{S> t}\) is \(\FF_t\)-measurable while \( \mathbf{1}_{S> t+1} \) may be not. Then by Lemma \ref{CondiExpe}, we have
	\begin{align*}
		\mathbb{E} \left[ L(t+1) \cdot \mathbf{1}_{S> t+1} \mid \mathcal{F}_t \right]
		\leq \mathbb{E} \left[ L(t+1) \cdot \mathbf{1}_{S> t} \mid \mathcal{F}_t \right]
		= \mathbb{E} \left[ L(t+1) \cdot \mathbf{1}_{S> t} \mid \mathcal{F}_t \right]
		\cdot \mathbf{1}_{S> t}.
	\end{align*}
Second, consider the difference between the interior loss $s_p(\theta)$ at two successive steps:
	\begin{align*}
		   s_p(\theta_{t+1})  - s_p(\theta_t) 
		& = \int_{0}^{1} \frac{\d}{\d\tau} s_p(\theta_t - \tau\eta \nabla \wt{L}(\theta_t)) \d\tau \\
		& = -\eta \int_{0}^{1} \langle \nabla s_p(\theta_t - \tau\eta \nabla \wt{L}(\theta_t)),
		\nabla \wt{L}(\theta_t) \rangle  \d\tau \\
		& = -\eta \int_{0}^{1} \langle \nabla s_p(\theta_t),\nabla \wt{L}(\theta_t) \rangle  \d\tau 
		-\eta \int_{0}^{1} \langle \nabla s_p(\theta_{t+\tau})-\nabla s_p(\theta_{t}),
		\nabla \wt{L}(\theta_t) \rangle  \d\tau.
	\end{align*}
By taking conditional expectation directly, we obtain
\begin{align*}
		\mathbb{E} \left[ \langle \nabla s_p(\theta_t),\nabla \wt{L}(\theta_t) \rangle
		\mid \mathcal{F}_t \right] 
		= \langle \nabla s_p(\theta_t),\nabla {L}(\theta_t) \rangle.
	\end{align*}
Meanwhile, using the argument in Lemma \ref{GramDyn}, we have
	\begin{align*}
		&  \llan \frac{\partial s_p(\bm{w}_{t+\tau},\bm{a}_{t+\tau})}
		{\partial \bm{w}_r} - \frac{\partial s_p(\bm{w}_t,\bm{a}_t)}{\partial \bm{w}_r},
		\frac{\partial \wt{L}(t)}{\partial \bm{w}_r} \rran \\
 \lesssim  &\left\| \frac{\partial \wt{L}(\bm{w}, \bm{a})}{\partial \bm{w}_r} \right\|_2 \cdot \frac{C_B B^3}{\sqrt{n_1 m}}\left( \|\bm{w}_r(t+\tau) - \bm{w}_r(t)\|_2 + \|a_r(t+\tau) - a_r(t)\|_2 \right) \\
 =& \frac{\tau \eta C_B B^3}{\sqrt{n_1 m}}   \left\| \frac{\partial \wt{L}(\bm{w}, \bm{a})}{\partial \bm{w}_r} \right\|_2 \left( \left\| \frac{\partial \wt{L}(\bm{w}, \bm{a})}{\partial \bm{w}_r} \right\|_2 + \left\| \frac{\partial \wt{L}(\bm{w}, \bm{a})}{\partial a_r} \right\|_2 \right)  \\
 \lesssim & \frac{\tau \eta C_B B^3}{\sqrt{n_1 m}}  \left( \left\| \frac{\partial \wt{L}(\bm{w}, \bm{a})}{\partial \bm{w}_r} \right\|_2^2 + \left\| \frac{\partial \wt{L}(\bm{w}, \bm{a})}{\partial a_r} \right\|_2^2 \right).
\end{align*}
Similarly, 
\begin{align*}
 \llan \frac{\partial s_p(\bm{w}_{t+\tau},\bm{a}_{t+\tau})}
		{\partial a_r} - \frac{\partial s_p(\bm{w}_t,\bm{a}_t)}{\partial a_r},
		\frac{\partial \wt{L}(t)}{\partial a_r} \rran 
		\lesssim  \frac{\tau \eta C_B B^3}{\sqrt{n_1 m}}  
		\left( \left\| \frac{\partial \wt{L}(\bm{w}, \bm{a})}{\partial \bm{w}_r} \right\|_2^2
		+ \left\| \frac{\partial \wt{L}(\bm{w}, \bm{a})}{\partial a_r} \right\|_2^2 \right).
	\end{align*}
By taking conditional expectation and summing over \(r\in [m]\), we get
	\begin{align*}
		\mathbb{E} \left[ \langle \nabla s_p(\theta_{t+\tau})-\nabla s_p(\theta_{t}),
		\nabla \wt{L}(\theta_t) \rangle 
		\cdot \mathbf{1}_{S> t+1} \mid \mathcal{F}_t \right] 
		\lesssim \tau\eta\frac{C_B^3 B^9}{\sqrt{n_1 m}}L(t),
	\end{align*}
	where we have used the following estimates from the proof of Lemma \ref{StepGap-w_t}
	\begin{align*}
		\mathbb{E} \left[ \left\|\frac{\partial \wt{L}(\bm{w}, \bm{a})}{\partial w_r} \right\|_2^2 
		\cdot \mathbf{1}_{S> t} \mid \mathcal{F}_t \right],\;
		\mathbb{E} \left[ \left\|\frac{\partial \wt{L}(\bm{w}, \bm{a})}{\partial a_r} \right\|_2^2 
		\cdot \mathbf{1}_{S> t} \mid \mathcal{F}_t \right] \lesssim \frac{C_B^2 B^6}{m}L(t).
	\end{align*}
Likewise, for the boundary loss $h_q$, we have
	\begin{align*}
		   h_q(\theta_{t+1})  - h_q(\theta_t) 
		& = \int_{0}^{1} \frac{d}{d\tau} h_q(\theta_t - \tau\eta \nabla \wt{L}(\theta_t)) \d\tau \\
		& = -\eta \int_{0}^{1} \langle \nabla h_q(\theta_t - \tau\eta \nabla \wt{L}(\theta_t)),
		\nabla \wt{L}(\theta_t) \rangle  \d\tau \\
		& = -\eta \int_{0}^{1} \langle \nabla h_q(\theta_t),\nabla \wt{L}(\theta_t) \rangle  \d\tau 
		-\eta \int_{0}^{1} \langle \nabla h_q(\theta_{t+\tau})-\nabla h_q(\theta_{t}),
		\nabla \wt{L}(\theta_t) \rangle  \d\tau.
	\end{align*}
	and the following estimates
	\begin{gather*}
		\mathbb{E} \left[ \langle \nabla h_q(\theta_t),\nabla \wt{L}(\theta_t) \rangle
		\mid \mathcal{F}_t \right] 
		= \langle \nabla h_q(\theta_t),\nabla {L}(\theta_t) \rangle,\\
		\mathbb{E} \left[ \langle \nabla h_q(\theta_{t+\tau})-\nabla h_q(\theta_{t}),
		\nabla \wt{L}(\theta_t) \rangle 
		\cdot \mathbf{1}_{S> t+1} \mid \mathcal{F}_t \right] 
		\lesssim \tau\eta\frac{C_B^3 B^7}{\sqrt{n_2 m}}L(t).
	\end{gather*}
Combining the preceding estimates gives
	\begin{align*}
		\mathbb{E} \left[ \begin{bmatrix}
			\bm{s}(\theta_{t+1}) \\
			\bm{h}(\theta_{t+1})
		\end{bmatrix} -  \begin{bmatrix}
			\bm{s}(\theta_{t}) \\
			\bm{h}(\theta_{t})
		\end{bmatrix}  
		\cdot \mathbf{1}_{S> t+1} \mid \mathcal{F}_t \right] 
		= - \eta \bm{G}_{\theta}(t)  \begin{bmatrix}
			\bm{s}(\theta_{t}) \\
			\bm{h}(\theta_{t})
		\end{bmatrix} + \begin{bmatrix}
			\chi_{\bm{s}}(\theta_{t}) \\
			\chi_{\bm{h}}(\theta_{t})
		\end{bmatrix},
	\end{align*}
	where for some constant \(c_1>0\), the residual satisfies
	\begin{align*}
		\left\| \begin{bmatrix}
			\chi_{\bm{s}}(\theta_{t}) \\
			\chi_{\bm{h}}(\theta_{t})
		\end{bmatrix} \right\|_2 \leq  c_1 \eta^2 \frac{C_B^3 B^9}{\sqrt{m}}L(t).
	\end{align*}
    Moreover, recall that 
	\begin{gather*}
		\begin{aligned}
			s_p(\bm{w},\bm{a})  = \frac{1}{\sqrt{n_1 m}} \left( 
			\sum_{i=1}^{d} 
			\sum_{r=1}^{m} a_r \sigma''(\bm{w}_r^\top \wt{\bm{x}}_p)w_{ri}^2 - \sqrt{m} f(\bm{x}_p) \right)
		\end{aligned}.
	\end{gather*}
Using the local Lipschitz property of the activation function $\sigma$ and polynomials, we get
\begin{align*}
\| s_p(t+1) - s_p(t) \|_2 & \lesssim \sum_{r=1}^{m} \frac{C_B B^3}{\sqrt{n_1 m}}
\left( \|\bm{w}_r(t+1) - \bm{w}_r(t)\|_2 + \|a_r(t+1) - a_r(t)\|_2 \right) \\
& = \frac{\eta C_B B^3}{\sqrt{n_1 m}} \sum_{r=1}^{m}  
		\left( \left\| \frac{\partial \wt{L}(\bm{w}, \bm{a})}{\partial \bm{w}_r} \right\|_2
		+ \left\| \frac{\partial \wt{L}(\bm{w}, \bm{a})}{\partial a_r} \right\|_2 \right).
	\end{align*}
	Squaring both sides and then taking the conditional expectation yield
	\begin{align*}
		\mathbb{E} \left[ \| s_p(t+1) - s_p(t) \|_2 ^2 \mid \mathcal{F}_t \right]
	    \lesssim  \eta^2 \frac{C_B^4 B^{12}}{n_1 m} L(t).
   \end{align*}
   A similar estimate holds for the boundary loss $h_q$. Then we obtain 
   for some \(c_2>0\),
	\begin{align*}
		\mathbb{E} \left[ \left\| \begin{bmatrix} \bm{s}(t+1) \\
			\bm{h}(t+1) \end{bmatrix} -
		\begin{bmatrix} \bm{s}(t) \\ \bm{h}(t) \end{bmatrix}  \right\|_2^2 
		\cdot \mathbf{1}_{S> t} \mid \mathcal{F}_t \right]
		\leq  c_2 \frac{\eta^2}{m} C_B^4 B^{12}L(t).
	\end{align*}
In summary, we obtain
	\begin{align*}
		& \mathbb{E} \left[ 2 L(t+1)
		\cdot \mathbf{1}_{S> t+1} \mid \mathcal{F}_t \right] \\
		=& \mathbb{E} \left[ \left\| \begin{bmatrix} \bm{s}(t+1) \\ 
			\bm{h}(t+1) \end{bmatrix} \right\|_2^2
		\cdot \mathbf{1}_{S> t+1} \mid \mathcal{F}_t \right] \\
		=& \mathbb{E} \left[ \left\| \begin{bmatrix} \bm{s}(t+1) \\ \bm{h}(t+1) \end{bmatrix} -
		\begin{bmatrix} \bm{s}(t) \\ \bm{h}(t) \end{bmatrix} +
		\begin{bmatrix} \bm{s}(t) \\ \bm{h}(t) \end{bmatrix} \right\|_2^2
		\cdot \mathbf{1}_{S> t+1} \mid \mathcal{F}_t \right] \\
		\leq& \mathbb{E} \left[ \left\| \begin{bmatrix} \bm{s}(t+1) \\ \bm{h}(t+1) \end{bmatrix} -
		\begin{bmatrix} \bm{s}(t) \\ \bm{h}(t) \end{bmatrix}  \right\|_2^2 
		\cdot \mathbf{1}_{S> t} \mid \mathcal{F}_t \right]
		+ \left\| \begin{bmatrix} \bm{s}(t) \\ \bm{h}(t) \end{bmatrix} \right\|_2^2 \\
		& \qquad \qquad - 2\eta  \begin{bmatrix} \bm{s}(t) \\ \bm{h}(t) \end{bmatrix}^\top 
		\bm{G}_{\theta}(t) \begin{bmatrix} \bm{s}(t) \\ \bm{h}(t) \end{bmatrix}  + 2
		\begin{bmatrix} \bm{s}(t) \\ \bm{h}(t) \end{bmatrix}^\top
		\begin{bmatrix} \chi_{\bm{s}}(t) \\ \chi_{\bm{h}}(t) \end{bmatrix}  \\
		\leq& c_2 \frac{\eta^2}{m} C_B^4 B^{12}L(t) + 2L(t) - 2\eta \cdot \frac{1}{2}\lambda_\theta \cdot 2L(t)
		 	 + 2c_1 \eta^2 \frac{C_B^3 B^9}{\sqrt{m}}L(t) \cdot \sqrt{2L(t)}.
	\end{align*}
	Using the inequality \(\sqrt{L(t)}\leq c_3 \sqrt{L(0)}\) for the last term, we get
	\begin{align*}
		\mathbb{E} \left[ L(t+1)
		\cdot \mathbf{1}_{S> t+1} \mid \mathcal{F}_t \right] & \leq  \left( 1- \eta \lambda_\theta
		+ \frac{1}{2}c_2 \frac{\eta^2}{m} C_B^4 B^{12} + 
		\sqrt{8} c_1c_3 \frac{\eta^2}{\sqrt{m}} C_B^3 B^{9}\sqrt{L(0)}\right) L(t).
	\end{align*}
	If \(m\) and \(\eta\) satisfy the given assumptions,
	then we have 
    $$\frac{\eta}{m} C_B^4 B^{12}\leq \frac{\eta}{\sqrt{m}} C_B^3 B^{9}\sqrt{L(0)}\lesssim \lambda_\theta$$ and
	\begin{align*}
		\mathbb{E} \left[ L(t+1) \cdot \mathbf{1}_{S> t+1} \mid \mathcal{F}_t \right] & \leq  
		\left( 1- \eta \frac{\lambda_\theta}{2} \right) L(t).
	\end{align*}
	Finally, under the above condition, we have
	\begin{align*}
		\mathbb{E} \left[ L(t+1) \cdot \mathbf{1}_{S> t+1} \mid \mathcal{F}_t \right]
		& \leq \min \left\{ \left( 1- \eta \frac{\lambda_\theta}{2} \right) L(t),
		\mathbb{E} \left[ L(t+1) \cdot \mathbf{1}_{S> t} \mid \mathcal{F}_t \right]
		\cdot \mathbf{1}_{S> t}\right\} \\ &\leq \left( 1- \eta \frac{\lambda_\theta}{2} \right) L(t) \cdot \mathbf{1}_{S> t}.
	\end{align*}

\subsection{Proof of Lemma \ref{w(S)-w(0)}}

By the triangle inequality, we have 
\begin{align*}
	& \E \left[ \| \bm{w}_*(S) - \bm{w}_*(0) \|_2 \right] 
	 \leq \E \left[ \sum_{t=0}^{S-1} \| \bm{w}_*(t+1) - \bm{w}_*(t) \|_2 
			\cdot \mathbf{1}_{S> t} \right] \\
	 =& \E \left[ \sum_{t=0}^{S-1} \mathbb{E} \left[ \left\| 
			\bm{w}_*(t+1) - \bm{w}_*(t) \right\|_2
			\cdot \mathbf{1}_{S> t} \mid \mathcal{F}_t \right] \right]  \lesssim \E \left[ \sum_{t=0}^{S-1} \eta \frac{C_B B^3}{\sqrt{m}}\sqrt{L(t)} \right] \\
		 =& \eta \frac{C_B B^3}{\sqrt{m}} \E \left[ \sum_{t=0}^{S-1} \mathbb{E} \left[ \sqrt{L(t)}
		\cdot \mathbf{1}_{S> t-1} \mid \mathcal{F}_{t-1} \right]  \right]  \leq \eta \frac{C_B B^3}{\sqrt{m}} \sum_{t=0}^{\infty} 
		\left( 1- \eta \frac{\lambda_\theta}{2} \right)^{t/2} \sqrt{L(0)}  \\
        \lesssim& \frac{C_B B^3}{\lambda_\theta} \sqrt{ \frac{L(0)}{m} }.
	\end{align*}
	Similarly, we have 
	\[ \E \left[ \| a_r(S) - a_r(0) \|_2 \right]
		\lesssim \frac{C_B B^3}{\lambda_\theta} \sqrt{ \frac{L(0)}{m} }. \]

\subsection{Proof of Theorem \ref{SGDThm}}
First, with probability at least \(1-4\delta\), the event \(\init\) happens if 
\[ m\gtrsim \frac{C_B^4d^6}{\min\left\{\lambda_{\bm{w}}^2,\lambda_{\bm{a}}^2\right\}} 
	\log^3\left(\frac{n_1+n_2}{\delta}\right). \]
Conditioned on this event, by Lemma \ref{w(S)-w(0)} and Markov's inequality, if
\[ m\gtrsim \max\left\{C_B^2 B^6,\log^3\left(\frac{n_1+n_2}{\delta}\right)\right\}\quad \text{ and } \quad
    \eta \lesssim \frac{\lambda_\theta }{C_B^3 B^{9}}\sqrt{\frac{m}{L(0)}}, \]
where $L_0$ is the upper bound of the initial loss \(L(0)\) in Lemma \ref{Loss0}, we have 
\begin{equation}\label{OverRan} 
    \begin{aligned}
		\P\left( \| \bm{w}_*(S) - \bm{w}_*(0) \|_2 > R_{\bm{w}} \right)  
		&\leq \frac{C_B B^3}{\lambda_\theta R_{\bm{w}}} \sqrt{ \frac{L_0}{m} } \\
		\P\left( \| a_*(S) - a_*(0) \|_2 > R_{\bm{a}} \right)  
		&\leq \frac{C_B B^3}{\lambda_\theta R_{\bm{a}}} \sqrt{ \frac{L_0}{m} }
\end{aligned}
\end{equation}
with the quantities $R_{\bm{w}} $ and $ R_{\bm{a}}$ defined in \eqref{RwRa}.
Thus the inequalities for all \(\bm{w}_r\) and \(a_r\) in \eqref{Stopping} hold with probability at least \(1-\wt{\delta}\) if
	\[ m\gtrsim  \frac{C_B^2 B^6 L_0}{{\wt{\delta}}^2 \lambda_\theta^2 R_{\bm{w}}^2}
	=  \frac{C_B^6 d^6 B^{6} L_0}
	{{\wt{\delta}}^2 \lambda_\theta^2 \min\left\{\lambda_{\bm{w}}^2,\lambda_{\bm{a}}^2\right\} }
    = \frac{C_B^8 d^9 B^{6} \log\left(\frac{n_1+n_2}{\delta}\right)}
	{{\wt{\delta}}^2 \lambda_\theta^2 \min\left\{\lambda_{\bm{w}}^2,\lambda_{\bm{a}}^2\right\} }. \]
Note that if all the inequalities in (\ref{Stopping}) are satisfied at step \(S\), then we have
\(S=\infty\) according to the definition of the stopping time.  
Therefore, if \(\eta\) and \(m\) satisfy the above conditions, we deduce that with probability at least \(1-4\delta\), there holds
	\( \P\left( S=\infty \right) \geq  1-\wt{\delta}. \)
In the case \(S=\infty\), the second assertion is implied by Corollary \ref{E[L(t)]}.

\begin{remark}
	If we obtain only the following estimate
	\[ \E \left[ \left\| \bm{w}_r(S) - \bm{w}_r(0) \right\|_2 \right] 
	\lesssim \frac{C_B B^3}{\lambda_\theta} \sqrt{ \frac{L(0)}{m} },\quad \forall r\in [m], \]
	then by Markov's inequality, we have
	\[ \P\left( \| \bm{w}_r(S) - \bm{w}_r(0) \|_2 > R_{\bm{w}} \right)  
	\leq \frac{C_B B^3}{\lambda_\theta R_{\bm{w}}} \sqrt{ \frac{L(0)}{m} },\quad \forall r\in [m]. \]
Then the right hand side has to be made smaller than \(\delta/m\) in order to ensure that \(\bm{w}_r(S)\) stays in a small ball of \(\bm{w}_r(0)\) for all \(r\in [m]\) with high probability, which however is impossible. Thus, we have to control \(\|\bm{w}_r(S) - \bm{w}_r(0)\|_2\) and 
\(\|a_r(S) - a_r(0)\|_2\) uniformly. The strategy is to control the distance using only the loss \(L(t)\) and uniform bound of \(\bm{w}_r\) within the stopping time $S$.
\end{remark}

\section{Proofs for section \ref{SecSGF}} \label{Pro-SecSGF}

\subsection{Proof of Lemma \ref{SDElogLoss}}
We may drop the variable \(\theta_t\) below. The proof employs the following multi-dimensional Ito's formula: for an Ito process \(X_t\) defined by \(\d X_t = b(w,t)\d t + \sigma(w,t)\d W_t\), the process \(Y_t = f(X_t)\) satisfies
\[ \d Y_t = \left( \nabla f \cdot b + \frac{1}{2}\mathrm{tr}\left(\sigma^\top \big(\nabla^2 f\big)\sigma \right) \right)\d t + \nabla f \cdot \sigma \d W_t. \]
We apply this result to $f=\log L$. Then direct computation gives 
\begin{align*}
	\mathrm{Hess}(\log L)  = \nabla \left( \frac{\nabla L}{L} \right)
	= \frac{1}{L^2}\left( L \nabla^2 L - \nabla L \nabla L ^\top \right) 
	= \frac{\mathrm{Hess}(L)}{L} - \frac{\nabla L \nabla L ^\top}{L^2}.
\end{align*}
Note that within the stopping time $t<S$, the loss \(L(\theta)\) is \(\lambda_\theta\)-Polyak-Lojasiewicz.
Using multi-dimensional Ito's formula on the function \(\log L(\theta_t)\) gives
\begin{align*}
	\log L(\theta_t) = &\log L(\theta_0) - \int_{0}^{t} \llan \nabla \log L(\theta_\tau),
	\nabla L(\theta_\tau) \rran \d\tau \\
	& + \eta \int_{0}^{t} \frac{1}{2}\mathrm{tr}
	\left( \sigma(\theta_\tau)^\top \mathrm{Hess}(\log L(\theta_\tau))\sigma(\theta_\tau) \right) \d\tau 
	+ \sqrt{\eta} \int_{0}^{t}  \nabla \log L(\theta_\tau)^\top
	\sigma(\theta_\tau) \d W_\tau  \\
	 =& \log L(\theta_0) - \int_{0}^{t} \frac{ \llan \nabla L(\theta_\tau),
	\nabla L(\theta_\tau) \rran}{L(\theta_\tau)} \d\tau + \eta \int_{0}^{t} \frac{1}{2L}\mathrm{tr}
	\left( \sigma(\theta_\tau)^\top \mathrm{Hess}( L(\theta_\tau)) \sigma(\theta_\tau) \right) \d\tau \\
	& - \eta \int_{0}^{t} \frac{1}{2L^2}\mathrm{tr}
	\left( \sigma(\theta_\tau)^\top \nabla L \nabla L ^\top \sigma(\theta_\tau) \right) \d\tau
	+ \sqrt{\eta} \int_{0}^{t}  
	\frac{\nabla L(\theta_\tau) ^\top \sigma(\theta_\tau)}{L(\theta_\tau)} \d W_\tau \\
	& \leq \log L(\theta_0) - \lambda_\theta t + \frac{1}{2}\eta \int_{0}^{t} \frac{1}{L}\mathrm{tr}
	\left( \sigma(\theta_\tau)^\top \mathrm{Hess}( L(\theta_\tau)) \sigma(\theta_\tau) \right) \d\tau 
	- \frac{1}{2} \langle M \rangle_t + M_t,
\end{align*}
with the indefinite integral \(M_t\) and its quadratic variation \( \langle M \rangle_t \) defined in \eqref{Martin-M_t}. The integrability of \(M_t\) follows from the following inequalities
\begin{align*}
    \left\|\nabla L(\theta_t) \right\|_2^2  & = \left\|\sum_{p=1}^{n_1}s_p \nabla s_p + \sum_{q=1}^{n_2}h_q
        \nabla h_q \right\|_2^2  \lesssim  C_B^2 B^6 L(t),\\
    \left\|\sigma(\theta_t) \right\|_2^2 & \leq \left\|\sigma \right\|_F^2 = 
        \mathrm{tr} \left( \sigma \sigma^\top \right) 
		 = \mathrm{tr} \left( \Sigma \right)
		 = \E\left[ ( \nabla \wt{L} -\nabla L)^\top
		( \nabla \wt{L} -\nabla L) \right] 
		 = \E \left[ \| \nabla \wt{L} \|_2^2 \right] - \| \nabla L \|_2^2 \\
	&\leq \E \left[ \| \nabla \wt{L} \|_2^2 \right] 
         \leq  \E \left[ \wt{L} \cdot
		\left( \sum_{p\in \mathcal{I}} \frac{n_1}{I} \left\| \nabla s_p  \right\|_2^2
		+ \sum_{q\in \mathcal{J}} \frac{ n_2}{J} \left\| \nabla h_q 
		\right\|_2^2 \right) \right] \lesssim C_B^2 B^6 L(t),
\end{align*}
from which we deduce that \(M_t\) is a local martingale. 

\subsection{Proof of Lemma \ref{tr(HessL)}}

Since the matrix \(\sigma(\theta_t) \sigma(\theta_t)^\top\) is positive semidefinite
and \(\mathrm{Hess}(L(\theta_t))\) is symmetric, by von Neumann's trace inequality \cite{Mirsky:1974}, we have 	\begin{align*}
\mathrm{tr} \left( \sigma^\top \mathrm{Hess}(L) \sigma \right)
	= \mathrm{tr} \left( \mathrm{Hess}(L) \sigma \sigma^\top \right)
	\leq \|\mathrm{Hess}(L)\|_2 \, \mathrm{tr} \left( \sigma \sigma^\top \right)
		\leq \|\mathrm{Hess}(L)\|_F \, \mathrm{tr} \left( \sigma \sigma^\top \right).
	\end{align*}
Within the stopping time \(t<S\), the estimates (\ref{sDeriBound}) and (\ref{hDeriBound}) hold. 
Consequently, by the proof of Lemma \ref{SDElogLoss}, we deduce
\[ \mathrm{tr} \left( \sigma(\theta_t) \sigma(\theta_t)^\top \right)
\leq \E \left[ \| \nabla \wt{L}(\theta_t) \|_2^2 \right] \lesssim C_B^2 B^6 L(\theta_t). \]
With $\otimes$ being the Kronecker product, the Hessian of the interior loss \(s_p(\theta)\) is given by 
		\begin{align*}
			\frac{\partial^2 s_p}{\partial \bm{w}_r^2}  = & \frac{1}{\sqrt{n_1 m}} 
			\sum_{i=1}^{d}  a_r \sigma^{(4)}(\bm{w}_r^\top \wt{\bm{x}}_p)w_{ri}^2
			\wt{\bm{x}}_p\otimes \wt{\bm{x}}_p + a_r \sigma^{(3)}(\bm{w}_r^\top \wt{\bm{x}}_p)2w_{ri}
			\wt{\bm{x}}_p\otimes \bm{e}_i \\
			& + a_r \sigma^{(3)}(\bm{w}_r^\top \wt{\bm{x}}_p)2w_{ri}\bm{e}_i \otimes \wt{\bm{x}}_p 
			+ a_r \sigma^{(2)}(\bm{w}_r^\top \wt{\bm{x}}_p)2\bm{e}_i \otimes \bm{e}_i.
		\end{align*}
	For two vectors \(\bm{u}\) and \(\bm{v}\), we have 
	\(\| \bm{u}\otimes \bm{v} \|_2 = \| \bm{u} \|_2 \| \bm{v} \|_2\). 
	Therefore, conditioned on \(t<S\), we get
	\[\left\| \frac{\partial^2 s_p}{\partial \bm{w}_r^2} \right\|_2 
		\lesssim \frac{1}{\sqrt{n_1 m}}C_B (\|\bm{w}_r(0)\|_2^3 +1).\]
Similarly, we have
\begin{align*}
&\frac{\partial^2 s_p}{\partial \bm{w}_r \partial a_r}  = \frac{1}{\sqrt{n_1 m}}
\sum_{i=1}^{d} \sigma^{(3)}(\bm{w}_r^\top \wt{\bm{x}}_p)w_{ri}^2 \wt{\bm{x}}_p 
+ \sigma^{(2)}(\bm{w}_r^\top \wt{\bm{x}}_p)2w_{ri} \bm{e}_i,\\
&\left\| \frac{\partial^2 s_p}{\partial \bm{w}_r \partial a_r} \right\|_2 
\lesssim \frac{1}{\sqrt{n_1 m}}C_B (\|\bm{w}_r(0)\|_2^3 +1),\\	
&\left\| \frac{\partial^2 h_q}{\partial \bm{w}_r^2} \right\|_2 
\lesssim \frac{1}{\sqrt{n_2 m}}C_B (\|\bm{w}_r(0)\|_2 +1),\\ 
&
\left\| \frac{\partial^2 h_q}{\partial \bm{w}_r \partial a_r} \right\|_2 
		\lesssim \frac{1}{\sqrt{n_2 m}}C_B (\|\bm{w}_r(0)\|_2 +1),
	\quad \frac{\partial^2 s_p}{\partial a_r^2} = \frac{\partial^2 h_q}{\partial a_r^2} = 0.
\end{align*}
For all \(r\in [m]\), we have
	\begin{align*}
		\frac{\partial^2 L}{\partial \bm{w}_r^2} 
		 &= \sum_{p=1}^{n_1} s_p \frac{\partial^2 s_p}{\partial \bm{w}_r^2} 
		+ \sum_{q=1}^{n_2} h_q \frac{\partial h_q^2}{\partial \bm{w}_r^2} 
		 + \sum_{p=1}^{n_1}  \frac{\partial s_p}{\partial \bm{w}_r} 
		\otimes  \frac{\partial s_p}{\partial \bm{w}_r} 
		+ \sum_{q=1}^{n_2}  \frac{\partial h_q}{\partial \bm{w}_r} 
		\otimes  \frac{\partial h_q}{\partial \bm{w}_r},\\		\frac{\partial^2 L}{\partial \bm{w}_r \partial a_r} 
		 &= \sum_{p=1}^{n_1} s_p \frac{\partial^2 s_p}{\partial \bm{w}_r \partial a_r} 
		+ \sum_{q=1}^{n_2} h_q \frac{\partial h_q^2}{\partial \bm{w}_r \partial a_r} 
		 + \sum_{p=1}^{n_1}  \frac{\partial s_p}{\partial \bm{w}_r} 
		\otimes  \frac{\partial s_p}{\partial a_r} 
		+ \sum_{q=1}^{n_2}  \frac{\partial h_q}{\partial \bm{w}_r} 
		\otimes  \frac{\partial h_q}{\partial a_r},
	\end{align*}
	and 
	\[ \frac{\partial^2 L}{\partial a_r^2} = 0.  \]
	These second derivatives of \(L\) are bounded by
	\begin{equation*}
		\left\| \frac{\partial^2 L}{\partial \bm{w}_r^2}  \right\|_2^2 
		\lesssim C_B^4 \left(\frac{1}{m} (\|\bm{w}_r(0)\|_2^6 +1)\right)^2 + C_B^2 \frac{1}{m} (\|\bm{w}_r(0)\|_2^6 +1)L.
	\end{equation*}
	For \(\alpha,\beta \in [m]\) such that \( \alpha \neq \beta \), we have
	\begin{gather*}
		\frac{\partial^2 L}{\partial \bm{w}_\alpha \partial \bm{w}_\beta}
		= \sum_{p=1}^{n_1}  \frac{\partial s_p}{\partial \bm{w}_\alpha} 
		\otimes  \frac{\partial s_p}{\partial \bm{w}_\beta} 
		+ \sum_{q=1}^{n_2}  \frac{\partial h_q}{\partial \bm{w}_\alpha} 
		\otimes  \frac{\partial h_q}{\partial \bm{w}_\beta},\\
		\frac{\partial^2 L}{\partial \bm{w}_\alpha \partial a_\beta}
		= \sum_{p=1}^{n_1}  \frac{\partial s_p}{\partial \bm{w}_\alpha} 
		\otimes  \frac{\partial s_p}{\partial a_\beta} 
		+ \sum_{q=1}^{n_2}  \frac{\partial h_q}{\partial \bm{w}_\alpha} 
		\otimes  \frac{\partial h_q}{\partial a_\beta},  \\
		\frac{\partial^2 L}{\partial a_\alpha \partial a_\beta}
		= \sum_{p=1}^{n_1}  \frac{\partial s_p}{\partial a_\alpha} 
		\otimes  \frac{\partial s_p}{\partial a_\beta} 
		+ \sum_{q=1}^{n_2}  \frac{\partial h_q}{\partial a_\alpha} 
		\otimes  \frac{\partial h_q}{\partial a_\beta}.
	\end{gather*}
	These second derivatives of \(L\) are bounded by
	\begin{equation*}
		\left\|\frac{\partial^2 L}{\partial \bm{w}_\alpha \partial \bm{w}_\beta}  \right\|_2^2 
		\lesssim C_B^4 \frac{1}{m} (\|\bm{w}_\alpha(0)\|_2^6 +1)\frac{1}{m} (\|\bm{w}_\beta(0)\|_2^6 +1).
	\end{equation*}
	By the inequality \eqref{moment-w(0)}, we have 
	\begin{gather*}
		\|\mathrm{Hess}(L)\|_F^2 \lesssim C_B^4 \left(\frac{1}{m}\sum_{r=1}^{m} (\|\bm{w}_r(0)\|_2^6 +1)\right)^2 + C_B^2 L \frac{1}{m}\sum_{r=1}^{m} (\|\bm{w}_r(0)\|_2^6 +1) \lesssim C_B^4 d^6 + C_B^2 d^3 L.
	\end{gather*}

\subsection{Proof of Lemma \ref{ConE[L(t)]}}

For \(t<S\), by Lemmas \ref{tr(HessL)} and \ref{Loss(t)}, we have for some  \(c_1\), 
	\begin{gather*}
		\log L(\theta_t) \leq \log L(\theta_0) - \lambda_\theta t + 
		\eta c_1 C_B^3 B^6\left( C_B d^3 + \sqrt{d^3 L(0)}\right)t
		+ M_t - \frac{1}{2} \langle M \rangle_t.
	\end{gather*}
By Lemma \ref{Loss0}, we have 
$$L(0) \lesssim C_B^2 d^3  \log\left( \frac{n_1+n_2}{\delta} \right).$$ 
By taking exponential, we get 
\[ 
L(t) \leq L(0) \exp \left( - \lambda_\theta t + \eta c_1 C_B^4 d^3 B^6 \log^{\frac{1}{2}}\left( \frac{n_1+n_2}{\delta} \right) t  \right)\mathcal{E}_t. 
\]
The desired conclusion follows by taking expectation of both sides of the inequality.

\subsection{Proof of Lemma \ref{Con_w(S)-w(0)}}

	Recall that the component $\bm{w}_r(t)$ of \(\theta_t\) satisfies
	\begin{gather*}
		\bm{w}_r(t) = \bm{w}_r(0) - \int_{0}^{t} \frac{\partial L(\theta_\tau)}{\partial \bm{w}_r} 
		\d\tau + \sqrt{\eta}\xi_{\bm{w}_r}, \quad \forall r\in[m],
	\end{gather*}
	where \(\xi_{\bm{w}_r}\) denotes the corresponding part in the noise vector
	\( \int_{0}^{t} \sigma(\theta_t) \d W_t. \)
Consequently,
	\begin{align*}
		\E\left[ \|\bm{w}_*(t) - \bm{w}_*(0)\|_2 \right] 
		\leq \E\left[ \int_{0}^{t} \left\| 
		\frac{\partial L(\theta_\tau)}{\partial \bm{w}_*} \right\|_2 \d\tau \right] 
		+ \sqrt{\eta} \E\left[ \left\| \int_{0}^{t} \sigma(\theta_\tau) \d W_\tau \right\|_2 \right].
	\end{align*}
Meanwhile, we have
	\begin{align*}
		\left\| \frac{\partial L(\theta_\tau)}{\partial \bm{w}_r} \right\|_2^2 
		\lesssim \frac{C_B^2 B^6}{m}L(\theta_\tau), \quad \forall r\in[m].
	\end{align*}
By the property of Ito integral, we deduce
	\begin{align*}
		\E\left[ \left\| \int_{0}^{t} \sigma(\theta_\tau) \d W_\tau \right\|_2^2 \right]
		& = \E\left[  \int_{0}^{t} \left\| \sigma(\theta_\tau) \right\|_F^2 \d\tau  \right]
		= \E\left[  \int_{0}^{t} \mathrm{tr} 
		\left( \sigma(\theta_\tau) \sigma(\theta_\tau)^\top \right)  \d\tau  \right] \\
		& \lesssim  \E\left[  \int_{0}^{t} C_B^2 B^6 L(\theta_\tau) \d\tau  \right] 
		= C_B^2 B^6  \int_{0}^{t} \E\left[  L(\theta_\tau) \right]  \d\tau\\
		& \leq C_B^2 B^6  \int_{0}^{t} L(0) \exp( - \lambda_\theta \tau/2)  \d\tau  \lesssim \frac{1}{\lambda_\theta} C_B^2 B^6 L(0).
	\end{align*}
	Conditioned on the event \(t<S\), by the Cauchy-Schwarz inequality and Lemma \ref{ConE[L(t)]}, we get
	\begin{align*}
		\E\left[ \|\bm{w}_*(t) - \bm{w}_*(0)\|_2 \right] 
		& \leq \E\left[ \int_{0}^{t} \left\| 
		\frac{\partial L(\theta_\tau)}{\partial \bm{w}_*} \right\|_2 \d\tau \right] 
		+ \sqrt{\eta} \E\left[ \left\| \int_{0}^{t} \sigma(\theta_\tau) \d W_\tau \right\|_2 \right] \\
		& \lesssim \frac{C_B B^3}{\sqrt{m}} \int_{0}^{t}  \sqrt{L(0)} \exp( - \lambda_\theta \tau/4) \d\tau
		+ \sqrt{\eta}\sqrt{ \frac{1}{\lambda_\theta}C_B^2 B^6 L(0) } \\
		& \lesssim \frac{C_B B^3}{\lambda_\theta} \sqrt{ \frac{L(0)}{m} } 
		+ C_B B^3 \sqrt{ \frac{\eta L(0)}{\lambda_\theta} }.
	\end{align*}
	In the same manner, we deduce
	\begin{align*}
		\E\left[ \|a_*(t) - a_*(0)\|_2 \right] \lesssim 
		\frac{C_B B^3}{\lambda_\theta} \sqrt{ \frac{L(0)}{m} } 
		+ C_B B^3 \sqrt{ \frac{\eta L(0)}{\lambda_\theta} }.
	\end{align*}
This completes the proof of the lemma.

\section{Generalization to other linear PDEs} \label{Gen-PDE}

The results in this work can be extended to general linear PDEs with linear boundary conditions since the analysis does not rely much on the specific form of the differential equation or its relevant well-posedness theory. To illustrate this point,  consider the following linear (elliptic, parabolic or hyperbolic) problem 
\begin{equation*}
	\left\{
	\begin{aligned}
		\mathcal{D} [u] &=f, \quad \mbox{in } \Omega, \\
		\mathcal{B}[u]&=g, \quad  \mbox{on }  \partial \Omega,
	\end{aligned}
	\right.
\end{equation*}
where $\mathcal{D} = \sum_{|\alpha|\leq K} a_\alpha(x)\partial^\alpha$ is a linear operator and \(\mathcal{B}\) denotes 
the linear boundary or / and initial condition(s). 
Consider the two-layer neural network \eqref{network} and sampling points $\{\bm{x}_p\}_{p=1}^{n_1}\cup\{{\bm{y}}_q\}_{q=1}^{n_2}$. Then the corresponding PINN loss is given by 
\begin{equation*}
		L(\theta) =
		\sum_{p=1}^{n_1}  \frac{1}{2n_1}\left( \mathcal{D} [\phi(\bm{x}_p;\theta)]-f(\bm{x}_p)\right)^2
		 + \sum_{q=1}^{n_2}\frac{\gamma }{2n_2} 
		\left(\mathcal{B}[\phi(\bm{y}_q;\theta)]-g(\bm{y}_q)\right)^2.
\end{equation*}
Suppose that Assumption \ref{AssAct} (ii) holds for \(\sigma\) up to order \(K+1\) and the associated Gram matrices $\bm{G}_{\bm{w}}^{\infty}$ and $\bm{G}_{\bm{a}}^{\infty}$ are positive definite. (The proof of the assumption may be highly challenging; see Remark \ref{rmk:GramInit} for relevant discussions). Then the estimates in \eqref{sDeriBound} for the interior loss $s_p(\bm{w},\bm{a})$ in Appendix \ref{Pro-StaRan} should be revised as 
\begin{equation*}
    \left\| \frac{\partial s_p(\bm{w},\bm{a})}{\partial \bm{w}_r} \right\|_2, \quad
    \left\| \frac{\partial s_p(\bm{w},\bm{a})}{\partial a_r} \right\|_2  
	\lesssim \frac{C_B}{\sqrt{n_1 m}} \left( \|\bm{w}_r\|_2^{K+1} + 1  \right)	\lesssim \frac{C_B B^{K+1}}{\sqrt{n_1 m}},\\
\end{equation*}
where the implicit constant of the inequality depends on the specific form of the differential operator \(\mathcal{D}\) and the property of its coefficients. This change applies also to the boundary loss $h_q(\bm{w},\bm{a})$. To estimate the initial loss \(L(0)\) in Lemma \ref{Loss0}, the width $m$ of the neural network and the bound on \(L(0)\) satisfy
\[	m \gtrsim \log^K\left( \frac{n_1+n_2}{\delta} \right)\quad \mbox{and} \quad L(0) \lesssim C_B^2 d^{K+1}\log\left(\frac{n_1+n_2}{\delta}\right). \] 
To ensure the positivity of the initial Gram matrices in Lemma \ref{GramInit}, the width \(m\) should satisfy
\[ m\gtrsim \frac{C_B^4d^{2K+2}}
	{\min\left\{\lambda_{\bm{w}}^2,\lambda_{\bm{a}}^2\right\}} 
	\log^{K+1}\left(\frac{n_1+n_2}{\delta}\right). \]
To ensure the positivity of the Gram matrices during training in Lemma \ref{GramDyn}, \(m\) and \(R_{\bm{w}},R_{\bm{a}}\) in \eqref{RwRa} should satisfy 
\[ m\gtrsim \log^{K+1}\left(\frac{1}{\delta}\right)\quad \mbox{and} \quad  R_{\bm{w}}, R_{\bm{a}} \approx 
 		\frac{\min\left\{\lambda_{\bm{w}},\lambda_{\bm{a}}\right\}}{C_B^2 d^{K+1}}. 
\]
For the dynamic randomness, the expectation of the difference of parameters between two consecutive steps in Lemma \ref{StepGap-w_t} is given by 
\begin{align*}
		\mathbb{E} \left[ \left\| \bm{w}_*(t+1) - \bm{w}_*(t) \right\|_2^2 
		\cdot \mathbf{1}_{S> t} \mid \mathcal{F}_t \right],\quad
		\mathbb{E} \left[ \left| a_*(t+1) - a_*(t) \right|^2 
		\cdot \mathbf{1}_{S> t} \mid \mathcal{F}_t \right] 
		\lesssim \eta^2 \frac{C_B^2 B^{2K+2}}{m}L(t). 
	\end{align*}
Then the conditions on \(\eta\) and \(m\) in Lemma \ref{LossGap} should be 
\[ m\gtrsim \max\left\{C_B^2B^{2K+2},\log^{K+1}\left(\frac{n_1+n_2}{\delta}\right)\right\}\quad \mbox{and}\quad
    \eta \lesssim \frac{\lambda_\theta }{C_B^3 B^{3K+3}}\sqrt{\frac{m}{L(0)}}, \]  
and the bounds on the parameters at the stopping time $S$ in Lemma \ref{w(S)-w(0)} read
\[ \E \left[ \left\| \bm{w}_*(S) - \bm{w}_*(0) \right\|_2 \right], \quad
		\E \left[ | a_*(S) - a_*(0) | \right] 
		\lesssim \frac{C_B B^{K+1}}{\lambda_\theta} \sqrt{ \frac{L(0)}{m} }. \]
Finally, Theorem \ref{SGDThm} holds if \(m\) and \(\eta\) satisfy
\begin{align*} 		m&\gtrsim \frac{C_B^4 d^{2K+2}}{\min\left\{\lambda_{\bm{w}}^2,\lambda_{\bm{a}}^2\right\}}
        \log\left( \frac{n_1+n_2}{\delta} \right) \cdot
        \max\left\{ \log^K\left( \frac{n_1+n_2}{\delta} \right),
		\frac{C_B^4 d^{K+1} B^{2K+2} }{{\wt{\delta}}^2 \lambda_\theta^2 }
            \right\},\\
	\eta & \lesssim \frac{\lambda_\theta }{C_B^3 B^{3K+3}}\sqrt{\frac{m}{L(0)}}. 
\end{align*}
Similar changes also apply to the lemmas in Section \ref{SecSGF}. We omit the details and state only the final conclusion. 
Theorem \ref{ConThm} holds with \(B\) and \(C_B\) defined therein if \(m\) and \(\eta\) satisfy 
\begin{align*}
m&\gtrsim \frac{C_B^4 d^{2K+2}}{\min\left\{\lambda_{\bm{w}}^2,\lambda_{\bm{a}}^2\right\}}
        \log\left( \frac{n_1+n_2}{\delta} \right) \cdot
        \max\left\{ \log^K\left( \frac{n_1+n_2}{\delta} \right),
		\frac{C_B^4 d^{K+1} B^{2K+2} }{{\wt{\delta}}^2 \lambda_\theta^2 }
            \right\},\\
\eta &\lesssim \frac{\wt{\delta}^2 \lambda_\theta \min\left\{\lambda_{\bm{w}}^2,\lambda_{\bm{a}}^2\right\}}{C_B^8 d^{3K+3} B^{2K+2}\log\left( \frac{n_1+n_2}{\delta} \right)}.
\end{align*}

\end{document}